\documentclass[manuscript,screen]{acmart}
\usepackage{adjustbox}
\usepackage{array}
\usepackage{bbm}
\AtBeginDocument{%
  \providecommand\BibTeX{{%
    \normalfont B\kern-0.5em{\scshape i\kern-0.25em b}\kern-0.8em\TeX}}}

\setcopyright{rightsretained}
\acmJournal{CSUR}
\acmYear{2023} \acmVolume{1} \acmNumber{1} \acmArticle{1} \acmMonth{1} \acmPrice{}\acmDOI{10.1145/3584741}

\begin{document}

\title{A Survey on Event-based News Narrative Extraction}%

\author{Brian Felipe Keith Norambuena}
\email{briankeithn@vt.edu}
\affiliation{%
  \institution{Virginia Tech}
  \city{Blacksburg}
  \state{Virginia}
  \country{USA}
}
\affiliation{%
  \institution{Universidad Católica del Norte}
  \city{Antofagasta}
  \country{Chile}
}

\author{Tanushree Mitra}
\email{tmitra@uw.edu}
\affiliation{%
  \institution{University of Washington}
  \city{Seattle}
  \state{Washington}
  \country{USA}
}
\author{Chris North}
\email{north@vt.edu}
\affiliation{%
  \institution{Virginia Tech}
  \city{Blacksburg}
  \state{Virginia}
  \country{USA}
}

\newcommand{\rot}[1]{\multicolumn{1}{c}{\adjustbox{angle=60,lap=\width-1em}{#1}}}
\newcommand{\rotn}[1]{\multicolumn{1}{c}{\adjustbox{angle=90,lap=\width-1em}{#1}}}
\renewcommand{\shortauthors}{Keith Norambuena et al.}

\begin{abstract}
Narratives are fundamental to our understanding of the world, providing us with a natural structure for knowledge representation over time. Computational narrative extraction is a subfield of artificial intelligence that makes heavy use of information retrieval and natural language processing techniques. Despite the importance of computational narrative extraction, relatively little scholarly work exists on synthesizing previous research and strategizing future research in the area. In particular, this article focuses on extracting news narratives from an event-centric perspective. Extracting narratives from news data has multiple applications in understanding the evolving information landscape. This survey presents an extensive study of research in the area of event-based news narrative extraction. In particular, we screened over 900 articles that yielded 54 relevant articles. These articles are synthesized and organized by representation model, extraction criteria, and evaluation approaches. Based on the reviewed studies, we identify recent trends, open challenges, and potential research lines.
\end{abstract}

\ccsdesc[500]{Computing methodologies~Information extraction}
\ccsdesc[500]{Computing methodologies~Knowledge representation and reasoning}
\ccsdesc[500]{General and reference~Surveys and overviews}
\keywords{computational narratives, narrative representation, narrative extraction, narrative analysis}

\maketitle

\section{Introduction}
Narratives are fundamental to our understanding of the world \cite{abbott2008cambridge} and they provide a framework that enables humans to associate and represent events over time \cite{burke1969grammar}. Moreover, narratives are a core element of collaborative sensemaking in society \cite{baber2011sensemaking, wilson2018assembling}. In this context, narratives are defined as a \emph{coherent system of interrelated stories} \cite{halverson2011master}, where stories themselves are defined as sequences of \textit{events} \cite{wake2013narrative}. These systems of stories help humans produce a shared understanding of the world \cite{szostek2017defence}. In particular, extracting narratives from data is a fundamental task in our efforts to achieve this goal of common understanding \cite{keith2021narrative}. 

In this survey, we focus on a specific type of narrative: news narratives. In particular, we analyze works that extract computational narrative representations from news articles. Work on general computational narratives started as early as the 1960s \cite{ryan2017grimes}. However, these early works focused mostly on narrative generation---usually through rule-based methods and grammars \cite{alhussain2021automatic}---rather than extracting narratives from data. In contrast, the narrative extraction works reviewed in this survey start around the 2000s (e.g., \cite{uramoto1998method,chieu2004query,nallapati2004event}).

From an information retrieval standpoint, extracting narratives from data relies on several techniques from this field, including event \cite{nallapati2004event} and entity extraction methods \cite{bogel2015time}, as well as elements from search and ranking \cite{liao2021wilson} and summarization techniques \cite{la2021summarize}. Furthermore, narrative extraction is supported by several artificial intelligence techniques, such as machine learning \cite{tannier2013building} and search and optimization \cite{shahaf2010connecting}. 

Despite the importance of news narrative extraction, relatively little work has focused on clarifying the past trajectory and future agenda of news narrative extraction. Our goal with this survey is to fill this gap. This article presents a literature review of narrative extraction screening over 900 papers from a variety of journals, conferences, and workshops. In particular, by thematically analyzing 54 articles we identify a taxonomy of representations, extractions methods, and evaluation methods, that helps organize prior work and chart the path forward for future research. Taken together, all these elements provide a detailed account of the core elements of event-based news narrative extraction.

\subsection{Scope of this Survey and Definitions}
\subsubsection{Narrative Definition}
There are many potential definitions of narrative in the literature. General narrative theory focuses explicitly on understanding the general rules of narrative and its different arrangements that make it meaningful \cite{abbott2008cambridge, puckett2016narrative}. The key intuition in formal narrative theory is that there is a distinction between the story itself and its representation. Narrative theory tries to understand the relationships between stories and their many possible representations \cite{puckett2016narrative}. Other definitions consider narratives as communication tools to construct a shared meaning of events with the purpose of influencing the behaviors \cite{miskimmon2014strategic}. 

Halverson et al. \cite{halverson2011master} define narratives not just as one story, but rather as a \textit{system of stories}. That is, narratives are a systematic collection of interrelated stories with coherent themes. Stories are defined as sequences of \textit{events} tied together in a coherent fashion. In this definition, events are the fundamental units of narrative action, they are either an act involving characters and entities or a happening where no entities are causally involved \cite{abbott2008cambridge}. We leverage this definition to model news narratives. Thus, we have a series of hierarchical definitions starting from the \textit{narrative}, then going into \textit{stories}, and finally into the fundamental units of the narrative: the \textit{event} and its related \textit{entities}. Furthermore, these definitions require an underlying \textit{order} for the events, as they have to be linked sequentially in the stories.

There are two fundamental units in our previous discussion: events and entities. These units provide different perspectives of the narrative, one is focused on the actions and happenings of the narrative, while the other is focused on the characters and other entities that participate in the events. However, to provide a more focused review we will focus exclusively on \textbf{event-based narrative representations}. Thus, we define computational narrative representations as an \textit{event structure} that represents different stories. We note that these event structures are discrete in nature (e.g., a graph or a timeline of events). Nevertheless, we note that some of the extraction methods that we review will leverage entity-based information, but they are not the focus of their representation. 

Finally, we note that the simplest way to computationally represent a narrative is through a linear structure representing sequences of events (i.e., a timeline). In fact, this is the most common approach in our survey. However, we also find that there are more complex representations, based on event graph structures.

\subsubsection{News Narrative Extraction}
The main focus of this survey is on \textbf{narrative extraction} from \textit{news data} (``How do we extract a news narrative from data?''). In particular, we focus exclusively on textual narratives extracted from a set of news articles published in traditional news sources---we exclude works that focus on mixed types of data (e.g., images and text, or videos and text). Thus, all of the surveyed works fall under the umbrella of Natural Language Processing.

Moreover, we note that extraction can be performed at a document level (i.e., extracting a narrative from a single document) or at a corpus level (i.e., extracting a narrative from multiple news articles). As part of our scope definition, we focus on corpus-level extraction methods, where the goal is to obtain a narrative representation from a set of articles, rather than on document-level extraction (e.g., extracting the narrative of a single document).

Throughout this work, we work under the assumption that most news articles focus on a single main event. This is a common assumption in story and narrative extraction methods \cite{keith2021narrative} and a natural assumption when dealing with breaking news articles, as they are likely to present a single event \cite{norambuenaevaluating}. We note that news articles may sometimes refer to previous or secondary events in their body, which can be used to link articles together. However, for the purposes of our definition, these references are not considered the \textit{main event} of that news article. Following this assumption, we deal with three levels of resolution in our works: events as sentences, events as documents, and events as clusters. Events may be represented by relevant sentences extracted from a news article, usually, a single sentence is used for these purposes. Events may also be represented by an entire document (i.e., a news article). We note that there is some overlap between these two representations when documents are associated with headlines. Finally, events may also be represented as sets of documents that refer to the same main event.

We note that there are more granular views of events in the literature—for example, the notion of event from TimeML \cite{pustejovsky2003timeml,minard2015semeval}, where events are a much more specific action (e.g., a perception or state) compared to a news event that may comprise multiple of these events \cite{hu2014exploring}. Contrasting with the granular specifications of TimeML, there are also works that view events as sets of terms (e.g., keywords or entities) \cite{swan2000automatic}, akin to how topics are sometimes characterized in traditional topic modeling works \cite{blei2006dynamic}, and construct timelines representing them as such. However, this view of events is too broad and lacks the specificity expected from news events. Thus, we do not consider narrative representations that use such approaches. Following these exclusion criteria, we removed approximately 10 articles from the final data set.

Leveraging our previous discussion of narratives as a structured system of interrelated stories, we define the (event-based) narrative extraction task as follows:

\fbox{\begin{minipage}{0.96\columnwidth}
\textbf{News Narrative Extraction}: Given a set of news articles, the news narrative extraction task generates a \textit{discrete structure} comprised of \textit{events} to represent the narrative.
\end{minipage}}

We note that the structure is left deliberately ambiguous to allow for different types of representations, such as event timelines or event graphs. However, we note that all these overarching narrative representations are discrete in nature (e.g., event graphs), even if the underlying event representations could be continuous (e.g., text embeddings). Furthermore, the representation of the event itself can be defined in different ways depending on the \textit{resolution} level (sentences, documents, or clusters) of the narrative representation. Furthermore, this definition excludes entity-based representations (e.g., character networks).

\subsubsection{Exclusions: Related Tasks}
We exclude works that focus on narrative generation, narrative forecasting, and narrative analysis. We also exclude works that only focus on representational issues without an associated method.

Narrative generation is a fundamentally different task from extraction that seeks to create new fictional narratives, rather than extract a narrative that already exists (either fictional or non-fictional) \cite{gervas2019long, gatt2018survey}. Furthermore, the focus of narrative generation is usually fictional narratives, not news narratives. Narrative forecasting (i.e., predicting the next events in the narrative) is a task that lies between extraction and generation, but its focus is on generating new events rather than on extracting the complete narrative \cite{zhao2021event}. Narrative analysis methods use existing extraction approaches to obtain a computational representation of the narrative and then use it to analyze the narrative \cite{ofek2013linking, schlachter2015leveraging}. However, they do not provide new insight into the extraction task itself, unless they include a novel extraction method as well.

Moreover, we exclude interactive narratives, as these are a fundamentally different type of narratives where the story can be changed through user feedback and actions \cite{cavazza2006narratology}, which would not make sense in the context of news narratives. However, while the underlying story cannot be changed, it might still be possible to interact with the narrative model. In fact, several works rely on interactivity at a presentation level. 

Finally, we exclude works that focus on news narratives extracted from social media \cite{lin2012generating,bandeli2020framework}, as social media narratives follow a different approach that requires not only analyzing content but also the users spreading it, leading to unique challenges that are left beyond the scope of this survey.

\subsubsection{Inclusion and Exclusion Criteria}
\begin{figure}
    \centering
    \includegraphics[width=0.8\textwidth]{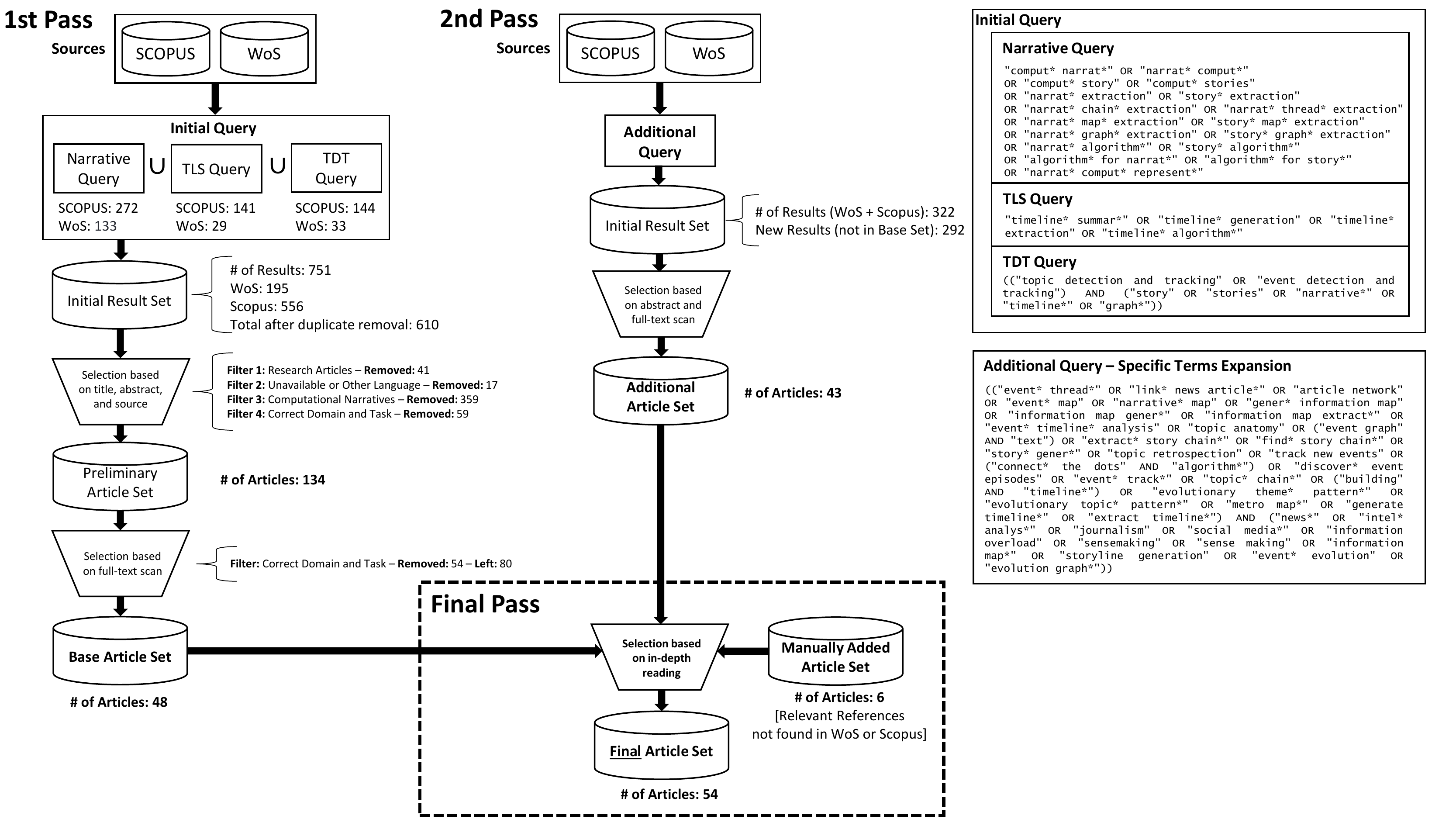}
    \caption{Overview of the article collection process and the inclusion/exclusion criteria used to construct the final article set.}
    \label{fig:inclusion}
\end{figure}

Having defined our scope, we provide details of our collection methodology and inclusion/exclusion criteria. We describe the query and steps used to generate the final article set in Figure \ref{fig:inclusion}.

We performed two article searches on SCOPUS and Web of Science. The first search was based on three queries that covered broad areas related to the news narrative extraction task: narrative extraction and computational narratives in general, topic detection and tracking (TDT) \cite{allan2012topic}, and timeline summarization (TLS) \cite{ghalandari2020examining}. These latter two fields are highly related to our task and provide a series of relevant works that we have examined in our survey. In particular, we note that most TDT works view news as flat collections \cite{nallapati2004event} of events without an underlying narrative structure. Instead, we view news data as an interconnected \textit{structure of events}. Nevertheless, some TDT works fit with our view of narratives and thus we include them in the review. In contrast, we consider most of the TLS line of works as a subset of the narrative extraction task and include many works from that field as part of the ``event as sentences'' resolution level. However, we exclude works that do not generate a full timeline and only focus on identifying relevant dates, as that is a different sub-task. Next, we performed a second query based on a series of keywords obtained from the initial results. We applied the same inclusion and exclusion criteria for this second set of articles. After this, some additional articles that were not caught by our two main searches were added based on references from some reviewed articles. Finally, we performed a final pass on all the articles based on an in-depth reading of each article.

The rest of this article is structured as follows. The rest of Section 1 discusses related surveys and reviews. Section 2 presents an overview and summary of each one of the reviewed articles. Section 3 discusses the different extraction criteria. Section 4 presents a discussion of the evaluation approaches and metrics. Section 5 presents a discussion of our findings and future research directions. This survey concludes with a brief summary and key takeaways in Section 6.

\subsection{Related Surveys}
Most surveys regarding computational narratology focus on the task of narrative generation rather than extraction. In fact, there is an extensive series of survey papers and literature reviews on generation in conferences \cite{valls2017computational} and journals \cite{kybartas2016survey,gervas2019long} that cover narrative generation and its different approaches in-depth. Moreover, there is even a book \cite{mani2012computational} on computational narrative representations for narrative generation and an extensive and in-depth book chapter on different cognitive approaches to narrative generation \cite{ogata2016computational}. Narrative generation is also covered as a specific subtask of the more general field of natural language generation \cite{gatt2018survey}. In contrast, general narrative extraction is not covered by any published survey. More specifically, our domain of interest---news narrative extraction---is also not covered in the literature. However, there are some surveys that touch on related topics. In the rest of this subsection, we provide a general description of these works and how they relate to our own survey. 

First, we note a survey on the evaluation of summarization methods by Ermakova et al. \cite{ermakova2019survey} as a related approach to narrative extraction. In particular, this survey provides a comprehensive overview of existing metrics for the evaluation of narrative summarization methods. Narrative summarization is related to both narrative extraction and generation, as it requires extracting an internal narrative representation from data and then generating the summary. In comparison, our survey presents evaluation metrics for narrative extraction methods, some of which overlap with the evaluation metrics discussed in the aforementioned survey. 

Second, we note the work of Richards et al. \cite{richards2009advancing} which discusses representation models for narratives. Most of the discussion is specific to narrative generation, but there are general models that could be applied in both generation and extraction contexts. Nevertheless, the discussion is focused on what constitutes a narrative in general rather than being directly useful for the news narrative extraction task as defined here.

Third, we note the survey on extracting character networks from fictional narratives by Labatut and Bost \cite{labatut2019extraction}. Their work is related to ours as it focuses on the narrative extraction task, but with a much more specific scope focused on character-based models (i.e., entity-based narrative extraction). In contrast, our work has a different scope that considers event-based models. Moreover, their scope focuses on extracting networks from fictional narratives, while we consider extraction methods for non-fictional narratives in news data.

Finally, we note the recent survey on timeline summarization approaches by Ghalandari and Ifrim \cite{ghalandari2020examining}. While there is plenty of overlap between this survey and our own, the news narrative extraction task that we cover is more general than just timeline summarization, as we include methods that treat events as documents and clusters, rather than at a sentence level. However, we highlight the empirical component of that survey, which includes an experimental section comparing the state-of-the-art methods in timeline summarization.

\section{News Narrative Extraction}
\subsection{Overview}
\begin{figure}[!htb]
    \centering
    \includegraphics[width=0.55\textwidth]{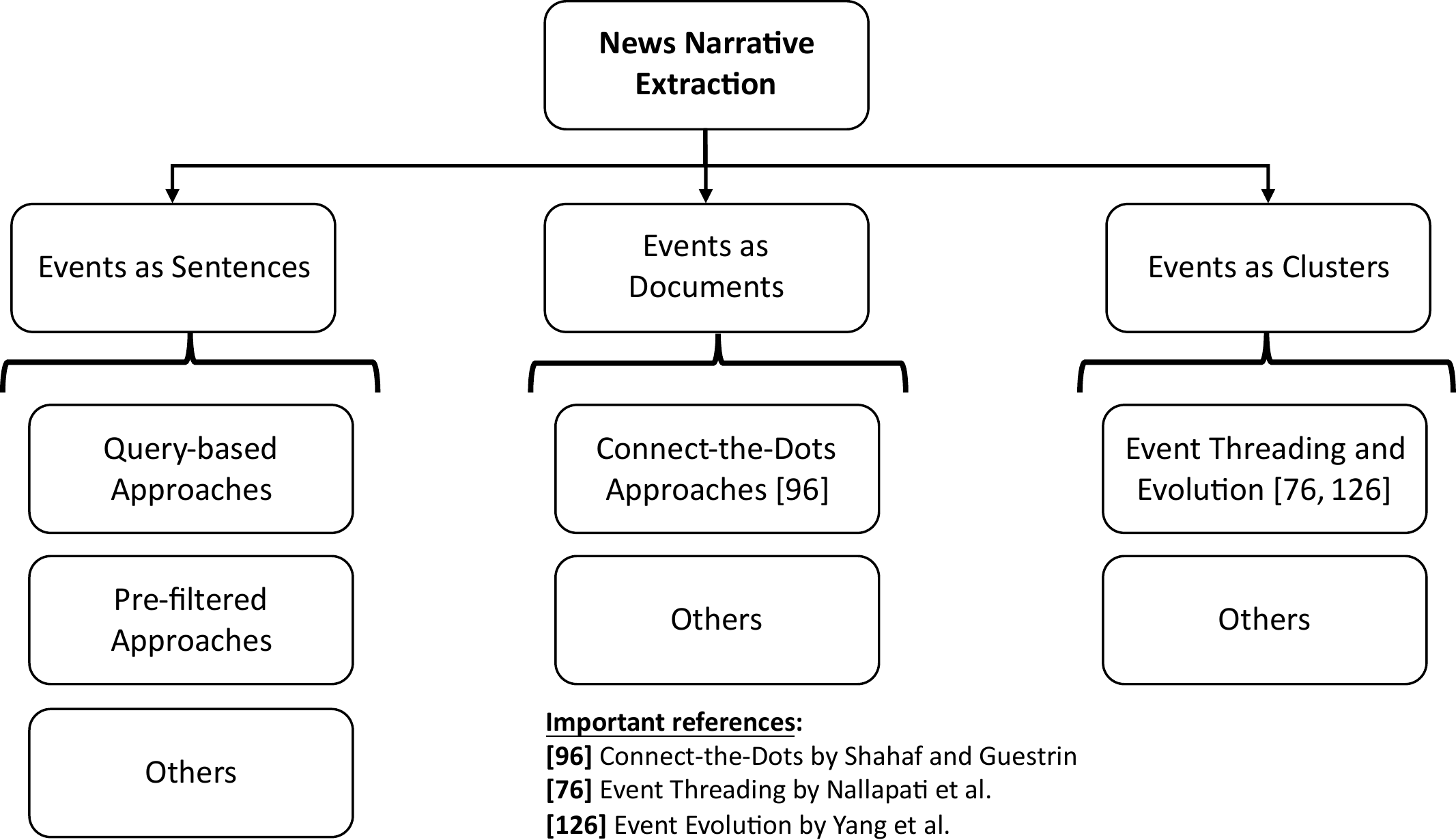}
    \caption{Overview of the different methods used in news narrative extraction categorized by event resolution.}
    \label{fig:methods}
\end{figure}

We found a total of 54 articles focusing on event-based news narrative extraction in our review. We present the articles based on the resolution level that they use: events as sentences, events as documents, and events as clusters. Figure \ref{fig:methods} summarizes the identified approaches categorized by event resolution and some relevant subsets of these categories. In the sentence-level resolution, \textit{query-based approaches} include an information retrieval step in addition to the narrative extraction itself. For example, these approaches require the user of the method to define a search query (e.g., ``COVID'' or ``Terrorism'') to find related documents in the data set through similarity-based techniques or other methods before extracting the narrative from the queried subset. In contrast, \textit{pre-filtered approaches} assume that the data set has been already filtered and do not require an explicit query. In the document-level resolution, \textit{Connect the Dots} approaches refer to the line of works derived from Shahaf and Guestrin's seminal method of the same name on storyline extraction \cite{shahaf2010connecting}. In the cluster-level resolution, event threading and evolution methods refer to a series of works based on Nallapati et al.'s \textit{event threading} concept \cite{nallapati2004event} or Yang et al.'s \textit{event evolution} concept \cite{yang2006tracing}. Works that fall under the ``Others'' do not fit in any of the defined subsets. 

Table \ref{tab:papers} summarizes the reviewed articles. In particular, we include the following columns in this table: \textit{event resolution}, \textit{number of stories}, \textit{structure}, \textit{type of approach}, and \textit{event representation}. We now provide a brief description of these elements and their possible values.

\begin{figure}[!htb]
    \centering
    \includegraphics[width=\textwidth]{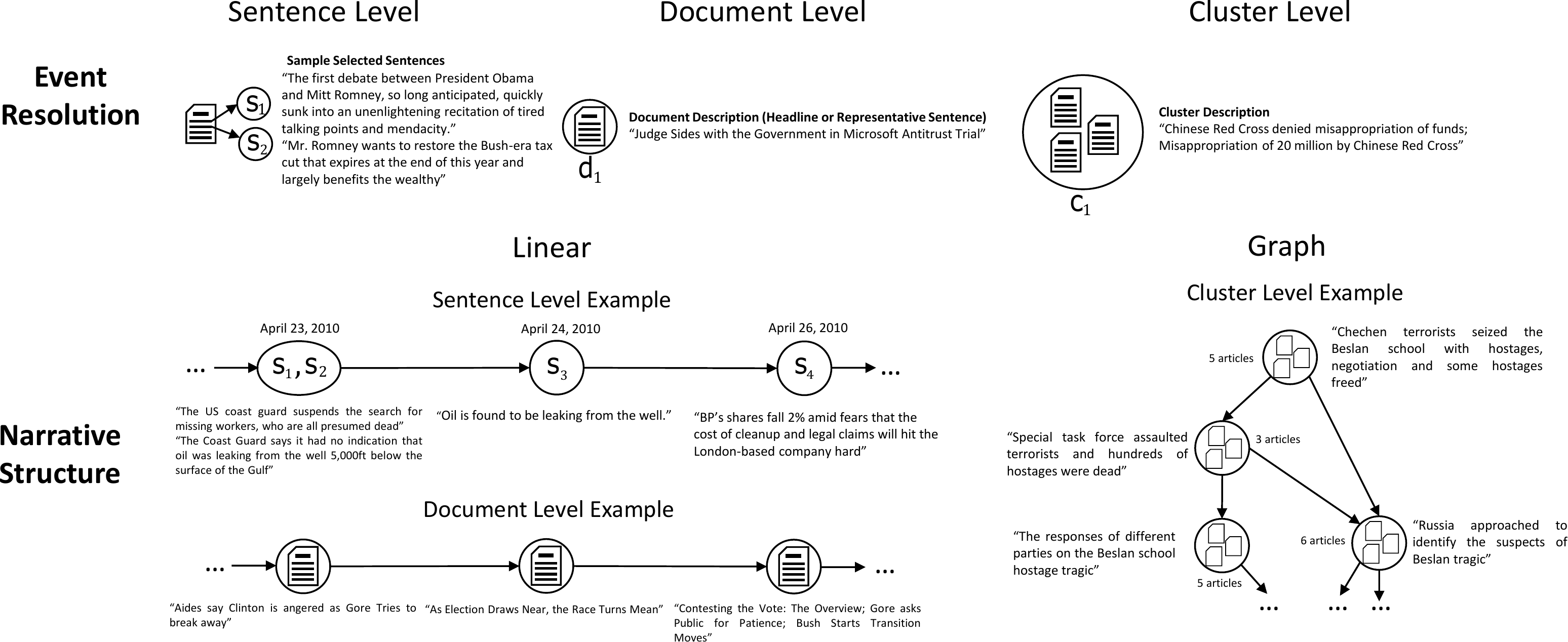}
    \caption{Resolution level and narrative structure. Examples adapted from several works in this survey.}
    \label{fig:examples}
\end{figure}

\textit{Event resolution} refers to the abstraction level at which the events are extracted. As mentioned in the scope definition, we consider three levels: sentences, documents, and clusters. Sentence-level works represent events as either a single sentence (e.g., the most important sentence or a headline) or a set of sentences (e.g., a sample of representative sentences). Document-level works represent events directly as a single document (e.g., a full news article). Cluster-level works represent events as sets of documents (e.g., multiple news articles that talk about the same basic event). Structure represents whether the extraction method generates a linear structure of events (e.g., a timeline \cite{shahaf2010connecting, yan2011timeline}) or a graph-like structure (e.g., a directed acyclic graph \cite{keith2021narrative} or tree \cite{liu2017growing}). Figure \ref{fig:examples} exemplifies these concepts.

\textit{Number of stories} refers to whether the method is designed to handle a single storyline or multiple storylines. Recall that our definition of a story as a sequence of events. Most timeline extraction methods extract a single story, but some of them extract \textit{parallel} timelines, where each timeline represents a different story from the data \cite{yu2021multi, laban2017newslens}. In contrast, most graph-based works are designed to represent multiple storylines, due to their inherent more complex nature compared to timelines. However, there are some works that represent a single story, but provide extra information by exploiting graph structures. For example, appending additional nodes with related events to the central story \cite{lin2008storyline}.

\textit{Type of approach} represents whether the method is supervised---which requires training data---or unsupervised---which does not require training data. In general, we considered any method where the authors had to train the model with labeled data before using it as supervised. However, some approaches only did this to find the optimal value of a small set of hyperparameters \cite{nallapati2004event, yan2011evolutionary, li2015tracking} and it could be possible to use them in an unsupervised manner, provided that those hyperparameters were fixed in some other way (e.g., heuristics or previous work information).

Finally, \textit{event representation} provides information about the computational representation of the events. Note that this is separate from the resolution level of the event. In general, we found four types of representations: word frequency models (e.g., TF-IDF and Bag of Words vectors), topic distribution models (e.g., Latent Dirichlet Allocation (LDA) vectors), neural embeddings (e.g., BERT), and entity-based models (e.g., entity frequency vectors). Some works combine these approaches and have a mixed event representation that leverages all these elements in some way to extract the final narrative model. There are some works that did not fit in any of these approaches and were marked as ``Other''.

\begin{table}[!htb]
\resizebox{\textwidth}{!}{\begin{tabular}{@{}cccccccccccccccc@{}}
\toprule
     &                                                   & \multicolumn{3}{c}{Event Resolution}           & \multicolumn{2}{c}{\# of Stories} & \multicolumn{2}{c}{Structure}   & \multicolumn{2}{c}{Approach}          & \multicolumn{5}{c}{Event Representation}                                 \\ \midrule
Year & Reference                                         & Sentences & Documents & Clusters               & Single         & Multiple         & Linear & Graph                  & Unsupervised & Supervised             & Word Vectors & Topic Distribution & Neural Embeddings & Entities & Other \\ \midrule
1998 & \multicolumn{1}{c|}{\citet{uramoto1998method}}     &           & $\times$         & \multicolumn{1}{c|}{}  &                & $\times$                &        & \multicolumn{1}{c|}{$\times$} & $\times$            & \multicolumn{1}{c|}{}  & $\times$            &                    &                   &          &       \\
2004 & \multicolumn{1}{c|}{\citet{nallapati2004event}}    &           &           & \multicolumn{1}{c|}{$\times$} &                & $\times$                &        & \multicolumn{1}{c|}{$\times$} &              & \multicolumn{1}{c|}{$\times$} & $\times$            &                    &                   & $\times$        &       \\
2004 & \multicolumn{1}{c|}{\citet{chieu2004query}}        &           & $\times$         & \multicolumn{1}{c|}{}  & $\times$              &                  & $\times$      & \multicolumn{1}{c|}{}  & $\times$            & \multicolumn{1}{c|}{}  & $\times$            &                    &                   &          &       \\
2005 & \multicolumn{1}{c|}{\citet{guha2005unweaving}}     &           & $\times$         & \multicolumn{1}{c|}{}  &                & $\times$                & $\times$      & \multicolumn{1}{c|}{}  & $\times$            & \multicolumn{1}{c|}{}  & $\times$            &                    &                   &          &       \\
2006 & \multicolumn{1}{c|}{\citet{yang2006tracing}}       &           &           & \multicolumn{1}{c|}{$\times$} &                & $\times$                &        & \multicolumn{1}{c|}{$\times$} & $\times$            & \multicolumn{1}{c|}{}  & $\times$            &                    &                   &          &       \\
2006 & \multicolumn{1}{c|}{\citet{lin2006topic}}          &           &           & \multicolumn{1}{c|}{$\times$} & $\times$              &                  &        & \multicolumn{1}{c|}{$\times$} & $\times$            & \multicolumn{1}{c|}{}  & $\times$            &                    &                   &          &       \\
2007 & \multicolumn{1}{c|}{\citet{lin2007individualized}} &           &           & \multicolumn{1}{c|}{$\times$} & $\times$              &                  &        & \multicolumn{1}{c|}{$\times$} & $\times$            & \multicolumn{1}{c|}{}  & $\times$            &                    &                   &          &       \\
2008 & \multicolumn{1}{c|}{\citet{chen2008tscan}}         &           &           & \multicolumn{1}{c|}{$\times$} &                & $\times$                &        & \multicolumn{1}{c|}{$\times$} & $\times$            & \multicolumn{1}{c|}{}  & $\times$            & $\times$                  &                   &          &       \\
2008 & \multicolumn{1}{c|}{\citet{qiu2008timeline}}       &           &           & \multicolumn{1}{c|}{$\times$} &                & $\times$                &        & \multicolumn{1}{c|}{$\times$} &              & \multicolumn{1}{c|}{$\times$} & $\times$            &                    &                   &          &       \\
2008 & \multicolumn{1}{c|}{\citet{lin2008storyline}}      &           &           & \multicolumn{1}{c|}{$\times$} & $\times$              &                  &        & \multicolumn{1}{c|}{$\times$} & $\times$            & \multicolumn{1}{c|}{}  & $\times$            &                    &                   &          &       \\
2009 & \multicolumn{1}{c|}{\citet{yang2009discovering}}   &           &           & \multicolumn{1}{c|}{$\times$} &                & $\times$                &        & \multicolumn{1}{c|}{$\times$} & $\times$            & \multicolumn{1}{c|}{}  & $\times$            &                    &                   &          &       \\
2010 & \multicolumn{1}{c|}{\citet{shahaf2010connecting}}  &           & $\times$         & \multicolumn{1}{c|}{}  & $\times$              &                  & $\times$      & \multicolumn{1}{c|}{}  & $\times$            & \multicolumn{1}{c|}{}  & $\times$            &                    &                   &          &       \\
2011 & \multicolumn{1}{c|}{\citet{yan2011evolutionary}}   & $\times$         &           & \multicolumn{1}{c|}{}  & $\times$              &                  & $\times$      & \multicolumn{1}{c|}{}  &              & \multicolumn{1}{c|}{$\times$} & $\times$            &                    &                   &          &       \\
2011 & \multicolumn{1}{c|}{\citet{yan2011timeline}}       & $\times$         &           & \multicolumn{1}{c|}{}  & $\times$              &                  & $\times$      & \multicolumn{1}{c|}{}  &              & \multicolumn{1}{c|}{$\times$} & $\times$            &                    &                   &          &       \\
2011 & \multicolumn{1}{c|}{\citet{hu2011generating}}      & $\times$         &           & \multicolumn{1}{c|}{}  & $\times$              &                  & $\times$      & \multicolumn{1}{c|}{}  & $\times$            & \multicolumn{1}{c|}{}  & $\times$            & $\times$                  &                   &          &       \\
2011 & \multicolumn{1}{c|}{\citet{khurdiya2011multi}}     &           &           & \multicolumn{1}{c|}{$\times$} &                & $\times$                &        & \multicolumn{1}{c|}{$\times$} & $\times$            & \multicolumn{1}{c|}{}  &              & $\times$                  &                   &          &       \\
2012 & \multicolumn{1}{c|}{\citet{zhu2012finding}}        &           & $\times$         & \multicolumn{1}{c|}{}  & $\times$              &                  & $\times$      & \multicolumn{1}{c|}{}  & $\times$            & \multicolumn{1}{c|}{}  & $\times$            &                    &                   &          &       \\
2012 & \multicolumn{1}{c|}{\citet{chen2010tscan}}         &           &           & \multicolumn{1}{c|}{$\times$} &                & $\times$                &        & \multicolumn{1}{c|}{$\times$} & $\times$            & \multicolumn{1}{c|}{}  & $\times$            & $\times$                  &                   &          &       \\
2012 & \multicolumn{1}{c|}{\citet{shahaf2012connecting}}  &           & $\times$         & \multicolumn{1}{c|}{}  & $\times$              &                  & $\times$      & \multicolumn{1}{c|}{}  & $\times$            & \multicolumn{1}{c|}{}  & $\times$            &                    &                   &          &       \\
2012 & \multicolumn{1}{c|}{\citet{shahaf2012trains}}      &           & $\times$         & \multicolumn{1}{c|}{}  &                & $\times$                &        & \multicolumn{1}{c|}{$\times$} & $\times$            & \multicolumn{1}{c|}{}  & $\times$            &                    &                   &          &       \\
2013 & \multicolumn{1}{c|}{\citet{binh2013predicting}}    & $\times$         &           & \multicolumn{1}{c|}{}  & $\times$              &                  & $\times$      & \multicolumn{1}{c|}{}  &              & \multicolumn{1}{c|}{$\times$} & $\times$            &                    &                   &          &       \\
2013 & \multicolumn{1}{c|}{\citet{li2013evolutionary}}    & $\times$         &           & \multicolumn{1}{c|}{}  & $\times$              &                  & $\times$      & \multicolumn{1}{c|}{}  &              & \multicolumn{1}{c|}{$\times$} &              & $\times$                  &                   &          &       \\
2013 & \multicolumn{1}{c|}{\citet{tran2013leveraging}}    & $\times$         &           & \multicolumn{1}{c|}{}  & $\times$              &                  & $\times$      & \multicolumn{1}{c|}{}  &              & \multicolumn{1}{c|}{$\times$} & $\times$            &                    &                   &          &       \\
2013 & \multicolumn{1}{c|}{\citet{huang2013optimized}}    & $\times$         &           & \multicolumn{1}{c|}{}  & $\times$              &                  & $\times$      & \multicolumn{1}{c|}{}  & $\times$            & \multicolumn{1}{c|}{}  &              & $\times$                  &                   &          &       \\
2013 & \multicolumn{1}{c|}{\citet{tannier2013building}}   &           & $\times$         & \multicolumn{1}{c|}{}  &                & $\times$                &        & \multicolumn{1}{c|}{$\times$} &              & \multicolumn{1}{c|}{$\times$} & $\times$            &                    &                   &          &       \\
2013 & \multicolumn{1}{c|}{\citet{shahaf2013metro}}       &           & $\times$         & \multicolumn{1}{c|}{}  &                & $\times$                &        & \multicolumn{1}{c|}{$\times$} & $\times$            & \multicolumn{1}{c|}{}  & $\times$            &                    &                   &          &       \\
2013 & \multicolumn{1}{c|}{\citet{shahaf2013information}} &           &           & \multicolumn{1}{c|}{$\times$} &                & $\times$                &        & \multicolumn{1}{c|}{$\times$} & $\times$            & \multicolumn{1}{c|}{}  & $\times$            &                    &                   &          &       \\
2014 & \multicolumn{1}{c|}{\citet{nguyen2014ranking}}     & $\times$         &           & \multicolumn{1}{c|}{}  & $\times$              &                  & $\times$      & \multicolumn{1}{c|}{}  & $\times$            & \multicolumn{1}{c|}{}  & $\times$            &                    &                   &          &       \\
2014 & \multicolumn{1}{c|}{\citet{zhu2014finding}}        &           & $\times$         & \multicolumn{1}{c|}{}  & $\times$              &                  & $\times$      & \multicolumn{1}{c|}{}  & $\times$            & \multicolumn{1}{c|}{}  & $\times$            &                    &                   &          &       \\
2014 & \multicolumn{1}{c|}{\citet{huang2014discovering}}  &           &           & \multicolumn{1}{c|}{$\times$} &                & $\times$                &        & \multicolumn{1}{c|}{$\times$} & $\times$            & \multicolumn{1}{c|}{}  & $\times$            & $\times$                  &                   & $\times$        &       \\
2014 & \multicolumn{1}{c|}{\citet{wei2014exploiting}}     &           &           & \multicolumn{1}{c|}{$\times$} &                & $\times$                &        & \multicolumn{1}{c|}{$\times$} & $\times$            & \multicolumn{1}{c|}{}  & $\times$            &                    &                   &          &       \\
2014 & \multicolumn{1}{c|}{\citet{hu2014exploring}}       &           & $\times$         & \multicolumn{1}{c|}{}  &                & $\times$                &        & \multicolumn{1}{c|}{$\times$} & $\times$            & \multicolumn{1}{c|}{}  &              & $\times$                  &                   &          &       \\
2014 & \multicolumn{1}{c|}{\citet{zhou2014generating}}    & $\times$         &           & \multicolumn{1}{c|}{}  &                & $\times$                &        & \multicolumn{1}{c|}{$\times$} & $\times$            & \multicolumn{1}{c|}{}  & $\times$            &                    &                   &          &       \\
2015 & \multicolumn{1}{c|}{\citet{tran2015timeline}}      & $\times$         &           & \multicolumn{1}{c|}{}  & $\times$              &                  & $\times$      & \multicolumn{1}{c|}{}  &              & \multicolumn{1}{c|}{$\times$} & $\times$            &                    &                   &          &       \\
2015 & \multicolumn{1}{c|}{\citet{li2015tracking}}        & $\times$         &           & \multicolumn{1}{c|}{}  & $\times$              &                  & $\times$      & \multicolumn{1}{c|}{}  &              & \multicolumn{1}{c|}{$\times$} &              & $\times$                  &                   &          &       \\
2015 & \multicolumn{1}{c|}{\citet{bogel2015time}}         &           & $\times$         & \multicolumn{1}{c|}{}  & $\times$              &                  &        & \multicolumn{1}{c|}{$\times$} & $\times$            & \multicolumn{1}{c|}{}  & $\times$            &                    &                   & $\times$        &       \\
2015 & \multicolumn{1}{c|}{\citet{chen2015multi}}         & $\times$         &           & \multicolumn{1}{c|}{}  & $\times$              &                  & $\times$      & \multicolumn{1}{c|}{}  & $\times$            & \multicolumn{1}{c|}{}  & $\times$            & $\times$                  &                   &          & $\times$     \\
2015 & \multicolumn{1}{c|}{\citet{shahaf2015information}} &           &           & \multicolumn{1}{c|}{$\times$} &                & $\times$                &        & \multicolumn{1}{c|}{$\times$} & $\times$            & \multicolumn{1}{c|}{}  & $\times$            &                    &                   &          &       \\
2017 & \multicolumn{1}{c|}{\citet{wu2017event}}           & $\times$         &           & \multicolumn{1}{c|}{}  & $\times$              &                  & $\times$      & \multicolumn{1}{c|}{}  & $\times$            & \multicolumn{1}{c|}{}  & $\times$            &                    &                   &          &       \\
2017 & \multicolumn{1}{c|}{\citet{liu2017growing}}        &           &           & \multicolumn{1}{c|}{$\times$} &                & $\times$                &        & \multicolumn{1}{c|}{$\times$} &              & \multicolumn{1}{c|}{$\times$} & $\times$            &                    &                   &          &       \\
2017 & \multicolumn{1}{c|}{\citet{laban2017newslens}}     &           & $\times$         & \multicolumn{1}{c|}{}  &                & $\times$                & $\times$      & \multicolumn{1}{c|}{}  & $\times$            & \multicolumn{1}{c|}{}  & $\times$            &                    &                   &          &       \\
2018 & \multicolumn{1}{c|}{\citet{wang2018event}}         & $\times$         &           & \multicolumn{1}{c|}{}  & $\times$              &                  & $\times$      & \multicolumn{1}{c|}{}  &              & \multicolumn{1}{c|}{$\times$} &              & $\times$                  &                   & $\times$        &       \\
2018 & \multicolumn{1}{c|}{\citet{tikhomirov2017news}}    & $\times$         &           & \multicolumn{1}{c|}{}  & $\times$              &                  & $\times$      & \multicolumn{1}{c|}{}  &              & \multicolumn{1}{c|}{$\times$} & $\times$            &                    & $\times$                 &          &       \\
2018 & \multicolumn{1}{c|}{\citet{xu2018generating}}      &           &           & \multicolumn{1}{c|}{$\times$} &                & $\times$                &        & \multicolumn{1}{c|}{$\times$} & $\times$            & \multicolumn{1}{c|}{}  &              &                    & $\times$                 &          &       \\
2018 & \multicolumn{1}{c|}{\citet{zhou2018new}}           & $\times$         &           & \multicolumn{1}{c|}{}  &                & $\times$                &        & \multicolumn{1}{c|}{$\times$} & $\times$            & \multicolumn{1}{c|}{}  & $\times$            &                    &                   &          &       \\
2019 & \multicolumn{1}{c|}{\citet{camacho2019analyzing}}  &           & $\times$         & \multicolumn{1}{c|}{}  &                & $\times$                &        & \multicolumn{1}{c|}{$\times$} & $\times$            & \multicolumn{1}{c|}{}  & $\times$            & $\times$                  &                   & $\times$        &       \\
2019 & \multicolumn{1}{c|}{\citet{cai2019temporal}}       &           &           & \multicolumn{1}{c|}{$\times$} &                & $\times$                &        & \multicolumn{1}{c|}{$\times$} & $\times$            & \multicolumn{1}{c|}{}  & $\times$            &                    &                   &          &       \\
2019 & \multicolumn{1}{c|}{\citet{yuan2019dtexsl}}        & $\times$         &           & \multicolumn{1}{c|}{}  &                & $\times$                &        & \multicolumn{1}{c|}{$\times$} & $\times$            & \multicolumn{1}{c|}{}  &              &                    & $\times$                 &          &       \\
2020 & \multicolumn{1}{c|}{\citet{duan2020comparative}}   & $\times$         &           & \multicolumn{1}{c|}{}  & $\times$              &                  & $\times$      & \multicolumn{1}{c|}{}  & $\times$            & \multicolumn{1}{c|}{}  & $\times$            &                    & $\times$                 &          &       \\
2020 & \multicolumn{1}{c|}{\citet{liu2020story}}          &           &           & \multicolumn{1}{c|}{$\times$} &                & $\times$                &        & \multicolumn{1}{c|}{$\times$} &              & \multicolumn{1}{c|}{$\times$} & $\times$            &                    &                   &          &       \\
2021 & \multicolumn{1}{c|}{\citet{la2021summarize}}       & $\times$         &           & \multicolumn{1}{c|}{}  & $\times$              &                  & $\times$      & \multicolumn{1}{c|}{}  & $\times$            & \multicolumn{1}{c|}{}  &              &                    &                   &          & $\times$     \\
2021 & \multicolumn{1}{c|}{\citet{yu2021multi}}           & $\times$         &           & \multicolumn{1}{c|}{}  &                & $\times$                & $\times$      & \multicolumn{1}{c|}{}  &              & \multicolumn{1}{c|}{$\times$} &              &                    & $\times$                 &          &       \\
2021 & \multicolumn{1}{c|}{\citet{liao2021wilson}}        & $\times$         &           & \multicolumn{1}{c|}{}  & $\times$              &                  & $\times$      & \multicolumn{1}{c|}{}  & $\times$            & \multicolumn{1}{c|}{}  &              &                    & $\times$                 &          &       \\
2021 & \multicolumn{1}{c|}{\citet{keith2021narrative}}    &           & $\times$         & \multicolumn{1}{c|}{}  &                & $\times$                &        & \multicolumn{1}{c|}{$\times$} & $\times$            & \multicolumn{1}{c|}{}  &              &                    & $\times$                 &          &       \\ \bottomrule
\end{tabular}}
\caption{Summary of the surveyed articles.}
\label{tab:papers}
\end{table}

\subsection{Events as Sentences}
We start with works that use a sentence-level resolution. Most of these methods fall under the umbrella of timeline summarization \cite{ghalandari2020examining}. However, not all of them fit with traditional TLS work. We split the discussion into three parts: query-based approaches, pre-filtered approaches, and others.

\subsubsection{Query-based Approaches}
These approaches perform an information retrieval step before or during the narrative extraction process based on a user-defined query. In some cases, the query just acts as a simple filter, in others, they explicitly include the query into the narrative extraction model.

Chieu and Lee \cite{chieu2004query} present a query-based timeline extraction approach where each event is represented as a sentence. This is the earliest form of the ``events as sentences'' that we could find in the literature. Sentences are first filtered based on the query and then ranked according to two criteria: \textit{interest}---based on the frequency of the reported event in the query---and \textit{burstiness}---based on the idea that important events form clusters around their date of occurrence. To determine whether two sentences are reporting the same event, the authors use cosine similarity. Furthermore, interest is determined based on a time window to avoid combining events that should be separated due to their temporal distance. To reduce \textit{redundancy}, duplicated sentences are removed based on a time window around an important event that depends on the interest value.

Yan et al. \cite{yan2011evolutionary} proposed a timeline summarization method based on balanced optimization and iterative substitution of sentences. Their optimization problem is defined in terms of \textit{relevance}, \textit{coverage}, \textit{coherence}, and \textit{diversity}. All these terms are based on the Kullback-Leibler divergence (KLD) \cite{kullback1951information} of the summary items with a target distribution. Relevance is related to a user-defined query and is defined as the KLD between the summary items and the internal representation of the query. Coverage is based on a global term---KLD between the summary items and the whole corpus---and a local term---KLD between the summary items and the set of sentences from the same date. Coherence is defined locally, based on the KLD between each summary item and its neighboring summaries by using an exponential temporal decay term (i.e., consecutive dates should have relatively similar summaries). Diversity is measured across dates and measures the average KLD of each sentence with respect to all other sentences in a leave-one-out manner. The final utility function is a weighted average of these terms with user-defined weights and can be defined at a local level (to evaluate individual time periods) and a global level (to evaluate the full timeline). To find the sentences, this utility function is optimized in an iterative manner by replacing sentences in the date summaries and improving the utility value in each step using a dynamic programming algorithm that considers both local and global constraints. 

Li et al. \cite{li2013evolutionary} propose a topic modeling approach for timeline extraction from news called Evolutionary Hierarchical Dirichlet Process (EHDP) to capture the evolution pattern of news topics. This model extends Hierarchical Dirichlet Process models \cite{teh2006hierarchical} by incorporating time dependencies and background information. In particular, it adds a new dynamic Dirichlet mixture model. Using this proposed topic model, a series of sentences are selected to represent each time period in the timeline based on the weighted average of three criteria: \textit{relevance} (the summary should be related to the overall query), \textit{coverage} (the summary should generalize the important topics in each time period), and \textit{coherence} (each summary should be coherent with neighboring time periods). To score these criteria, the authors propose a topic scoring algorithm based on KLD that leverages their new topic model. The selected sentences are used to represent the relevant events in each time period.

RaRE (Rank and RErank) \cite{nguyen2014ranking} is a system for building timelines of events from news articles based on a user query. In particular, it extracts timelines in three steps: temporal clustering based on salient dates, event relevance, salience scoring, and sentence re-ranking using an iterative algorithm that seeks to reduce redundancy. The method has an underlying assumption that each document represents a single event that can be described by a single sentence. The temporal clustering step identifies salient dates based on the \textit{number of occurrences} of the date in the documents. The sets of events linked to a specific salient date are called \textit{temporal clusters}. Furthermore, as a preprocessing step, events are clustered into \textit{thematic clusters} inside each date using hierarchical clustering based on normalized Manhattan distance and a user-specified threshold. The event relevance and salience scoring steps use these criteria to rank events (i.e., documents) inside each temporal cluster. In particular, it uses four metrics: \textit{event relevance}, \textit{thematic cluster relevance}, \textit{event salience}, and \textit{date salience}. Event relevance is based on cosine similarity with the initial query. Thematic cluster relevance is based on the similarity of its thematic cluster with the initial query based on the average relevance of each event in the cluster. Event salience is based on the frequency of terms on a specific date. Date salience is based on the (normalized) total relevance of all events happening on that date. Finally, the sentence re-ranking step measures the frequency of unused terms on each date for a specific event to reduce redundancy.

Another topic modeling approach uses a time-dependent Hierarchical Dirichlet Tree Model \cite{li2015tracking} to capture the evolution of news topics using the Dirichlet Tree distribution---a generalization of the Dirichlet distribution \cite{dennis1991hyper}. In particular, the model represents topic distributions in sentences using a tree of fixed depth. Each sentence is associated with a path and with a topic vector and each node has its own topic distribution over words. Using the proposed topic model, sentences are selected by first locating candidate words on the nodes of the tree based on the Jensen-Shannon (JS) divergence of sentences and KLD between word collections. Next, the candidate sentences are scored based on the weighted average of the following criteria: \textit{focus} (the timeline should be relevant to a given query), \textit{coherence} (the sentences should be correlated), and \textit{coverage} (the sentences and documents should be representative).

Wu et al. \cite{wu2017event} propose a sentence-based approach to generate timelines. In particular, all the sentences that contain a user-defined query word are split by date and used to generate a \textit{date vector} representing that specific date. Sentences that do not include parseable dates are grouped based on \textit{similarity} with the date vector. All sentences are then ranked based on similarity with their corresponding date vector and unrelated sentences are filtered out based on a user-defined threshold. The highest-ranking sentence is used to summarize each date. 

Tikhomirov and Dobrov \cite{tikhomirov2017news} propose a news timeline generation approach from a query based on three steps: query extension, inter-document graph extraction, and intra-document sentence ranking. Query extension is based on \textit{pseudo-relevance feedback} and consists of three query levels, which are constructed using the most significant terms based on TF-IDF weights. Next, as a preprocessing step, dates that have a \textit{frequency} below a statistically determined threshold are discarded. The next two steps use an \textit{inverted pyramid} \cite{norambuenaevaluating} heuristic, which assumes that the upper part of the article contains the most important information and the lower part of the article may contain \textit{references} to important events from the past. In particular, the inter-document graph extraction step constructs a \textit{similarity matrix} between the upper and lower parts of the documents. If the similarity is above a specified threshold, then the articles are considered to be linked, creating a similarity graph. Next, a ranking algorithm---LexRank \cite{erkan2004lexrank}---is used to determine the importance of each document. Documents that are above a specified importance threshold are used to further expand the original query one more time. Finally, to rank the final selected sentences for the summary, a ranking metric is defined by taking into account content similarity (using cosine similarity) with the extended query (i.e., maximizing relevance) and subtracting similarity with already extracted sentences (i.e., minimizing redundancy).

WILSON (neWs tImeLine SummarizatiON) \cite{liao2021wilson} is a query-based timeline summarization method for news based on a divide-and-conquer approach consisting of two major components: date selection and text summarization for each selected date. For date selection, the method first tags temporal expressions in sentences and constructs a date reference graph based on these annotations. Next, the method assigns weights to the edges of the date reference graph by taking the product of the \textit{number of references} and \textit{temporal distances} with the references. Then, it uses the PageRank algorithm \cite{page1999pagerank} on the extracted graph to find the most salient dates. However, this approach leads to a bias towards older dates, as they have had more time to get references. Thus, the model is augmented with an exponential \textit{recency adjustment} weight, which is used to initialize the Personalized PageRank algorithm \cite{bahmani2010fast}, which allows for non-uniform initial distributions. Next, the daily summarization can be done using any multi-document summarization approach. Specifically, the authors use TextRank \cite{mihalcea2004textrank} based on BERT \cite{devlin2018bert} representations to generate the summaries.

\subsubsection{Pre-filtered Approaches}
These approaches assume that the data set has already been filtered as part of a pre-processing step. Thus, they do not explicitly model the query in their extraction model.

Yan et al. \cite{yan2011timeline} propose a system to generate news timelines using a trans-temporal summarization approach, where the summary for each time period depends on its context---that is, nearby time periods. Before generating the timeline, the system chooses the important time periods (e.g., specific days) to be summarized based on \textit{burstiness}. The timeline extraction approach is based on two components: a \textit{global component}---which defines the structure of the overall summary and the inter-temporal relationships between each period of the timeline---and a \textit{local component}---which defines the summary in each time period. The global component is based on a global graph that uses inter-date dependency, which is computed using \textit{temporal proximity} and a global \textit{affinity} model for each sentence based on PageRank. Furthermore, to ensure a diverse set of sentences in the global component, the system incorporates DivRank into the affinity model \cite{mei2010divrank} to penalize the lack of \textit{diversity} in the sentence selection. Next, the local component is based on a local sentence graph for each time period following a similar approach to the global graph. To generate the final sentence selection in each time period, the system optimizes a weighted ranking generated by both components.

Hu et al. \cite{hu2011generating} propose a timeline overview method for news based on the concept of breakpoints---points in time where a significant development or change occurs (i.e., important events). Their extraction approach consists of three steps. First, they analyze \textit{topic activity} using a Topic-Activeness Hidden Markov Model (HMM) and discard inactive periods. In practice, this is done by measuring whether there is new information using KLD and document frequency. Next, the breakpoints are identified by detecting topic variations in each time period using a topic mixture model---in particular, a generative probabilistic mixture model \cite{mei2005discovering}---and a Theme-Transition HMM model to model topic evolution. Specifically, breakpoints are identified by using JS divergence to measure \textit{topic variation} between two consecutive time points. Then, a summary for each breakpoint is generated by selected representative sentences---based on Jaccard \textit{similarity with topic keywords} and \textit{relevant entities}.

Tran et al. \cite{binh2013predicting} present a supervised learning method to extract timelines from news articles based on Linear Regression. Their model first identifies salient dates based on \textit{burstiness} (i.e., high-frequency periods), and then selects the most representative sentences from the news articles on each of these dates. In particular, the model uses \textit{surface-level features} (e.g., length and position of the sentences), \textit{coherence features} (e.g., causal and temporal signals), \textit{topic features} (e.g., TF-IDF information and cross-entropy), and \textit{time-related features} (e.g., popularity over time and use of temporal expressions) to determine the key sentences of each date. Subsequent work by Tran et al. \cite{tran2013leveraging} used SVM-Rank instead of linear regression and expanded upon this supervised framework. In particular, they leverage three metrics to evaluate the event sentences: \textit{relevance}, \textit{novelty}, and \textit{continuity}. Relevance is learned using the SVM-Rank mentioned earlier. Novelty is evaluated by measuring the non-overlapping n-grams over the total n-grams between a candidate sentence and previously selected sentences. Continuity is a measure of local coherence---there should be smooth transitions in the timeline---that is computed as the average n-gram overlap of all sentences in the current day with the previous summary. The final score is based on a weighted average of these metrics. To learn the relevance function, the authors leverage the same set of features from their previous work \cite{binh2013predicting}, but they also added an extra set of features about the event itself. For example, they evaluated whether the sentences properly represent the main event of the article, using the fact that the first sentences should contain the most relevant information (following the \textit{inverted pyramid structure}). Thus, they evaluate the similarity between the sentence summary and the first four sentences. Once the SVM-Rank method was trained, the ranking is fed to a dynamic programming approach to optimize the final score.

Huang and Huang \cite{huang2013optimized} present an event storyline generation method based on a mixture-event-aspect probabilistic model that can detect and distinguish the different types of sub-events in the article data set. Their model is an extension of Probabilistic Latent Semantic Analysis \cite{hofmann1999probabilistic} and LDA \cite{blei2003latent}. In particular, their model detects \textit{global} aspects (i.e., terms that are important throughout the whole story) and \textit{local} aspects (i.e., terms that are important in a specific event inside the story). Based on the extracted aspect model, the \textit{bursty} periods for each aspect are extracted to measure their popularity on a certain date and detect relevant events. Based on these results, it is possible to extract a timeline and select the most representative sentences associated with both global and local aspects to compose the final storyline with adjustable weights for the aspects. Sentences are selected by minimizing the overall \textit{information loss} over each aspect. In particular, the LexPageRank algorithm \cite{erkan2004lexpagerank} is used to rank sentences and KLD is used for sentence similarity.

Tran et al. \cite{tran2015timeline} propose a timeline summarization approach based on article headlines. Their approach is based on a random walk model using a topic-sensitive version of PageRank \cite{haveliwala2003topic} that selects relevant headlines from the data set for each time period. There are three key metrics to evaluate the relevance of a headline: \textit{informing value}, \textit{spread}, and \textit{influence}. The informing value depends on whether the headline provides factual information or an opinion, review, or another non-informing category. It is a binary value computed using a supervised learning approach based on an SVM classifier to separate facts from opinion \cite{yu2003towards}. Influence tries to measure the impact of an event on future events (e.g., ``the president resigns'' would lead to a ``new election'' event) based on references from future events using similarity between future news articles and the headline of the event. Spread is based on the intuitive idea that a relevant event will be reported in multiple news outlets---that is, its reporting will be spread over multiple headlines. Thus, it is a measure of \textit{positive} redundancy and it is formally defined as the probability of a headline being duplicated. To estimate whether two headlines are duplicates, the system uses a supervised logistic regression model trained on semantic similarity measures based on paraphrase detection literature \cite{mihalcea2006corpus}. Having defined these elements, the goal is to maximize all three aspects to select the best headlines. This is done by using PageRank on a graph of headlines, taking into account both spread (graph edges) and influence (random walk probability), to generate the final rankings. Next, to generate the final timeline for each day, the resulting rankings are selected greedily, subject to redundancy constraints, informativeness constraints, and a maximum number of headlines per day.

Chen et al. \cite{chen2015multi} present a supervised timeline summarization algorithm based on aging theory for news data sets. Aging theory \cite{chen2003life} is a model that tracks the life cycle of events using an energy function, which increases when an event becomes popular and diminishes with time. The method works by extracting sentences (i.e., specific events) and the publication time from news articles and using a classification model built with SVM to determine whether they belong in the output timeline. This approach is based on \textit{surface-level features} (e.g., noun frequencies and stop word frequencies), \textit{importance features} (e.g., latent semantic analysis scores), \textit{topic features} (e.g., topic word frequencies), an \textit{aging score feature} (i.e., changing coverage of an event over time), and a \textit{novelty feature}. The aging score is used to measure the life cycle of each term over time using a recurrence relation with TF-IDF representations. The novelty score is based on the Jaccard similarity of the current summary and the candidate sentence.

\subsubsection{Others}
Here we present works that use a sentence-level resolution but differ from the majority of the other works that follow the traditional timeline summarization approach. In particular, we consider works on extracting disaster storylines from news and works that present variations on the traditional TLS task.

\textbf{Disaster Storylines}
Zhou et al.'s works \cite{zhou2014generating,zhou2018new} present a framework to construct spatio-temporal storylines for disaster management from news data based on how the disaster location moves over time (e.g., a typhoon moving through different areas). This approach generates timelines for two levels of representation: a \textit{global level} that follows the progress of the disaster through each location and a \textit{local level} that focuses on a specific location. To extract the storyline, a series of snippets (i.e., event sentences) are extracted from the news articles using named entity recognition methods and grouped together based on a similarity graph. Then, a set of representative sentences is selected by finding the minimum dominating set \cite{shen2010multi} using a greedy algorithm. Next, an integer linear programming approach is used to select the optimal sequence for the main route of the disaster by maximizing the \textit{coherence} of the story chain, subject to a series of \textit{structural}, \textit{chronological}, and \textit{length} constraints. In this method, coherence is defined based on consecutive content similarity rather than word influence. However, the key difference is that this formulation includes a \textit{smoothness} constraint, which is specifically designed to track the moving location of disasters through time. Smoothness is based on simulating the natural trajectory of a disaster. In particular, the constraints set a maximum distance for consecutive events (i.e., avoiding jumps to locations too far away) and seek to avoid acute angles that could be formed by two consecutive connections (i.e., avoiding sharp turns in the trajectory of the disaster). Once the main storyline has been constructed, the next step is to analyze the local level storylines. For each main storyline event, a set of similar articles are selected and used to construct a multi-view graph that represents the event relationships based on content similarity. Then, a Steiner tree algorithm is used on the multi-view graph to generate a local storyline for that location.

Yuan et al. \cite{yuan2019dtexsl} propose dTexSL, a disaster storyline extraction approach that extends Zhou et al.'s works \cite{zhou2014generating,zhou2018new}. Unlike the previous approach, the news articles are first divided into different subsets based on location and are represented using neural embeddings. Locations are found by measuring the distance of the locations described in each article---using named entity recognition to find location references---and merging locations that are close enough based on a user-defined threshold. Then, an integer linear programming approach is used to select the \textit{key locations} (i.e., document clusters). Instead of choosing events to maximize coherence like before, the goal is to maximize the \textit{number of documents covered} on the map. The model has similar constraints as the original approach: \textit{chronological order}, \textit{length}, and \textit{smoothness}. Once the main storyline has been constructed, a word embedding method is used to construct a multi-view graph that represents the event relationships based on content similarity. Using this graph, a set of representative articles are selected based on two criteria: \textit{uniqueness}---computed using information gain---and \textit{relevance}---computed using a measure of node importance. Then, a dynamic Steiner tree algorithm is used on the multi-view graph to generate a local storyline for that specific location. Finally, a traditional multi-document summarization method  \cite{goldstein2000creating} is applied to generate a high-level event description for that specific location.

\textbf{Task Variations}
Duan et al. \cite{duan2020comparative} introduce another variation on the timeline summarization task called comparative timeline summarization. In this task, the goal is to provide timelines consisting of major contrasting events from two data sets. Their approach is based on three core characteristics: \textit{coverage}, \textit{distinctness}, and \textit{diversity}. Coverage is based on the idea that the timelines should cover most of the important information or topics from each data set. Distinctness is based on the idea that the events in a timeline should be distinct from the events on the other timeline at each time point, to allow for a proper contrast between them. Diversity is based on the idea that each timeline should cover a diverse set of events from its data set. To model these attributes, the authors propose a dynamic Markov model that is built around sentence similarity at a document level for each time step. In particular, sentences are selected from news articles to describe events based on local and global importance measures through the use of an affinity-preserving mutually reinforced Markov random walk model based on the PageRank algorithm. The output is a timeline that contains contrasting events from both data sets.

Yu et al. \cite{yu2021multi} propose a variation on the basic timeline summarization task, called Multi-TimeLine Summarization (MTLS). In this task, events are represented as sets of sentences and computationally represented by the neural embedding model sentence-BERT \cite{reimers2019sentence}. Given a set of time-stamped news articles, MLTS seeks to automatically extract timelines for important and different stories found in the data set. The authors propose a framework to solve this task called 2SAPS (Two-Stage Affinity Propagation Summarization). There are two key components in their framework: an event generation module and a timeline generation module. The event generation module seeks to extract important events from the document collection. To do so, it uses an \textit{affinity} propagation approach to cluster similar sentences \cite{frey2007clustering} and to identify the event of the article and any other previously \textit{referenced} event. Furthermore, there is a \textit{temporal similarity term} that uses an exponential decay function to penalize similarities of events that are temporally far away. Once the events are identified, a subset of these events is selected based on a weighted average of a \textit{salience metric}---based on event frequency---and a \textit{consistency metric}---based on the intra-event similarity. Next, the timeline generation module has three internal steps: event link, time selection, and timeline summarization itself. Event linking is based on the weighted average between a \textit{co-reference score} (based on entities or terms shared between events) and \textit{semantic similarity} (e.g., cosine similarity). Based on these average scores, the system builds an event graph and uses affinity propagation on it to determine the initial clusters (i.e., timeline sets). Next, there is a timeline selection based on the weighted average of \textit{timeline salience}---the average event salience of the timeline---and \textit{timeline coherence}---the average semantic similarity scores between chronologically adjacent events. The timeline summarizing step selects an exemplar sentence for each event in the timelines, as the most typical and representative member of each event. Finally, there is an add-on timeline tagging step which assigns a label to each timeline, based on the most frequent words of the events. 

Summarize Dates First (SDF) \cite{la2021summarize} is a timeline summarization pipeline that follows a different paradigm for timeline summarization based on generating a summary for each individual date first, and then selecting the most relevant dates using these summaries. This is different from the traditional approach where the relevant dates are selected first. Furthermore, this approach aggregates dates by leveraging higher-level temporal references (i.e., references to previous events in the article). SDF consists of three steps: temporal tagging, per-date summary extraction, and summary-drive date selection. In the temporal tagging stage, the raw text is annotated to identify date-level references (e.g., 31 December 2021) and high-level references (e.g., last December). The per-date summary extraction step uses any traditional sentence-based summarization algorithm from the multi-document summarization literature (e.g., TextRank \cite{mihalcea2004textrank}). Summary-driven date selection is the last step and uses a selection strategy, called Graph-Based Date Selection, which uses graph ranking algorithms (e.g., PageRank, HITS). In particular, a directed date graph model is built using the temporal references of the data set, where the edge weight connecting two dates is influenced by the count of date-level references and the similarity between the date summary and the high-level references to the earlier date.

\subsection{Events as Documents}
Here we present works that use a document-level resolution. We split the discussion into two parts: methods that build upon the Connect the Dots approach by Shahaf and Guestrin \cite{shahaf2010connecting}---a seminal work in the field of news narrative extraction---and others. We further divide the presentation based on whether the methods are linear or graph-based. We note that the works cataloged as others did not have a discernible pattern beyond using a document-level resolution.

\subsubsection{Connect the Dots Approaches}
\textbf{Linear Representations}
Shahaf and Guestrin \cite{shahaf2010connecting} proposed the Connect the Dots algorithm to extract temporal chains of documents (i.e., timelines). In particular, they use an optimization approach that seeks to maximize the overall \textit{coherence} of the timeline. Coherence measures the smoothness of a storyline, a coherent story should not have drastic changes in content or topic. To implement this metric, they propose an approach based on \textit{word influence}---a measure of word relevance computed through random walks on a word-document graph---and \textit{word activations}---which measure whether a specific word is active at a given point in the storyline. To extract the story chains, they used linear programming to maximize coherence subject to structural and temporal constraints. However, since linear programming provides non-integer solutions, it required additional heuristics to find the best chain by defining a rounding method. The linear programming approach used in the original Connect the Dots implementation was computationally expensive. Thus, Shahaf and Guestrin \cite{shahaf2012connecting} proposed a new method to reduce computational costs and avoid the approximate solutions from the linear program. In particular, they used a best-first search algorithm based on an extension heuristic---given a chain of documents, adding a new document to the chain will at most keep the same level of coherence---and the original linear program to individually evaluate each chain. 

Expanding upon the Connect the Dots method, Zhu and Oates \cite{zhu2012finding} propose an algorithm to extract story chains from newswire articles that connect two user-defined endpoints based on the following characteristics: \textit{relevance} (the articles on the chain should be relevant to the endpoints), \textit{coherence} (the transition between events should be smooth), low \textit{redundancy} (there should only be one representative article for every event of the chain), and \textit{coverage} (the chain should cover every important event). To compute measures of these characteristics, the article proposes using random walks on bipartite graphs formed by articles and words, where the weights are given by TF-IDF representations. Thus, based on these criteria, their proposed algorithm consists of two iterative stages. The first phase consists of a divide-and-conquer bisecting search problem that adds articles to the story chain. In particular, in this phase the algorithm finds the best article to insert in the middle of each current link of the story chain (i.e., it bisects the current links) based on coherence and relevance criteria. The second phase consists of pruning redundant articles---by removing a certain percentage based on how much coherence they would add to the current story chain---and irrelevant articles---by removing events that are similar to each other and temporally close with an exponential decay function. These phases are repeated until there are no more articles to add or prune. A subsequent article by the same authors \cite{zhu2014finding} revisits the story chain algorithm and extends this approach by adding an intermediate clustering step that groups documents into document clusters and words into word clusters. These clusters are used to generate a new bipartite \textit{correlation graph} that combines the weight of individual documents and words through a weighted average to assign the edge weights. Furthermore, the model adds a named \textit{entity bias} that assigns a higher weight to named entities compared to other terms. This is modeled through a co-occurrence frequency matrix for entity pairs, which is then used to compute a relevance score for each document in the data set based on the named entities. In turn, these elements are used to modify the cluster and document weights in the correlation graph.

Camacho Barranco et al. \cite{camacho2019analyzing} propose a storyline extraction algorithm that takes a set of user-defined articles as a seed and generates a timeline of articles based on a series of evaluation metrics. First, the authors propose a temporal criterion to filter candidate documents based on a range between the latest publication date of the seed articles and a maximum threshold away from the earliest publication date of the seed articles (i.e., in the interval $[t_{\min} - t_{threshold}, t_{\max}]$). Next, there is a topical criterion that measures how much a candidate article can deviate from the seed articles based on KLD and LDA topics. Having defined their basic framework, the authors then formalize an optimization problem to extract the storylines by selecting article connections based on different criteria: \textit{incoherence}, \textit{similarity}, \textit{overlap}, and \textit{uniformity}. Incoherence is based on the average pairwise Soergel distance between documents---measured using TF-IDF information for the entities of the document---with a temporal factor to penalize temporally distant articles. Similarity is used as a penalty factor to enforce diversity in non-adjacent articles of the storyline, implemented as a negative exponential factor based on the Soergel distance. Both of these metrics are weighted by a relevance factor of the documents and are smoothed using modified Gaussian distributions to measure event overlap. Next, an overall overlap factor for the storyline is computed, assigning a penalty based on the difference between publication dates and a user-defined threshold. The overlap factor ensures that the breakpoints occur at sufficiently distinct dates. The uniformity penalty seeks to avoid the case where the optimal solution selects purely irrelevant events as optimal by penalizing uniform weights. The objective function to minimize consists of the sum of the product between incoherence and similarity, multiplied by the overlap and uniformity penalties.

\textbf{Graph-based Representations}
Metro Maps \cite{shahaf2012trains,shahaf2013metro} are an extension of the Connect the Dots approach that represents more than a single storyline using a directed acyclic graph of events. In particular, the metro maps method is a structured summarization approach that captures the evolution of multiple stories and their interactions. The stories are represented using a metro map metaphor, where each metro line represents a story and stations represent key events. Metro lines intersect in specific stations, representing how storylines connect with each other. This representation is extracted by solving an optimization problem. In particular, the goal is to maximize \textit{connectivity}, subject to \textit{coverage} and \textit{coherence} constraints. Coverage is computed based on how well specific terms or keywords are represented in the selected events and is defined using a \textit{submodular function} that encourages diversity (e.g., if a term is already covered, adding a document that covers it provides little extra coverage). These keywords depend on the specific corpus or domain of application. Coherence is defined following Shahaf et al.'s previous work \cite{shahaf2010connecting,shahaf2012connecting}. Finally, connectivity is defined as the number of stories that intersect which is used to ensure that the final metro map is connected. The optimization problem is solved in phases. First, a series of coherent candidate metro lines are selected based on a divide-and-conquer approach, which constructs long lines from shorter ones and encodes them in a graph. Then, the method extracts a set of coherent lines that maximize coverage using an approximation algorithm based on the submodularity of the coverage function (otherwise finding these lines is an NP-hard problem). Finally, connectivity is increased using a local search approach that substitutes lines without sacrificing coverage.

Similar to the metro maps metaphor, the Narrative Maps model \cite{keith2021narrative} provides a framework to extract and represent narratives based on a route map metaphor. The narrative and its stories are shown as a series of routes through landmarks, which represent the events. In computational terms, the narrative is modeled through a directed acyclic graph of events. The events are represented through neural embeddings of article headlines. The graph is extracted by solving an optimization problem defined following a linear programming formulation similar to the Connect the Dots approach. The optimization problem is based on maximizing \textit{coherence} subject to \textit{coverage} constraints. Coherence measures how much sense it makes to connect two events together and is defined as the geometric mean of the \textit{content similarity} of events---using cosine or angular similarity---and their \textit{topical similarity}---based on JS similarity of their topic distributions based on clustering. Coverage is measured by the average percentage of topical clusters covered by the selected events based on their topic distributions. Once the optimal map has been found, the main storyline is extracted by normalizing the coherence values of the edges into probabilities and finding the maximum likelihood path. Then, a set of representative landmarks (i.e., important events) of each story by finding the maximum antichain, which corresponds to the point of the maximum width of the graph.

\subsubsection{Others}
\textbf{Linear Representations}
Guha et al. \cite{guha2005unweaving} propose an \textit{event threading} approach based on a graph decomposition method that generates document timelines. In particular, they propose decomposing a directed acyclic graph into disjointed node paths that ensure that as many nodes as possible participate in at least one path (i.e., they seek to maximize a notion of \textit{coverage}). The first step is to construct the graph, they propose doing this based on \textit{important terms} (or even entities) in the document collection and their co-occurrence. Furthermore, documents are modeled following a bag of words approach, although the method is also designed to handle TF-IDF representations. Once the graph is constructed, the next step is to solve the event thread extraction problem. To do this, they propose three formulations: an exact algorithm, a maximum approach, and a dynamic programming approach. The first method is an exact algorithm based on minimum cost flow, which has a high computational cost and is impractical. The second is an approximation algorithm based on maximum matching in bipartite graphs that solves the thread extraction problem for a fixed maximum size. The third method is based on an approximation algorithm that uses dynamic programming to solve the thread extraction problem for a range of thread sizes.

Laban and Hearst \cite{laban2017newslens} present newsLens, a system to build and visualize long-ranging news stories. In particular, their system groups news articles based on their \textit{topics}---based on a graph clustering approach---and then selects a sample of headlines from salient dates---based on the \textit{frequency} of publications. In more detail, the first step in their extraction approach is to construct a keyword graph for a starting time period using TF-IDF representations of the articles. Next, a local topic graph is created based on a user-defined threshold for the number of shared keywords between articles. After the initial time period, a sliding window approach with a user-defined length is used to handle the rest of the data. For each time period, a local topic graph is created and compared with the graph from the previous period to check for three types of relationships: linking (connecting a topic from the current graph to a pre-existing topic), splitting (dividing a pre-existing topic into new topics in the current period), or merging (combining separate topics from the previous step into a single one of the current period). However, this approach is not able to handle stories that have \textit{long-time gaps} between publications. To handle these cases, the content similarity of non-overlapping stories is analyzed and merged if above a specific threshold. Afterward, their method assigns a name to the storyline by extracting noun phrases from the news articles and scoring them based on multiple criteria (e.g., length, type of noun, abstractness, and frequency). Finally, salient dates are selected based on local frequency changes, and representative headlines are sampled randomly from these dates to generate the final timeline visualization.

\textbf{Graph Representations}
Uramoto and Takeda \cite{uramoto1998method} proposed a graph-based approach to model the relationships between news articles. In particular, they use a directed graph based on temporal ordering and \textit{event similarity}. This is the earliest article that fits with our definitions of event-based narrative representations for news narratives that we found. In particular, the authors use the concepts of \textit{genus} and \textit{differentia} words. For adjacent articles, genus words are computed using the intersection of their word sets and represent already known information in the story. In contrast, differentia words are built from the set difference between the articles (in temporal order) and represent new knowledge in the story. Thus, differentia words are more important when trying to find coherent sequences of articles. The events are represented with a variation of TF-IDF that assigns more weight to \textit{differentia} words.

Tannier and Moriceau \cite{tannier2013building} propose an approach for building multi-document event threads from news articles. In particular, they use a supervised learning approach with a series of classifiers to define the type of relationship between news articles: \textit{same-event}, \textit{continuation}, or \textit{reaction}. The output of this method is a temporal event graph, where the nodes correspond to events (represented as news articles) and the edges are labeled with the corresponding relationships. In particular, the first step is to determine whether there is a connection at all between the articles. To do so, an initial classifier is implemented using a series of \textit{content similarity} features (e.g., word overlap, cosine similarity, and similarity of the first sentences) to construct the initial temporal graph. However, this is not enough to find all potential relationships and a second-level classifier is included that takes into account the results from the previous classifier by using \textit{degree-based features} from the temporal graph. Next, after a connection has been established, another classifier determines whether this connection is based on the articles referring to the same news event---same-event connection---or based on a continuation---when an event is a direct continuation or consequence of a previous one. This classifier relies on \textit{date-based features} (e.g., differences in publication time, date references, and references between events themselves) and \textit{keyword-based features} (e.g., usage of temporal words, reaction words, or opinion words). The output is fed into another classifier that leverages degree-based features again to find more relationships. Due to the \textit{transitive} nature of the same-event and continuation relationships, a post-processing step takes the graph and constructs the \textit{transitive closure} for these specific relations. Afterward, a final classifier uses the same features to determine whether a continuation is a reaction---a subset of continuations that relate the reactions of people (or organizations) to an event. 

Hu et al. \cite{hu2014exploring} propose a system to model storyline interactions from news events. Their approach generates a series of event timelines focusing on specific entities or topics and their interactions with each other. In particular, this results in a directed graph connecting multiple events. In contrast to other approaches, the underlying representation of events is based on the \textit{main event descriptors} (i.e., the answers to Who, What, When, Where, Why, and How) \cite{norambuenaevaluating} which are extracted directly from each article and represent the key elements of the event. Based on this information, a coherence graph is constructed and used to identify the storylines through a random walk. Coherence is defined by three factors: \textit{subtopic consistency}, \textit{entity relatedness}, and \textit{time continuity}. To measure subtopic consistency, the first step is to use a generative probabilistic mixture model to discover latent subtopics. Then, JS divergence is used to measure the distance of topic distributions between articles. Next, entity relatedness is measured by the average \textit{affinity} of the entities from each pair of articles using normalized point-wise mutual information. The time continuity factor is simply defined as an exponential penalty term dependent on the temporal distance between events. The coherence graph is built by creating edges between documents that have a coherence score above a given threshold. Based on the coherence graph, a series of \textit{informative events} that connect multiple storylines are identified. Specifically, a topic-sensitive PageRank algorithm \cite{haveliwala2003topic} is used to discover these events. In turn, these events feed the storyline generation algorithm, an iterative algorithm that selects a single informative event for each story for each day. 

Bögel and Gertz \cite{bogel2015time} present a temporal linking framework based on the concept of article references. In particular, they exploit the structure of news articles to construct an information network. Instead of comparing articles based on overall content similarity, they exploit the use of \textit{lead paragraphs}, \textit{explanatory paragraphs}, and \textit{additional information paragraphs} in typical news articles. Specifically, they construct the network based on \textit{temporal expressions}, \textit{keywords}, and \textit{entity names}. To select valid event connections, the first step is to filter based on temporal information contained in the text based on a temporal tagger. Next, connections are evaluated based on the \textit{similarity} of the lead paragraph of a news article with all the other paragraphs of another news article (i.e., capturing references to the event). Similarity is computed based on the entities and keywords mentioned in each paragraph based on a weighted average of Jaccard and cosine similarity. Finally, irrelevant edges are pruned based on a user-defined threshold. However, some non-relevant edges are kept if they fulfill the role of a \textit{support path}---paths that have non-relevant edges but share endpoints with fully relevant paths---that provide more evidence of two events being connected. The output is a directed graph based on references, not necessarily acyclic, as there are future temporal references in some articles.

\subsection{Events as Clusters: Event Evolution and Threading}
Now, we present works that use a cluster-level resolution. We divide the discussion into two parts: works related to event threading \cite{nallapati2004event} and evolution \cite{yang2006tracing}, and others. 

\subsubsection{Event Threading and Evolution}
Nallapati et al. \cite{nallapati2004event} use a directed graph model to represent to capture the structure and dependencies of events in a news topic. They call this extraction process \textit{event threading}. They represent each event as a cluster of news articles. Event threading is a supervised method that consists of two phases: clustering documents and modeling dependencies. The clustering process starts with a cluster for each document in the data set and merges them iteratively based on similarity until the similarities fall below a predefined threshold. The authors evaluate three types of cluster similarity on the average link, complete link, or single link of the clusters based on document similarities. Document similarities are based on \textit{content similarity} (e.g., cosine similarity), \textit{common locations}, and \textit{common entities}. Furthermore, there is an exponential decay term based on the \textit{temporal distance} to penalize larger temporal distances between documents. Next, dependency modeling uses surface-level features of the document clusters, such as word distributions and time-ordering of the news articles. Based on this information, the authors propose several link extraction criteria (complete-link, simple threshold, nearest parent, best similarity, and maximum spanning tree). These approaches rely on temporal order, similarity information, or structural information. 

SToRe (Storyline-based Topic Retrospection) is a topic retrospective system \cite{lin2006topic, lin2007individualized, lin2008storyline} that extracts the main storyline from a given news topic and provides a summary of the topic based on this storyline. In particular, the extraction process consists of four phases: event identification, topic structure identification, main storyline construction, and storyline-based summarization. In the event identification phase, similar news articles will be clustered together to represent a single event using self-organizing maps. In the topic structure identification step, the events are linked together based on whether their \textit{similarity} exceeds a specific threshold. To compute similarity, the events are represented with a vector of term weights using the concepts of \textit{genus} and \textit{differentia} words \cite{uramoto1998method}. Then, cosine similarity is used to compare the event vectors. Next, in the main storyline construction step, an MST is extracted from the constructed topic structure. The MST is based on the relevance of each event with respect to the topic. The MST is used to generate a timeline of events, and it is further extended with small side branches of other relevant events based on a specific threshold. Finally, in the storyline-based summarization, a summary is generated for each event based on the news articles contained in its cluster using accumulated weight summary \cite{goldstein2000creating}.

Yang et al. \cite{yang2006tracing, yang2009discovering} use directed acyclic graphs to represent the evolution of events in online news. They call their approach \textit{event evolution graphs}, which represent temporal and causal relationships between events. Events are defined as sets of news articles and are represented as the average of the TF-IDF vectors of each article they contain. We note that the proposed method assumes that events and their corresponding articles are already computed. In practice, this would require a clustering step before constructing the graph. These events are linked together based on their similarity and a user-specified threshold, which is computed based on \textit{content similarity} (e.g., cosine similarity), \textit{temporal proximity}, and \textit{document distributional proximity} (which penalizes bursty periods with many articles about the same event). The latter two terms are represented through exponential decay factors. Furthermore, users are able to reduce the temporal granularity of the event evolution graph, which merges specific events that occur in short time frames. 

Qiu et al. \cite{qiu2008timeline} propose another event evolution graph extraction method. Their construction method follows an iterative approach based on \textit{content similarity} and \textit{temporal order}. In particular, documents are first grouped into clusters using the OHC method \cite{qiu2007topic} in the first time period, which gives rise to the initial events. Next, the PRAC method \cite{qiu2007novel} is used to build classifiers and determine whether the documents of the next time period are continuations of a cluster identified in the previous period. If so, a new event node is created using the identified cluster as its parent. This process is repeated until the last time period. Next, \textit{twigs}---paths that die before the end of the timeline---are removed based on a user-set tolerance, and equivalent event nodes are merged to reduce graph complexity.

TSCAN (Topic Summarization and Content ANatomy) \cite{chen2008tscan, chen2010tscan} is a method to analyze news data that produces a global summary and constructs an event evolution graph. We focus on the event graph component of this method. First, news articles are grouped into themes obtained through a matrix factorization approach with TF-IDF document representations. Next, the news articles of each theme are temporally segmented using an \textit{energy value} threshold based on eigenvalues from the matrix representation. In practice, this generates clusters of documents based on \textit{frequency}, which are associated with the nodes of the event evolution graph. The evolution graph is a directed acyclic graph, where the edges are constructed using \textit{temporal similarity}---computed using the temporal distance between events, with special cases to consider event overlap---and \textit{content similarity}---based on cosine similarity.

Khurdiya et al. \cite{khurdiya2011multi} propose a system that extracts directed graphs to represent stories from news data using multi-perspective links. Each node of this graph is associated with multiple news articles. The system uses LDA to extract topics in each time unit (e.g., a day). The extracted topics are associated with sets of articles based on the strength of the topic in each article and form the basis of the story identification model. We note that these topics and their article sets correspond to the notion of event that we use in this survey. Next, article sets are linked chronologically based on \textit{topic correlation} (e.g., Pearson's correlation coefficient) and a user-defined threshold, generating a directed graph of events. 

Wei et al. \cite{wei2014exploiting} identify event episodes in news data sets and construct a temporal episode graph (i.e., an event graph under our definitions in the survey). In particular, this article shows a discovery mechanism that organizes news documents into events using novel TF-IDF representations that incorporate a temporal component. Then, the system builds a link structure based on intercluster similarity measures. The first proposed event representation, called TF-IDF\textsubscript{Tempo}, gives more weight to features with \textit{consecutive occurrences} in a sequence of documents (i.e., it incorporates the surrounding context of the document) by modifying the IDF component of TF-IDF to consider the order of the documents. However, this approach is too strict and is unable to model overlapping events. Moreover, it also has a high bias towards low-frequency articles that are temporally close. Thus, the authors propose a second representation, called TF-Enhanced-IDF\textsubscript{Tempo} which modifies the IDF component by adopting the \textit{significance factor} proposed by Luhn \cite{luhn1958automatic} and a \textit{temporal gap threshold} to allow for short discontinuities in feature appearances. These representations are used with Hierarchical Agglomerative Clustering (HAC) \cite{voorhees1986implementing} to construct the article clusters that represent the events. For the purposes of clustering, document similarity is defined by content similarity (e.g., cosine similarity) and a negative exponential penalty for temporally distant documents.

Huang et al. \cite{huang2014discovering} propose a different event evolution approach to build and analyze event relationships based on three types of event connections. In particular, they define a \textit{co-occurrence dependence relationship}, an \textit{event reference relationship}, and a \textit{temporal proximity relationship}. The authors define events as a set of news articles and identify them through clustering and topic modeling using a combined similarity measure that leverages LDA and a TF-IDF document model with cosine similarity. Once the events are identified, the method extracts a series of core features (i.e., key entities and terms of the article) by analyzing the \textit{lead} of the articles and evaluating whether their frequency is above a specified threshold. These core features are used to construct a vectorial representation of the events. For the co-occurrence relationship, the method computes the aggregation of all mutual information between all features of the event, generating a symmetric matrix that represents all event-event relationships. For the event reference analysis, the method identifies shared core features and defines the degree of event reference based on the frequency of references in an event to the core features of a previous event, adjusted by the weight of these terms in the referencing event. Temporal dependency is evaluated using an exponential decay formula.

Event Phase Oriented News Summarization (EPONS) \cite{wang2018event} is a timeline summarization approach that assumes that a story summary contains multiple timelines, each one corresponding to a specific event. To model the semantic relations of news articles, EPONS uses a graph model, called Temporal Content Coherence Graph (TCCG), which is an event graph based on two metrics: \textit{content coherence} and \textit{temporal influence}. Content coherence is based on the weighted average of \textit{topic level similarity}---modeled by JS divergence over an LDA topic distribution---and \textit{entity-level similarity}---modeled over a ranking of named entities using the Tanimoto coefficient. Temporal influence is modeled through a Hamming (cosine) kernel to properly separate temporally distinct events. The TCCG is built by selecting edges that are above user-specified thresholds in each metric. Based on this graph, EPONS uses a modified structural clustering approach to group the news articles into different events. Furthermore, small clusters of similar articles are filtered out to ensure that the events are modeled properly. This post-processing is done by using four quality metrics on a pre-trained logistic regression classifier: percentage of new articles, time interval length, pairwise topic similarity, and pairwise entity similarity. Having identified the events, it is now necessary to construct the individual summaries and finalize the timeline. To do so, a vertex-reinforced random walk \cite{mei2010divrank,pemantle1992vertex} is used to rank the relevance of news articles inside each event, in a similar manner to PageRank. Next, a supervised model is used to determine whether the headlines are factual (i.e., they are reporting a specific event) or an opinion, as opinion-based headlines are not considered useful for timelines and must be filtered out. Finally, an optimization method is used to maximize the total \textit{relevance}, subject to \textit{non-redundancy} constraints (i.e., disallowing events that are too similar) to select the news articles. 

Cai et al. \cite{cai2019temporal} propose a method to extract Temporal Event Maps (TEM) based on the \textit{content dependence} degree and \textit{component event reference} degree for each pair of events. TEMs are directed graphs that have events as nodes, relations as edges, edge weights representing the strength of event relationships, and node weights representing the importance of each event. Events are defined as groups of related documents and identified using a LDA model. After obtaining the events, the next step is to compute the two core metrics that define the temporal event maps. The content dependence degree is defined as the aggregation of all mutual information among the features of each event. The content reference degree is defined by the presence of \textit{core features} of an event---salient terms based on frequency---in other events. Unlike content dependence, this is not a symmetric relationship between events. To construct the temporal event maps, the first step is to order events based on starting time. Then, connections are added for events that surpass a user-specified threshold for the product of content dependence and event reference degrees, which provides the edge weights for the graph. Finally, a ranking procedure based on PageRank is used to generate the event importance values.

\subsubsection{Others}
\textbf{Information Cartography}
Continuing with their work on metro maps, Shahaf et al. \cite{shahaf2013information,shahaf2015information} propose a new framework called Information Cartography that features \textit{zoomable} metro maps, allowing users of the map to visualize the news at different levels of resolution, allowing the user to zoom in to specific metro stops and generate a new map. Metro stops and events are no longer represented as single documents but as clusters of events. The articles are segmented into time windows and clusters are computed using a community-detection algorithm on word co-occurrence graphs. To extract the maps, an optimization problem is defined based on finding the best \textit{structure} for the map, relying on the idea of minimizing the total number of storylines (to reduce unneeded complexity) and maximizing the \textit{number of covered clusters} (to ensure that the stories are well covered). This approach leads to simple stories being modeled as a single metro line and more complex stories requiring the use of multiple shorter lines. Furthermore, a series of additional constraints for \textit{story coherence}, \textit{cluster quality}, and \textit{map size} is imposed. 

Building upon the concept of metro maps and information cartography, Xu and Tang \cite{xu2018generating} propose a narrative representation in the context of societal risk events (e.g., earthquakes) called Risk Maps. These maps follow the same basic representation of information cartography with events being represented as clusters of documents. However, one key difference is that this approach leverages advances in text representation by using neural word embeddings for news articles before clustering. To obtain the risk map, the authors choose to maximize \textit{coverage} as their primary objective, followed by \textit{connectivity}, subject to a minimal \textit{coherence} constraint. Coverage is defined based on how well each cluster is covered by the different storylines. Connectivity is simply the number of storylines that intersect. Coherence is defined based on the Jaccard similarity of consecutive clusters in the storylines. The optimization problem is solved using a greedy algorithm that finds the best path among clusters at each step.

\textbf{Story Forests}
Liu et al. \cite{liu2017growing,liu2020story} propose the Story Forest approach, where different stories are constructed and represented as a forest of event trees. First, events are clustered using a community detection approach on word co-occurrence graphs using betweenness centrality. Next, documents are associated with each topic through a similarity based on TF-IDF representations. Afterward, a second step groups documents together based on a supervised classifier (SVM) to determine whether pairs of documents refer to the same event based on TF-IDF features and similarities between the contents and titles of articles. The story forest is built iteratively by adding events into its trees by using three operations: merge, extend and insert. Before adding the events it is necessary to determine the correct story tree. This is done based on a measure of \textit{compatibility}, computed as the Jaccard similarity of the keywords of the event and the tree. If no trees are related to the event, a new tree is created with the event as its root. To add the event to an existing tree, the method first tries to \textit{merge} it with any of the existing events into the same node using the previously trained SVM classifier. Otherwise, the method scans all the nodes to identify which tree to \textit{extend} based on a measure of connection strength determined by three elements: \textit{compatibility}, \textit{coherence}, and \textit{time penalty}. Compatibility is measured by the similarity of their centroids based on cosine similarity. Coherence is a story-level measure that takes into account the path of events from the root of the tree to the newly appended event by measuring the average consecutive compatibility value. Finally, the time penalty is an exponential decay factor that depends on temporal distance. If none of the events are appropriate, the event is \textit{inserted} as a new node connected to the root.

\section{Narrative Extraction Criteria}
In this section, we present a summary of the different construction criteria found in the reviewed articles. These criteria refer to either an evaluation metric or additional information used in the extraction algorithms themselves as part of an objective function (e.g., coherence optimization), selection criteria (e.g., filtering based on content similarity or topic distribution similarity), and other types of extraction heuristics (e.g., leveraging article structure to compute content similarity or evaluating the use of opinionated language). The first part of Table \ref{tab:metrics-criteria} provides an overview of the different construction criteria. We note that these criteria are not mutually exclusive and can be combined as needed.

\begin{table}[!htb]
\resizebox{0.8\textwidth}{!}{
\begin{tabular}{@{}ccccccccccccccccccccccccccc@{}}
\toprule
     &                                                   & \multicolumn{14}{c}{Extraction Criteria}                                                                                                                                                                                                                                            & \multicolumn{11}{c}{Evaluation Metrics}                                                                                                                                                                  \\ \midrule
Year & Reference                                         & \rotn{Relevance} & \rotn{Surface Similarity} & \rotn{Topic Distribution} & \rotn{Entities} & \rotn{Coherence} & \rotn{Coverage} & \rotn{Dispersion} & \rotn{Diversity and Redundancy} & \rotn{Output Structure} & \rotn{Article Structure} & \rotn{Content References} & \rotn{Temporal References} & \rotn{Burstiness and Frequency} & \rotn{Temporal Distance} & \rotn{Traditional IR Metrics} & \rotn{Summarization Metrics} & \rotn{Ranking Metrics} & \rotn{Clustering Metrics} & \rotn{Coherence} & \rotn{Coverage} & \rotn{Dispersion} & \rotn{Diversity and Redundancy} & \rotn{User Performance} & \rotn{User Perception} & \rotn{Other or N/A} \\ \midrule
1998 & \multicolumn{1}{c|}{\citet{uramoto1998method}}     &           & $\times$                  &                    &          &           &          &            &                          &                  &                   &                    &                     &                          & \multicolumn{1}{c|}{}                  &                        &                       &                 &                    &           &          &            &                          &                  &                 & $\times$            \\
2004 & \multicolumn{1}{c|}{\citet{nallapati2004event}}    &           & $\times$                  &                    & $\times$        &           &          &            &                          &                  &                   &                    &                     &                          & \multicolumn{1}{c|}{$\times$}                 & $\times$                      &                       &                 &                    &           &          &            &                          &                  &                 &              \\
2004 & \multicolumn{1}{c|}{\citet{chieu2004query}}        &           & $\times$                  &                    &          &           &          &            &                          &                  &                   &                    &                     & $\times$                        & \multicolumn{1}{c|}{$\times$}                 &                        &                       &                 &                    &           &          &            &                          &                  & $\times$               &              \\
2005 & \multicolumn{1}{c|}{\citet{guha2005unweaving}}     &           & $\times$                  &                    &          &           & $\times$        &            &                          &                  &                   &                    &                     &                          & \multicolumn{1}{c|}{}                  &                        &                       &                 &                    &           &          &            &                          &                  &                 & $\times$            \\
2006 & \multicolumn{1}{c|}{\citet{yang2006tracing}}       &           & $\times$                  &                    &          &           &          &            &                          &                  &                   &                    &                     & $\times$                        & \multicolumn{1}{c|}{$\times$}                 & $\times$                      &                       &                 &                    &           &          &            &                          &                  &                 &              \\
2006 & \multicolumn{1}{c|}{\citet{lin2006topic}}          &           & $\times$                  &                    &          &           &          &            &                          &                  &                   &                    &                     &                          & \multicolumn{1}{c|}{}                  &                        &                       &                 &                    &           &          &            &                          & $\times$                & $\times$               &              \\
2007 & \multicolumn{1}{c|}{\citet{lin2007individualized}} &           & $\times$                  &                    &          &           &          &            &                          &                  &                   &                    &                     &                          & \multicolumn{1}{c|}{}                  & $\times$                      &                       &                 &                    &           &          &            &                          &                  &                 &              \\
2008 & \multicolumn{1}{c|}{\citet{chen2008tscan}}         &           & $\times$                  &                    &          &           &          &            &                          &                  &                   &                    &                     & $\times$                        & \multicolumn{1}{c|}{$\times$}                 &                        & $\times$                     &                 &                    &           &          &            &                          &                  &                 &              \\
2008 & \multicolumn{1}{c|}{\citet{qiu2008timeline}}       &           & $\times$                  &                    &          &           &          &            &                          &                  &                   &                    &                     &                          & \multicolumn{1}{c|}{}                  &                        &                       &                 &                    &           &          &            &                          &                  &                 & $\times$            \\
2008 & \multicolumn{1}{c|}{\citet{lin2008storyline}}      &           & $\times$                  &                    &          &           &          &            &                          &                  &                   &                    &                     &                          & \multicolumn{1}{c|}{}                  &                        &                       &                 &                    &           &          &            &                          & $\times$                & $\times$               &              \\
2009 & \multicolumn{1}{c|}{\citet{yang2009discovering}}   &           & $\times$                  &                    &          &           &          &            &                          &                  &                   &                    &                     & $\times$                        & \multicolumn{1}{c|}{$\times$}                 & $\times$                      &                       &                 &                    &           &          &            &                          &                  &                 &              \\
2010 & \multicolumn{1}{c|}{\citet{shahaf2010connecting}}  &           &                    &                    &          & $\times$         &          &            &                          &                  &                   &                    &                     &                          & \multicolumn{1}{c|}{}                  &                        &                       &                 &                    &           &          &            &                          &                  & $\times$               &              \\
2011 & \multicolumn{1}{c|}{\citet{yan2011evolutionary}}   & $\times$         &                    &                    &          & $\times$         & $\times$        &            & $\times$                        &                  &                   &                    &                     &                          & \multicolumn{1}{c|}{}                  &                        & $\times$                     &                 &                    &           &          &            &                          &                  &                 &              \\
2011 & \multicolumn{1}{c|}{\citet{yan2011timeline}}       & $\times$         &                    &                    &          &           &          &            &                          &                  &                   &                    &                     & $\times$                        & \multicolumn{1}{c|}{$\times$}                 &                        & $\times$                     &                 &                    &           &          &            &                          &                  &                 &              \\
2011 & \multicolumn{1}{c|}{\citet{hu2011generating}}      &           & $\times$                  & $\times$                  & $\times$        &           &          &            &                          &                  &                   &                    &                     &                          & \multicolumn{1}{c|}{}                  & $\times$                      &                       &                 &                    &           &          &            &                          &                  &                 &              \\
2011 & \multicolumn{1}{c|}{\citet{khurdiya2011multi}}     &           &                    & $\times$                  &          &           &          &            &                          &                  &                   &                    &                     &                          & \multicolumn{1}{c|}{}                  & $\times$                      &                       &                 &                    &           &          &            &                          &                  &                 &              \\
2012 & \multicolumn{1}{c|}{\citet{zhu2012finding}}        & $\times$         & $\times$                  &                    &          & $\times$         &         &            & $\times$                        &                  &                   &                    &                     &                          & \multicolumn{1}{c|}{}                  &                        &                       &                 &                    &           &          &            &                          &                  & $\times$               &              \\
2012 & \multicolumn{1}{c|}{\citet{chen2010tscan}}         &           & $\times$                  &                    &          &           &          &            &                          &                  &                   &                    &                     & $\times$                        & \multicolumn{1}{c|}{$\times$}                 &                        & $\times$                     &                 &                    &           &          &            &                          &                  &                 &              \\
2012 & \multicolumn{1}{c|}{\citet{shahaf2012connecting}}  &           &                    &                    &          & $\times$         &          &            &                          &                  &                   &                    &                     &                          & \multicolumn{1}{c|}{}                  &                        &                       &                 &                    &           &          &            &                          &                  & $\times$               &              \\
2012 & \multicolumn{1}{c|}{\citet{shahaf2012trains}}      &           &                    &                    &          & $\times$         & $\times$        &            &                          & $\times$                &                   &                    &                     &                          & \multicolumn{1}{c|}{}                  &                        &                       &                 &                    &           &          &            &                          & $\times$                &                 &              \\
2013 & \multicolumn{1}{c|}{\citet{binh2013predicting}}    & $\times$         & $\times$                  & $\times$                  &          & $\times$         &          &            &                          &                  & $\times$                 &                    & $\times$                   & $\times$                        & \multicolumn{1}{c|}{}                  & $\times$                      & $\times$                     &                 &                    &           &          &            &                          &                  &                 &              \\
2013 & \multicolumn{1}{c|}{\citet{li2013evolutionary}}    & $\times$         &                    & $\times$                  &          & $\times$         & $\times$        &            &                          &                  &                   &                    &                     &                          & \multicolumn{1}{c|}{}                  &                        & $\times$                     &                 &                    &           &          &            &                          &                  &                 &              \\
2013 & \multicolumn{1}{c|}{\citet{tran2013leveraging}}    & $\times$         & $\times$                  & $\times$                  &          & $\times$         &          &            &                          &                  & $\times$                 &                    & $\times$                   & $\times$                        & \multicolumn{1}{c|}{}                  &                        & $\times$                     &                 &                    &           &          &            &                          &                  &                 &              \\
2013 & \multicolumn{1}{c|}{\citet{huang2013optimized}}    &           &                    & $\times$                  &          &           &          &            &                          &                  &                   &                    &                     & $\times$                        & \multicolumn{1}{c|}{}                  &                        & $\times$                     &                 &                    &           &          &            &                          &                  &                 &              \\
2013 & \multicolumn{1}{c|}{\citet{tannier2013building}}   &           & $\times$                  &                    &          &           &          &            &                          & $\times$                &                   &                    & $\times$                   &                          & \multicolumn{1}{c|}{}                  & $\times$                      &                       &                 &                    &           &          &            &                          &                  &                 &              \\
2013 & \multicolumn{1}{c|}{\citet{shahaf2013metro}}       &           &                    &                    &          & $\times$         & $\times$        &            &                          & $\times$                &                   &                    &                     &                          & \multicolumn{1}{c|}{}                  &                        &                       &                 &                    &           &          &            &                          & $\times$                &                 &              \\
2013 & \multicolumn{1}{c|}{\citet{shahaf2013information}} &           &                    &                    &          & $\times$         & $\times$        &            &                          & $\times$                &                   &                    &                     &                          & \multicolumn{1}{c|}{}                  &                        &                       &                 &                    &           &          &            &                          & $\times$                &                 &              \\
2014 & \multicolumn{1}{c|}{\citet{nguyen2014ranking}}     & $\times$         & $\times$                  & $\times$                  &          &           &          &            & $\times$                        &                  &                   &                    &                     & $\times$                        & \multicolumn{1}{c|}{}                  & $\times$                      & $\times$                     &                 &                    &           &          &            &                          &                  &                 &              \\
2014 & \multicolumn{1}{c|}{\citet{zhu2014finding}}        & $\times$         & $\times$                  &                    & $\times$        & $\times$         &       &            & $\times$                        &                  &                   &                    &                     &                          & \multicolumn{1}{c|}{}                  &                        &                       &                 &                    &           &          &            &                          &                  & $\times$               &              \\
2014 & \multicolumn{1}{c|}{\citet{huang2014discovering}}  &           & $\times$                  & $\times$                  &          &           &          &            &                          &                  &                   & $\times$                  &                     &                          & \multicolumn{1}{c|}{$\times$}                 & $\times$                      &                       &                 &                    &           &          &            &                          &                  &                 &              \\
2014 & \multicolumn{1}{c|}{\citet{wei2014exploiting}}     &           & $\times$                  &                    &          &           &          &            &                          &                  &                   &                    &                     & $\times$                        & \multicolumn{1}{c|}{$\times$}                 & $\times$                      &                       &                 &                    &           &          &            &                          &                  &                 &              \\
2014 & \multicolumn{1}{c|}{\citet{hu2014exploring}}       &           &                    & $\times$                  & $\times$        &           &          &            &                          &                  &                   &                    &                     &                          & \multicolumn{1}{c|}{$\times$}                 & $\times$                      & $\times$                     &                 &                    &           &          &            &                          &                  & $\times$               &              \\
2014 & \multicolumn{1}{c|}{\citet{zhou2014generating}}    &           & $\times$                  &                    &          & $\times$         &          &            &                          & $\times$                &                   &                    &                     &                          & \multicolumn{1}{c|}{}                  &                        & $\times$                     &                 &                    &           &          &            &                          &                  &                 &              \\
2015 & \multicolumn{1}{c|}{\citet{tran2015timeline}}      & $\times$         &                    &                    &          &           &          &            & $\times$                        &                  &                   &                    & $\times$                   & $\times$                        & \multicolumn{1}{c|}{}                  & $\times$                      &                       &                 &                    &           &          &            &                          &                  & $\times$               &              \\
2015 & \multicolumn{1}{c|}{\citet{li2015tracking}}        & $\times$         & $\times$                  & $\times$                  &          & $\times$         & $\times$        &            &                          &                  &                   &                    &                     &                          & \multicolumn{1}{c|}{}                  &                        & $\times$                     &                 &                    &           &          &            &                          &                  &                 &              \\
2015 & \multicolumn{1}{c|}{\citet{bogel2015time}}         &           & $\times$                  &                    & $\times$        &           &          &            &                          &                  & $\times$                 &                    & $\times$                   &                          & \multicolumn{1}{c|}{}                  & $\times$                      &                       &                 &                    &           & $\times$        &            &                          &                  &                 &              \\
2015 & \multicolumn{1}{c|}{\citet{chen2015multi}}         &           & $\times$                  & $\times$                  &          &           &          &            &                          &                  &                   &                    &                     &                          & \multicolumn{1}{c|}{$\times$}                 & $\times$                      & $\times$                     &                 &                    &           &          &            &                          &                  &                 &              \\
2015 & \multicolumn{1}{c|}{\citet{shahaf2015information}} &           &                    &                    &          & $\times$         & $\times$        &            &                          & $\times$                &                   &                    &                     &                          & \multicolumn{1}{c|}{}                  &                        &                       &                 &                    &           &          &            &                          & $\times$                &                 &              \\
2017 & \multicolumn{1}{c|}{\citet{wu2017event}}           &           & $\times$                  &                    &          &           &          &            &                          &                  &                   &                    &                     &                          & \multicolumn{1}{c|}{}                  &                        &                       &                 &                    &           &          &            &                          &                  &                 & $\times$            \\
2017 & \multicolumn{1}{c|}{\citet{liu2017growing}}        &           & $\times$                  &                    &          & $\times$         &          &            &                          &                  &                   &                    &                     &                          & \multicolumn{1}{c|}{$\times$}                 &                        &                       &                 & $\times$                  &           &          &            &                          &                  & $\times$               &              \\
2017 & \multicolumn{1}{c|}{\citet{laban2017newslens}}     &           &                    & $\times$                  &          &           &          &            &                          &                  &                   &                    &                     & $\times$                        & \multicolumn{1}{c|}{$\times$}                 &                        &                       &                 &                    &           &          &            &                          &                  &                 & $\times$            \\
2018 & \multicolumn{1}{c|}{\citet{wang2018event}}         & $\times$         &                    & $\times$                  & $\times$        &           &          &            & $\times$                        &                  &                   &                    &                     &                          & \multicolumn{1}{c|}{$\times$}                 & $\times$                      &                       & $\times$               &                    &           &          &            &                          &                  &                 &              \\
2018 & \multicolumn{1}{c|}{\citet{tikhomirov2017news}}    & $\times$         & $\times$                  &                    &          &           &          &            &                          &                  & $\times$                 & $\times$                  &                     & $\times$                        & \multicolumn{1}{c|}{}                  &                        & $\times$                     &                 &                    &           &          &            &                          &                  &                 &              \\
2018 & \multicolumn{1}{c|}{\citet{xu2018generating}}      &           &                    &                    &          & $\times$         & $\times$        &            &                          & $\times$                &                   &                    &                     &                          & \multicolumn{1}{c|}{}                  &                        &                       &                 &                    & $\times$         & $\times$        &            &                          &                  &                 &              \\
2018 & \multicolumn{1}{c|}{\citet{zhou2018new}}           &           & $\times$                  &                    &          & $\times$         &          &            &                          & $\times$                &                   &                    &                     &                          & \multicolumn{1}{c|}{}                  &                        & $\times$                     &                 &                    &           &          &            &                          &                  &                 &              \\
2019 & \multicolumn{1}{c|}{\citet{camacho2019analyzing}}  &           & $\times$                  & $\times$                  & $\times$        & $\times$         &          &            & $\times$                        &                  &                   &                    &                     &                          & \multicolumn{1}{c|}{}                  &                        &                       &                 &                    &           &          & $\times$          &                          &                  & $\times$               &              \\
2019 & \multicolumn{1}{c|}{\citet{cai2019temporal}}       &           & $\times$                  &                    &          &           &          &            &                          &                  &                   & $\times$                  &                     &                          & \multicolumn{1}{c|}{}                  & $\times$                      &                       & $\times$               &                    &           &          &            &                          &                  &                 &              \\
2019 & \multicolumn{1}{c|}{\citet{yuan2019dtexsl}}        &           &                    &                    &          &           & $\times$        &            &                          & $\times$                &                   &                    &                     &                          & \multicolumn{1}{c|}{}                  &                        & $\times$                     &                 &                    &           &          &            &                          &                  &                 &              \\
2020 & \multicolumn{1}{c|}{\citet{duan2020comparative}}   &           &                    &                    &          &           & $\times$        &            & $\times$                        &                  &                   &                    &                     &                          & \multicolumn{1}{c|}{}                  & $\times$                      & $\times$                     &                 &                    &           &          &            & $\times$                        &                  &                 &              \\
2020 & \multicolumn{1}{c|}{\citet{liu2020story}}          &           & $\times$                  &                    &          & $\times$         &          &            &                          &                  &                   &                    &                     &                          & \multicolumn{1}{c|}{$\times$}                 &                        &                       &                 & $\times$                  &           &          &            &                          &                  & $\times$               &              \\
2021 & \multicolumn{1}{c|}{\citet{la2021summarize}}       &           & $\times$                  &                    &          &           &          &            &                          &                  &                   &                    & $\times$                   &                          & \multicolumn{1}{c|}{}                  &                        & $\times$                     &                 &                    &           &          &            &                          &                  &                 &              \\
2021 & \multicolumn{1}{c|}{\citet{yu2021multi}}           &           & $\times$                  &                    &          & $\times$         &          &            &                          &                  &                   & $\times$                  &                     & $\times$                        & \multicolumn{1}{c|}{$\times$}                 &                        & $\times$                     &                 &                    &           &          &            &                          &                  &                 &              \\
2021 & \multicolumn{1}{c|}{\citet{liao2021wilson}}        &           &                    &                    &          &           &          &            &                          &                  &                   &                    & $\times$                   &                          & \multicolumn{1}{c|}{}                  &                        & $\times$                     & $\times$               &                    &           &          &            &                          &                  & $\times$               &              \\
2021 & \multicolumn{1}{c|}{\citet{keith2021narrative}}    &           & $\times$                  & $\times$                  &          & $\times$         & $\times$        &            &                          &                  &                   &                    &                     &                          & \multicolumn{1}{c|}{}                  &                        &                       &                 &                    &           &          &            &                          &                  & $\times$               &              \\ \bottomrule
\end{tabular}}
\caption{Summary of the extraction criteria and evaluation metrics used in the reviewed articles.}
\label{tab:metrics-criteria}
\end{table}

\textbf{Relevance}
Relevance metrics evaluate whether the events in the narrative are relevant or significant to a given query or topic  \cite{yan2011evolutionary,yan2011timeline,li2013evolutionary,nguyen2014ranking,li2015tracking, tikhomirov2017news}. In general, relevance is measured by borrowing techniques from traditional search methods in Information Retrieval, such as PageRank and its variations \cite{zhu2012finding, zhu2014finding, tran2015timeline, wang2018event}. However, some approaches use supervised methods to learn a ranking function \cite{binh2013predicting, tran2013leveraging}. The results from such techniques are used to feed other parts of the algorithm or could be directly used to select relevant events, turning this issue into more of a traditional information retrieval problem rather than a narratological one.

\textbf{Content Similarity}
Another approach to extracting narratives is based on modeling content similarity between events. Over two-thirds of the methods use some sort of content similarity measure. There are many ways to do this, in particular, we found the following approaches: surface-level similarity comparisons (e.g., Jaccard similarity or cosine similarity) \cite{hu2011generating, nguyen2014ranking,uramoto1998method}, topic similarity based on topic distribution information (e.g., comparing topic vectors extracted from LDA models) \cite{li2015tracking,chen2015multi}, and entity-based comparisons (e.g., entity co-occurrence in events) \cite{nallapati2004event,zhu2014finding}. 

The exact choice of approach is highly dependent on the event representation. In recent years, researchers have started leveraging advances in text representation with neural embeddings (e.g., BERT) \cite{yu2021multi, keith2021narrative, liao2021wilson, duan2020comparative,tikhomirov2017news}, which have several advantages over traditional frequency-based models and are better able to capture semantic similarities. 

The use of entity-based information in event-based narrative extraction methods to measure event content similarity remains limited in scope, with sparse usage over the years compared to other content similarity measures \cite{camacho2019analyzing, wang2018event, bogel2015time, zhu2014finding, hu2014exploring, hu2011generating, nallapati2004event}. Combining entity information with other types of similarities would provide a much more holistic view of content similarity. Furthermore, expanding upon this approach, content similarity metrics could exploit the \textit{main event descriptors} \cite{norambuenaevaluating} to compute a more precise similarity measure.

\textbf{Coherence}
Coherence metrics evaluate whether the narrative makes sense. Due to their importance as an extraction metric, we show some mathematical formulations of coherence and coherence-like metrics in Table \ref{tab:coherence}. 

While coherence has a formal definition in narratological terms \cite{abbott2008cambridge}, it is just as complex and ill-defined as relevance in computational terms. One particular motivation for the definition of coherence that stands out is the idea of \textit{smoothness} from the Connect the Dots \cite{shahaf2010connecting,shahaf2012connecting,shahaf2013information,shahaf2015information} series of works. In particular, they use the concept of \textit{word influence} and \textit{word activations} (i.e., the sustained importance of the word in a storyline) to construct stories that have smooth transitions. 

Other approaches compute coherence based on content similarity. These works also seek to generate \textit{smooth} stories by avoiding drastic local changes based on content similarity \cite{yan2011evolutionary, zhu2012finding, tran2013leveraging,zhou2014generating, zhu2014finding,li2015tracking,zhou2018new, liu2017growing,xu2018generating,camacho2019analyzing,liu2020story,keith2021narrative}, without explicitly defining active words or topics like the original Connect the Dots approach. Finally, one approach also considers coherence around the idea of causality \cite{binh2013predicting,tran2013leveraging} in a supervised setting (e.g., causal signals in text).

\begin{table}[!htb]
\small
\resizebox{\textwidth}{!}{\begin{tabular}{@{}ccp{8cm}c@{}}
\toprule
\textbf{Formula}                                                                                                                                                                 & \textbf{Resolution Level} & \textbf{Description}                                                                                                                                                                                                                                                       & Source                      \\ \midrule
$\displaystyle\frac{1}{|S| - 1}\sum_{(c_i, c_j) \in S}{\mathsf{JaccardSim}(c_i ,c_j)}$                                                                                                                          & Events as Clusters        & This is a measure of coherence based on average Jaccard Similarity along a story $S$ based on cluster words.                                                                                                                                                               & \cite{xu2018generating}     \\
$\displaystyle\frac{1}{|S| - 1}{\mathsf{CosineSim}(c_i ,c_j)}$                                                                                                                                         & Events as Clusters        & This is a measure of coherence based on average Cosine Similarity along a story $S$ based on cluster centroids.                                                                                                                                                            & \cite{liu2017growing}       \\
$\displaystyle\max_{\mathsf{activations}}\left\{\min_{(d_i, d_j) \in S}{\sum_{w \in \mathsf{Words}}{\mathsf{Influence}(d_i, d_j|w)\cdot \mathbbm{1}(w \text{ active in } d_i, d_j)}}\right\}$ & Events as Documents       & This is the full form of the coherence for a storyline $S$ from the original Connect the Dots algorithm. It is based on maximizing the sum of word influences over active words in the storyline. Influence can be changed for any other type of scoring mechanism.    & \cite{shahaf2010connecting} \\
$\displaystyle\min_{(d_i, d_j) \in S}{\mathsf{CosineSim}(d_i ,d_j)}$                                                                                                                                   & Events as Documents       & This is a measure of coherence based on the minimum Cosine Similarity along a story $S$ based on document vectors.                                                                                                                                                         & \cite{zhou2014generating}   \\
$\displaystyle\min_{(d_i, d_j) \in N}{\sqrt{\mathsf{SurfaceSim}(d_i,d_j) \cdot \mathsf{TopicSim}(d_i, d_j)}}$                                                                                 & Events as Documents       & This is a measure of coherence for a narrative $N$ based on the minimum geometric mean of Surface-level similarity (e.g., cosine similarity) and Topic-level similarity (e.g., Jensen-Shannon divergence). It is based on document vectors and topic distribution vectors..                    & \cite{keith2021narrative}   \\
$\displaystyle\frac{\sum\limits_{i, j \in |D|\times|D|}{w_i \cdot w_j \cdot \Phi \cdot \mathsf{Soergel}(d_i, d_j) \cdot |t_i - t_j|}}{\sum\limits_{i, j \in |D|\times|D|}{w_i \cdot w_j \cdot \Phi }}$      & Events as Documents       & This is a measure of incoherence rather than coherence. It is based on the average Soergel distance and includes a temporal distance term as well. The events are weighted by their relevance ($w_i$ and $w_j$) and their temporal distance using a custom kernel $\Phi$. & \multicolumn{1}{l}{}        \\
$\displaystyle\sum_{s \in E}{\frac{\mathsf{Count}_{\mathsf{match}(s, E)}(\mathsf{gram}_n)}{\mathsf{Count}_{\mathsf{Previous}(E)}(\mathsf{gram}_n)}}$                                                         & Events as Sentences       & This is a measure of coherence based on the n-gram overlap between the current event sentences and the sentences of the previous summary of the timeline.                                                                                                                  & \cite{tran2013leveraging}   \\
$\displaystyle\frac{1}{1 + \exp{(JS(E, \mathsf{Previous}(E))))}}$                                                                                                                             & Events as Sentences       & This is a measure of coherence based on the Jensen-Shannon divergence between the current event sentence and the previous event sentence of the timeline.                                                                                                                  & \cite{li2015tracking}       \\
$\displaystyle\frac{\sum\limits_{\delta=-\Delta/2}^{\delta=\Delta/2}{\exp{(-\delta / v)} \cdot KLD(E_t, E_{t-\delta}})}{\sum\limits_{\delta=-\Delta/2}^{\delta=\Delta/2}{\exp{(-\delta / v)}}}$              & Events as Sentences       & This is a measure of coherence based on Kullback-Leibler divergence between the current event at time $t$ and all the other local events in a $\Delta$ time window surrounding the event. The events are weighted by their temporal distance based on parameter $v$.                              & \cite{li2013evolutionary}\\
\bottomrule
\end{tabular}}%
\caption{A sample of different formulations of coherence from the reviewed articles.}
\label{tab:coherence}
\end{table}

\textbf{Coverage-like Metrics}
Coverage-like metrics evaluate whether the extracted narrative properly covers the relevant events, stories, or topics. These metrics include coverage itself and related metrics, such as redundancy and diversity. The most basic form of coverage is simply the percentage of topics or relevant events covered by the extracted representation (or some variation of this metric)
\cite{guha2005unweaving,yuan2019dtexsl,duan2020comparative,keith2021narrative}, or a probability estimation \cite{shahaf2012trains,shahaf2013metro,shahaf2013information,shahaf2015information,xu2018generating}. Equation \ref{eq:coverage} shows an example formulation of coverage for a cluster $c$, where $\Pi$ represents an extracted narrative with storylines $l$.
\begin{equation}
\mathsf{Cover}_{\Pi}(c) = 1 - \prod_{l \in \Pi}{\left(1 - \mathsf{Cover}_{l}(c)\right)}
\label{eq:coverage}
\end{equation}

Another approach to compute coverage is to do a content similarity comparison between the output and the full data set (or a relevant subset) \cite{yan2011evolutionary,li2013evolutionary,li2015tracking}. In contrast, redundancy and diversity \cite{zhu2014finding,zhu2012finding,yan2011evolutionary,duan2020comparative,wang2018event} metrics are based on the idea that events should not be covered more than necessary, thus high redundancy can lead to coverage problems.

\textbf{Structural Information}
Some works evaluate the structure of the output narrative representation. In particular, these metrics consider aspects such as size (in general) or connectivity (in graph-based narratives). 

Size can be used as a proxy for complexity (e.g., length of the timeline) \cite{xu2018generating}. In most cases, rather than as an evaluation metric, size is used as a constraint (e.g., setting a maximum story length) \cite{yuan2019dtexsl,zhou2018new,zhou2014generating,shahaf2015information, shahaf2013information, shahaf2013metro, shahaf2012trains}. 

Connectivity metrics \cite{shahaf2015information, shahaf2013information, shahaf2013metro, shahaf2012trains} are used to ensure that narrative graphs avoid isolated stories, as they should be interwoven throughout the narrative. Structure metrics are mostly analyzed at a global level (e.g., the total number of connected stories). However, it is possible to consider local structural features, such as node degrees \cite{tannier2013building}.

Exploiting the internal article structure \cite{tran2013leveraging,bogel2015time,tikhomirov2017news} is another piece of structural information used by some methods. Most breaking news articles are written following the \textit{inverted pyramid structure} \cite{norambuenaevaluating}, where the most important information---the main event descriptors---is shown first in the \textit{lead}. Thus, the first few lines of an article describe its main event \cite{tran2013leveraging} and subsequent paragraphs may contain more details and reference previous events. 

\textbf{Content References}
Another criterion to consider in news narrative extraction is the use of content references. As mentioned before, some news articles make explicit references to previous works in their body. Note that this differs from explicit date-based references discussed before, which rely on explicit temporal information. This approach also differs from general content similarity because of its goal of identifying specific references rather than global similarity.

One way to identify these references is to compare the lead of a news article with the additional information paragraphs of another article \cite{tikhomirov2017news}. Other approaches identify references based on sentence co-occurrence without considering article structure \cite{yu2021multi}. Alternatively, a set of core features \cite{huang2014discovering,cai2019temporal} (e.g., relevant keywords or main event descriptors) could be identified and used to detect references in other articles. Once identified, these references can be used to identify relevant events based on reference-based metrics (e.g., bibliographic coupling). 

\textbf{Temporal Features}
Temporal information, such as the temporal distance between events or specific date references, has been used. In particular, temporal distance is commonly used to penalize events that would otherwise be similar in content. For example, consider two articles describing separate protests in a city, one during the year 2000 and another in the year 2010. These two articles would likely be very similar in terms of content, including both surface-level features and topic distributions. However, given the temporal separation between them, they would likely refer to different events. Thus, a common strategy is to define an exponentially decreasing term of the form $C_0\exp{\left(\frac{-\Delta t}{\sigma}\right)}$ (or similar), where $C_0$ and $\sigma$ are pre-defined constants \cite{yang2006tracing,yang2009discovering,yu2021multi, wei2014exploiting, yan2011timeline,liu2020story,liu2017growing,hu2014exploring,nallapati2004event,huang2014discovering}, although there are other approaches, such as kernels to perform temporal proximity projections \cite{yan2011timeline,wang2018event} or overlap-based measures \cite{chen2010tscan,chen2008tscan}. However, we note that the use of a temporal penalty is not always desired. Some events are continuations of stories that did not have anything new to report for a long time. For example, the investigation results of a flight accident might come much after the accident itself has been covered, leading to temporal gaps in story coverage \cite{laban2017newslens,wei2014exploiting}. Thus, it is necessary to distinguish between continuations and completely new storylines when the time gap is high enough. 

Burstiness and frequency measures and metrics based on these (e.g., energy values) are other time-based criteria used to identify relevant events and dates \cite{chen2008tscan,yang2009discovering,chen2010tscan,yan2011timeline,binh2013predicting,huang2013optimized,wei2014exploiting,nguyen2014ranking,tran2015timeline,tikhomirov2017news, laban2017newslens,yu2021multi, chen2015multi}. For example, periods with many publications are likely to contain important events. Alternatively, a specific event might be reported several times by different outlets. Finally, other temporal features include the use of specific temporal expressions or date references in the text \cite{binh2013predicting,tran2013leveraging,tran2015timeline,bogel2015time,la2021summarize,liao2021wilson,tannier2013building} to identify temporal cross-references between documents.

\section{Evaluation Metrics}
In this section, we discuss the evaluation approaches for the narrative output of the extraction methods. In particular, we show the evaluation metrics used to assess the quality of the extracted narratives. These output metrics are generally intended to be interpreted by humans, unlike the extraction criteria which may or may not be easy to interpret. In particular, user-based evaluation metrics (e.g., task performance or user perception) are an important subset of output evaluation criteria. The second part of Table \ref{tab:metrics-criteria} provides an overview of the different output evaluation criteria.

\subsection{Computational Metrics}
These metrics seek to evaluate the extracted narrative based on computational measures of narrative quality. These metrics are usually supervised, requiring a gold standard data set to be computed.

\subsubsection{Supervised}
We first discuss supervised approaches. In particular, we identified three broad types of metrics here: traditional information retrieval metrics, summarization metrics, and ranking metrics.

\textbf{Traditional IR Metrics}
Several works---about a third of the reviewed articles---rely on classical evaluation metrics such as accuracy, precision, recall, and the $F_1$ score \cite{nallapati2004event, yang2006tracing, lin2007individualized, yang2009discovering, khurdiya2011multi, hu2011generating, tannier2013building, huang2014discovering, wei2014exploiting, hu2014exploring, tran2015timeline, bogel2015time, chen2015multi, wang2018event, cai2019temporal, duan2020comparative} taken from traditional information retrieval and machine learning literature. In particular, these approaches evaluate the quality of the output by measuring whether \textit{events} or their \textit{connections} were identified correctly.  Some methods also use variations of these basic metrics, such as the mean average precision \cite{binh2013predicting,nguyen2014ranking} over multiple dates.

\textbf{Summarization Metrics}
Specialized metrics from the summarization domain have also been used to evaluate narratives in several works---about a third of the reviewed works use them. 

In particular, ROUGE (Recall-Oriented Understudy for Gisting Evaluation) metrics \cite{ogawa2011recall} have been used to evaluate the output of narrative extraction methods, mostly in timeline summarization works with a sentence-level event resolution  \cite{binh2013predicting,tran2013leveraging,nguyen2014ranking,chen2015multi,liao2021wilson,yan2011evolutionary,li2013evolutionary,li2015tracking,yu2021multi,huang2013optimized,tikhomirov2017news,la2021summarize}, but also in works with a document-level resolution  \cite{hu2014exploring,zhou2014generating,zhou2018new,yuan2019dtexsl}. ROUGE metrics include variations such as ROUGE-N (which measures the overlap of N-grams), ROUGE-L (which measures the longest common subsequence), ROUGE-W (a weighted version of ROUGE-L that favors consecutive subsequences), ROUGE-SU (skip-bigram and unigram-based co-occurrence statistics) and their precision, recall, and $F_1$ score variants. The most common variant is ROUGE-N, which we show in Equation \ref{eq:ROUGE}.
\begin{equation}
    \mathsf{ROUGE\text{-}N}=\frac{{\sum}_{s\in R}{\sum}_{\mathsf{gram}_{n}\in s}\mathsf{Count}_{\mathsf{match}}(gram_{n})}{{\sum}_{s\in R}{\sum}_{\mathsf{gram}_{n}\in s} \mathsf{Count}(\mathsf{gram}_{n})}
    \label{eq:ROUGE}
\end{equation}
In Equation \ref{eq:ROUGE}, $n$ represents the length of the $n$-gram, and $R$ represents the reference summaries (i.e., the ground truth). $\mathsf{Count}_\mathsf{match}(\mathsf{gram}_n)$ represents the maximum number of $n$-grams that co-occur in a candidate summary and the reference summaries. $\mathsf{Count}_{R}(\mathsf{gram}_n)$ represents the number of $n$-grams in the reference summaries. 

An alternative is to measure the average \textit{summary-to-document content similarity} where the summary is compared against the documents in the data set using a text similarity measure (e.g., cosine similarity) \cite{chen2008tscan,chen2010tscan}.

We note that these metrics are mostly used with linear representations rather than graph-based models---only three of the reviewed works that extract graphs use summarization metrics \cite{chen2008tscan,chen2010tscan,hu2014exploring}. This contrasts heavily with the case of traditional information retrieval metrics, where the split was much more balanced between linear (\textasciitilde{}40\%) and graph representations (\textasciitilde{}60\%). This might be due to the inability of these metrics to handle complex structures.

\textbf{Ranking Metrics}
Other works rely on ranking-based metrics, like those used in traditional search tasks from information retrieval. For example, Wang et al. \cite{wang2018event} use a relevance-based approach to evaluate their event phase summaries. Liao et al. \cite{liao2021wilson} evaluate the ranking performance of WILSON with the mean reciprocal rank and discounted cumulative gain \cite{avramidis2013rankeval}. Cai et al. \cite{cai2019temporal} use the normalized discounted cumulative gain \cite{yilmaz2008simple} to evaluate all their events.  

\textbf{Clustering Metrics}
Liu et al. \cite{liu2017growing, liu2020story} used clustering metrics to evaluate the event nodes---which are represented as clusters of articles---in their Story Forest method. In particular, they use the homogeneity, completeness, and V-measure scores \cite{rosenberg2007v}. These metrics require labeled data sets to be computed, thus they are supervised despite being designed to evaluate unsupervised clustering methods. In particular, homogeneity is larger when each extracted cluster only contains members of a single class. In contrast, completeness is maximized when all members of a true class are in the same cluster. Finally, the V-measure takes both of these metrics and computes the harmonic mean between them, similar to how the $F_1$ score treats precision and recall in traditional classification metrics. We note that none of the other events as clusters methods used these metrics or other similar clustering metrics to evaluate their models. Instead, they relied on traditional information retrieval metrics like accuracy, precision, recall, and the $F_1$ score.

\subsubsection{Unsupervised Metrics}
We now discuss unsupervised approaches. In general, there are far fewer works relying on unsupervised metrics to evaluate the final narrative output. 

\textbf{Coherence} In general, coherence is not used to evaluate the output narrative despite being a useful metric during the extraction process. One exception is Xu et al. \cite{xu2018generating}, who evaluate their output using a weighted average of story coherence (based on Jaccard similarity) and story size.

\textbf{Coverage}
Xu et al. \cite{xu2018generating} evaluate their output by treating coverage as a structural measure, making the assumption that good coverage of topics means that the structure of their metro map representation is good. However, due to the formulation of coverage based on whether the topical clusters of the data set are covered, it does not explicitly consider the structure of the output, which makes it inappropriate to evaluate the structure. In contrast, Bögel et al. \cite{bogel2015time} use a notion of coverage based on event connections in a graph (i.e., an article is covered if there is at least one edge connecting it) that could be treated more as a structural measure than the topical concept of coverage.

\textbf{Dispersion} Camacho et al. \cite{camacho2019analyzing} use the \textit{dispersion coefficient}---originally proposed as an evaluation metric for storytelling in the intelligence analysis domain \cite{hossain2011helping}---to evaluate their storyline. In particular, the dispersion coefficient is based on the Soergel distance, although other distance metrics could be used \cite{rigsby2018storytelling}. In particular, dispersion is based on Swanson’s complementary but disjoint hypothesis \cite{swanson1991complementary}---where articles that have no explicit common elements yield important inferences or insights when combined. These insights are not apparent from the separate documents. Furthermore, the authors propose a new evaluation metric to measure story flow based on Swanson's hypothesis called the dispersion coefficient, shown in Equation \ref{eq:dispersion}. We note that this particular version is based on the Soergel distance ($S$), but any other distance metric between documents could be used in practice.
\begin{equation}\mathsf{Dispersion}(d_1, \ldots, d_n) =
1 - \frac{1}{n - 2}\sum_{i = 1}^{n - 2}{\sum_{j = i + 2}^{n}\mathsf{D}(d_i, d_j)}, \quad \text{ with } \quad    \mathsf{D}(d_i, d_j) = \begin{cases}
    \frac{1}{n + i - j}, & \text{if } \mathsf{S}(d_i, d_j) < \theta\\
    0, & \text{otherwise.}
  \end{cases}
\label{eq:dispersion}
\end{equation}

\textbf{Diversity and Redundancy}
Finally, another alternative is a diversity metric to ensure proper coverage or low redundancy. In particular, Duan et al. \cite{duan2020comparative} used diversity---based on the average pairwise similarity of sentences (see Equation \ref{eq:pairwise}---to evaluate the performance of their comparative timeline extraction method.
\begin{equation}
\mathsf{Diversity} = 1 - \frac{1}{|S|^2}\sum{s_i \in S}{\sum{s_j \in S}{1 - \mathsf{CosineSim}(s_i, s_j)}}
\label{eq:pairwise}
\end{equation}

\subsection{User Evaluation Metrics}
These metrics seek to evaluate the extracted narrative based on subjective user measures or task performance measures. 

\textbf{Task-oriented Evaluation}
Task-oriented metrics require designing a series of benchmark tasks to measure the number of correct answers, accuracy, how much time the users take to complete the task or some other measure of correctness or quality. Few works use task-oriented evaluation metrics: Metro Maps \cite{shahaf2012trains, shahaf2013metro}, Information Cartography \cite{shahaf2013information, shahaf2015information}, and the SToRe system \cite{lin2006topic,lin2008storyline}. These works rely on event-based representations and all of them evaluate extraction methods as retrieval tools following a similar approach. In particular, there are \textit{micro-knowledge} tasks that measure how the extracted narratives help users \textit{retrieve information} faster and \textit{macro-knowledge} tasks that measure how the extracted narratives help users \textit{understand the big picture}.

For micro-knowledge tasks, all works create a series of simple retrieval questions such that the answers can be easily classified as right or wrong. For example, retrieving dates, facts, relevant entities or the main event descriptors. Users are evaluated by measuring how many correct answers (i.e., accuracy) they can get in a fixed amount of time and the rate at which they answer these questions \cite{shahaf2012trains, lin2006topic, lin2008storyline}. Another metric used at the micro-knowledge level is the ease of navigation, estimated by the number of documents that users clicked per correct answer \cite{shahaf2012trains, lin2006topic, lin2008storyline}.

For macro-knowledge, some form of summarization is used to evaluate the narratives. Shahaf et al. \cite{shahaf2012trains} asked users to create summaries based on different narrative representations and then used crowdsourcing to evaluate user preference over those summaries. However, these benchmark tasks do not go beyond basic retrieval and summarization. Tasks that require higher levels of knowledge and cognitive work (e.g., analysis tasks) are not covered by these evaluations. In general, the inherent difficulty of designing benchmark tasks that can be easily evaluated might be one of the reasons why user-based evaluations of extraction methods usually rely on subjective ratings rather than task-oriented metrics.

\textbf{Subjective Evaluation}
Most of the works that rely on user evaluations use subjective measures (i.e., user perception metrics). These subjective metrics include concepts from usability, including criteria such as user preference \cite{shahaf2013information, hu2014exploring, keith2021narrative}, visual presentation \cite{keith2021narrative}, and ease of use \cite{keith2021narrative,lin2006topic}. Other metrics include effectiveness as perceived by the users (e.g., perceived helpfulness or usefulness), satisfaction, and comprehensibility \cite{lin2006topic,lin2008storyline, tran2015timeline}. Alternatively perceived familiarity before and after using the extracted narrative can be a useful measure of usefulness \cite{shahaf2010connecting}.

Lastly, user-perceived quality is another widely used approach to evaluate extracted narratives. The user-perceived quality criteria mostly correspond to the quality criteria metrics defined before \cite{shahaf2010connecting, shahaf2012connecting, zhu2012finding, zhu2014finding, tran2015timeline, camacho2019analyzing, keith2021narrative}, including coherence, coverage, redundancy, relevance, dispersion, and similar variations (e.g., broadness). We note that these user perception metrics suffer a similar problem as their computational counterparts---they are fuzzy concepts that could be defined differently. This is further compounded by the subjective nature of these evaluations. 

Other works rely on asking users whether they consider specific elements of the narrative as correct. For example, asking whether a specific connection is correct, whether the selected documents are relevant, whether a specific storyline is logically coherent, or about the number of coherent and relevant documents \cite{liu2017growing, liu2020story,camacho2019analyzing}. This is similar to traditional information retrieval metrics that rely on ground truth information. However, in this case, rather than using a previously defined gold standard, the accuracy measures are defined purely on subjective perceptions. Finally, another approach is to ask users to compare the ground truth with the output narrative---from potentially multiple methods---and rank them according to their preference based on their knowledge of the topic \cite{liao2021wilson}.

\section{Discussion}
We now discuss our findings. We start by addressing the structural choices in narrative representation. Next, we address some of the challenges of extraction methods. Then, we turn our attention toward evaluation methods, including benchmark data sets, computational metrics, and user-based evaluations. Afterward, we discuss practical applications of news narrative extraction. Finally, we discuss recent trends, open challenges, and potential research directions.

\subsection{Narrative Structure}
The choice of the core structure is an important aspect of narrative representation. Using a linear structure provides a simple approach to represent a narrative with a single storyline, but it does not appropriately model the nuances of narratives with multiple stories. In contrast, graph-based structures allow the modeling of different interactions between storylines (e.g., convergent and divergent stories) \cite{keith2021narrative}. Linear representations are implicitly directed, but graph-based representations may or may not be directed. Directed graphs usually exploit the underlying temporal relationships to determine the direction of the connections between elements. When the connection between basic units is guided by temporal constraints it naturally gives rise to directed acyclic graphs. Directed acyclic graphs provide the most flexibility while also accounting for the temporal nature of a narrative. However, not all directed graph models are acyclic, as some use specific types of relationships that allow the creation of cycles (e.g., same-event relations).

A representation that falls between linear and fully graph-based representations is the tree-based representation \cite{liu2017growing, liu2020story, zhou2014generating, zhou2018new, yuan2019dtexsl}. Such models allow for more flexible structures than linear representations. In particular, they are able to model story divergence (i.e., multiple storylines splitting off from the root or other nodes). Unlike graph-based models, they are not able to model story convergence (e.g., two stories joining into a final event), as that would break the tree structure. Tree-based structures have not been deeply explored in the literature and could provide an intermediate approach between linear and graph-based representations in terms of complexity, allowing easier understanding by users while retaining some flexibility. However, the inability to model story convergence might limit their applications.

\subsection{Extraction Methods}
\textbf{Scalability and Computational Cost}
Most extraction methods discussed in this survey suffer from issues when dealing with big data, as the processing pipelines are quite expensive in terms of computational power and they might not be easily parallelizable. One of the simplest methods to reduce computational cost is to filter the data beforehand. This turns the computational cost problem into an information retrieval problem, where the most relevant documents must be retrieved before extracting the narrative itself. Many methods assume that the data has been pre-filtered to a relevant set of news articles. Including a filtering step adds an additional element to the pipeline, thus increasing the risk of errors. Moreover, defining an adequate concept of relevance for this method might prove problematic in itself. Nevertheless, this provides a simple approach to mitigate the ever-increasing available amount of data. 

Another approach is to deal with extraction in an online manner \cite{liu2020story}. Most news narrative extraction methods are offline methods that analyze an entire set of news articles. However, extracting the stories in an online manner without disrupting the pre-existing structure would offer a computationally cheaper alternative. This is similar to the approach used by traditional TDT systems that sought to track the events of a topic in an online manner \cite{allan2012topic}. However, it would also require handling the structure of events associated with the narrative, which is not considered by traditional TDT.

\textbf{Unified Metrics}
One of the limitations of current approaches is that there are multiple versions of coherence and similar metrics. Coherence itself is an ill-defined term in practice and formalizing it in a computational or mathematical definition is a difficult task. The different definitions of coherence-like metrics focus on measuring different aspects of the narrative. Moreover, additional constraints can be considered to enforce coherence beyond numerical metrics (e.g., events sharing common entities). In general, a hybrid extraction approach that mixes multiple metrics (e.g., through a linear combination) and also includes such constraints might provide better results. 

\subsection{Evaluation Methods}
\textbf{Benchmark Data Sets}
In general, most works collect their own data or use a subset of a pre-existing news data repository. For example, some use data sets from TDT literature \cite{nallapati2004event,wei2014exploiting,chen2008tscan,chen2010tscan}, DUC/TAC conferences \cite{chen2015multi}, or other general news repositories \cite{chieu2004query}. Most works do not publish their data sets. However, there is a subset of timeline summarization works that have provided evaluation data sets that have been adopted in several works as benchmark data. We present these data sets in Table \ref{tab:tls-data}. These data sets are appropriate for the ``events as sentences'' resolution level that timeline summarization uses. However, they do not provide a direct way to evaluate methods that use other resolution levels. Furthermore, we note that there are no such benchmark data sets for the other resolution levels of the narrative extraction tasks considered in this survey. The lack of appropriate benchmark data for the document-level and cluster-level resolutions makes comparing methods harder and makes replicability harder. 

\begin{table}[!htb]
\small
\begin{tabular}{@{}ccc@{}}
\toprule
\textbf{Data Set} & \textbf{Source}                & \textbf{URL}                                     \\ \midrule
Timeline 17       & \cite{tran2013leveraging}      & \url{https://github.com/complementizer/news-tls}        \\
Crisis            & \cite{tran2015balancing}      & \url{https://github.com/complementizer/news-tls}        \\
COVID-TLS         & \cite{la2021summarize}         & \url{https://github.com/MorenoLaQuatra/SDF-TLS}  \\
TLS-COVID19       & \cite{pasquali2021tls}         & \url{https://github.com/LIAAD/tls-covid19}  \\
Entities          & \cite{ghalandari2020examining} & \url{https://github.com/complementizer/news-tls} \\
MTLS Data         & \cite{yu2021multi}             & \url{https://yiyualt.github.io/mtlsdata/}        \\ \bottomrule
\end{tabular}
\caption{Benchmark data in the timeline summarization works.}
\label{tab:tls-data}
\end{table}

\textbf{Computational Metrics Limitations}
We note that most of the narratives discussed in this survey only consider the content (e.g., traditional information retrieval metrics) without accounting for the order nor the structure of the narrative. Some metrics consider ordering information, but only at a linear structure level. For example, story-level measures of coherence consider the connections between consecutive documents \cite{shahaf2010connecting} or the dispersion coefficient which models story flow \cite{camacho2019analyzing}. The ranking evaluation metrics also include some underlying notion of order, however, this notion is limited to a linear structure at best. The structural version of coverage based on event connections used by Bögel \cite{bogel2015time} is based on local connections only, but it does not account for the full structure of the graph. Thus, current metrics for the narrative extraction task are unable to deal with complex narrative structures. 

Given this limitation, it would be ideal to consider metrics that account for both order and structure to provide a proper evaluation of a narrative.  For linear narratives, it would be sufficient to consider content and order, as the structure itself is fixed. An approach to solve this would be to consider a metric based on weighted edit distance \cite{weigel1994normalizing} as it considers both the order of the elements and their contents (by defining weights according to event similarities or an adaptation of the previously discussed metrics). For non-linear narratives, a similar approach could use graph edit distance \cite{abu2015exact} with custom costs, as this metric would consider structure, order, and content.

However, the previous proposal would be supervised, as we would need a narrative against which to compare the output. Devising an unsupervised approach is a more challenging issue, particularly for graph-based narrative representations. One alternative is to attempt to extend coherence and dispersion measures to such graphs. For example, given a directed graph with a single starting event and a single ending event, it could be possible to compute all routes from start to end and obtain a weighted average of the coherence or dispersion of these routes. However, for more complex graph structures it might be too costly in computational terms to do such computation. Designing an unsupervised evaluation metric remains an open challenge.

\textbf{Benchmark Tasks and User Evaluations}
User-based evaluations usually focus on subjective measurements rather than objective task performance. This is due to the lack of properly defined evaluation frameworks and benchmark tasks. Current approaches rely on micro-knowledge tasks---tasks focused on information retrieval---that evaluate the number of correct answers (i.e., user accuracy) over time, and macro-knowledge tasks---tasks focused on summarization---that indirectly evaluate the quality of the extracted narrative by measuring the quality of a user-generated summary. These approaches are limited and do not capture all the nuances associated with narrative sensemaking. Moreover, they do not cover more complex tasks beyond retrieval and summarization.

One possible solution would be to design more holistic evaluations based on different types of benchmark tasks. In particular, the use of Bloom's taxonomy \cite{bloom1956taxonomy} could provide a useful framework to define such tasks as seen in other sensemaking applications \cite{burns2020evaluate} or cognitive tasks in general \cite{collins2016assessing}. Another possible solution would be to borrow the concept of insight-based evaluations \cite{north2011comparison}. Rather than focusing on benchmark tasks with specifically defined tasks and correct answers, the evaluation would be open-ended and would focus on analyzing the insights generated by the users.

\subsection{Practical Applications}
Event-based news narrative extraction has several practical applications beyond journalistic analysis tasks. Most of these applications seek to help with the issue of \textit{information overload} in different contexts \cite{shahaf2013information}. We briefly discuss some potential applications explored or mentioned in some of the reviewed works.

\textbf{Disaster Management}
Disaster management \cite{zhou2014generating,xu2018generating,zhou2018new,yuan2019dtexsl} could benefit from using extraction approaches to keep track of disasters or other similarly negative incidents. In particular, to minimize losses caused by a disaster, one of the critical tasks in disaster management is to efficiently analyze and understand situation updates. Doing this requires effective methods to navigate a multitude of documents such as news or reports related to the disaster. Domain experts need to obtain condensed information about the disaster and its evolution \cite{li2013empirical}. Thus, news narrative extraction could help experts to understand the evolving situation and devise a proper strategy.

\textbf{Open Source Intelligence}
Open source intelligence (OSINT) is intelligence that is synthesized using publicly available data \cite{evangelista2021systematic}. While OSINT data sources leverage more than just traditional news articles \cite{gibson2016acquisition}, OSINT could still benefit from news narrative extraction techniques. In particular, news narrative extraction methods could help intelligence analysts explore the information landscape and find key events \cite{keith2021narrative}. Furthermore, these techniques could help analysts in prediction tasks by providing support and evidence \cite{camacho2019analyzing}.

\textbf{Misinformation and Fact-Checking}
News narrative extraction methods could aid fact-checkers in their tasks by providing them with an overview of the current narrative and highlighting key relevant events \cite{keith2021narrative}. However, current methods do not include explicit ways to model misleading or outright false information.

\textbf{Financial Markets}
News narrative extraction could aid financial analysts to understand the information landscape \cite{cai2019temporal}. For example, market news is regarded as an important data source in the context of financial analysis \cite{cai2019temporal}. In particular, being able to understand and exploit the hidden information in the raw news data could help analysts adapt their strategies and reduce their financial risk.

\subsection{Recent Trends and Open Challenges}
\textbf{Timeline Summarization Variations}
Recent works have proposed some variations on the traditional timeline summarization task. In particular, Duan et al. \cite{duan2020comparative} proposed the \textit{comparative} TLS task and Yu et al. \cite{yu2021multi} proposed the Multi-TLS task. These two works highlight the fact that simple linear representations of narratives are naturally limiting unless applied to the most simple of narratives. Thus, the creation of similar tasks to address some of the shortcomings of these representations is a natural progression. However, it raises the question of whether these extensions would benefit from borrowing elements from the methods that use more complex representations discussed in this survey. A natural extension would be to consider a graph-based representation that allows for multiple storylines and comparisons without further modifications. This approach would address both the comparative TLS and MLTS tasks.

In this context, we note that most of the reviewed articles with a sentence-level event resolution used a linear structure (see Table \ref{tab:papers}). The only exceptions were the disaster storyline extraction systems \cite{zhou2014generating,zhou2018new,yuan2019dtexsl} with their local tree representation. However, these methods are designed with a specific news topic in mind---disaster news---and are able to leverage specific characteristics of the topic (e.g., the disaster moves over time). Thus, it would not be possible to directly adapt it to other types of news without addressing this issue.

Furthermore, we note that there are no inherent limitations to sentence-level representations that prevent them from being extended beyond linear narratives, which makes the lack of graph-based approaches an opportunity for future research. Finally, while we did not find such a suitable graph-based approach in the traditional news domain, there is one example from the social media domain---which has its own set of challenges in terms of narrative extraction---that can be found in Ansah et al. \cite{ansah2019graph}. This work proposes a tree-based narrative representation with sentence-level event representation using tweets. This approach extends the traditional TLS by allowing divergent storylines to emerge instead of just a single timeline. Such an approach could be adapted to traditional news narrative extraction.

\textbf{Multi-resolution Methods}
Currently, all the narrative extraction approaches that we reviewed work on a singular resolution level (sentences, documents, or clusters). Existing attempts at multiple resolution levels only change the scope of the data \cite{shahaf2013information, shahaf2015information} (i.e., applying the method again on a new subset of the data), they do not seek to change the underlying event resolution. Another perspective corresponds to the multi-level presentations of disaster storylines by Zhou et al. and Yuan et al. \cite{zhou2014generating,zhou2018new,yuan2019dtexsl}, which use global and local levels to represent the narrative. However, the underlying event representation remains the same and no efforts have been made to make a model that handles multiple levels of event resolution. Developing models that provide a multi-resolution approach remains an open challenge.

\textbf{Interactivity}
Most works on news narrative extraction provide surface-level interactions \cite{tannier2013building,shahaf2013information, shahaf2015information} such as re-arranging elements and changing the layout, showing details on demand (e.g., all details about a news article), zooming, or performing basic filtering, highlighting, and searching. However, there is still a need for better interaction models that give users more control and feedback when exploring and manipulating the narrative. Some models \cite{shahaf2010connecting,shahaf2012connecting} allow more in-depth refinement by letting the user specify elements that need to be changed and then evaluating all possible replacement and insertion actions. Building upon this feature-based feedback, Shahaf et al. \cite{shahaf2012trains} designed a method to learn a \textit{personalized coverage function} that can be optimized to find a personalized narrative. 

Another approach by Bögel et al. \cite{bogel2015time} allows parametric interaction to modify the extracted graph in real time, helping the user understand how the narrative changes based on the parameters. However, this approach requires the users to understand the underlying model parameters. In this context, \textit{semantic interactions} could be useful to aid users modify the model without deep understanding of the underlying parameters. Semantic interactions \cite{wenskovitch2018effect} are used in sensemaking applications to directly reflect the analytical thought process of analysts about data (e.g., by using information about how analysts organize documents or highlight text), as opposed to parametric interaction that manipulates model parameters (e.g., sliders and keyword weights). Thus, capturing a user model through semantic interaction could lead to a better narrative model. 

\textbf{Misinformation in News}
Recent works have highlighted the need for future work to model source bias, information validity, transparency, and credibility as an effort to model and counter misinformation \cite{laban2017newslens,keith2021narrative}. Existing narrative representations could be enhanced by including additional attributes in their representations and extraction algorithms. 

Works on misinformation detection focus on the propagation structure and content to determine whether a certain article or publication contains misinformation \cite{guo2019future,wu2019misinformation}. Other methods rely on crowdsourcing \cite{guo2019future} to detect misinformative content early. However, these methods do not model misinformation as part of an overarching narrative. Instead, they focus on local elements (e.g., a specific event). Thus, a holistic narrative approach could be useful in this context.

The issue of misinformation is also highly relevant for a series of recent works on disaster tracking by using news narrative extraction \cite{xu2018generating,zhou2018new,yuan2019dtexsl}. However, none of these methods address this issue and rely on the underlying assumption that the set does not contain false or misleading information. Thus, creating a narrative extraction model that accounts for misinformation would be of vital importance in the context of disaster tracking.

\section{Conclusions}
This literature review focused on narrative extraction and its related tasks of representation and analysis, synthesizing findings from 54 studies and identifying recurring types of representational structures, extraction criteria, and evaluation metrics. We further analyzed the articles and identified a series of recent trends, open challenges, and potential research directions. In terms of limitations, we highlight the lack of benchmark data sets, the need for better evaluation metrics that are capable of handling complex narratives properly, the high computational costs of most methods, and the lack of standardized benchmark tasks for user-based evaluations. In terms of open challenges, we note the need for better interaction models that allow users to explore the narrative with more control. Finally, we note that current models do not handle misleading or false content, a rising challenge as misinformation compounds with information overload to make understanding the information landscape even harder.  

As with other literature reviews, this work has some limitations related to the inclusion and exclusion of relevant pieces of work. In particular, we used the Scopus and Web of Science databases as our initial sources. Previous studies have shown that Scopus and Web of Science are inclusive and extensive sources for literature reviews \cite{falagas2008comparison}. Regardless, multiple studies were not included in our initial results and thus we had to include them through other means, such as extracting relevant citations from reviewed works. Moreover, the choice of keywords might have caused some studies that use different terminology to not show up in our searches.

\begin{acks}
This work was supported by NSF grants CNS-1915755 and DMS-1830501, ANID/Doctorado Becas Chile/2019 - 72200105, and a Virginia Tech ICTAS Junior Faculty Award received by Dr. Mitra.
\end{acks}

\bibliographystyle{ACM-Reference-Format}
\bibliography{sample-base}


\begin{thebibliography}{136}


\ifx \showCODEN    \undefined \def \showCODEN     #1{\unskip}     \fi
\ifx \showDOI      \undefined \def \showDOI       #1{#1}\fi
\ifx \showISBNx    \undefined \def \showISBNx     #1{\unskip}     \fi
\ifx \showISBNxiii \undefined \def \showISBNxiii  #1{\unskip}     \fi
\ifx \showISSN     \undefined \def \showISSN      #1{\unskip}     \fi
\ifx \showLCCN     \undefined \def \showLCCN      #1{\unskip}     \fi
\ifx \shownote     \undefined \def \shownote      #1{#1}          \fi
\ifx \showarticletitle \undefined \def \showarticletitle #1{#1}   \fi
\ifx \showURL      \undefined \def \showURL       {\relax}        \fi
\providecommand\bibfield[2]{#2}
\providecommand\bibinfo[2]{#2}
\providecommand\natexlab[1]{#1}
\providecommand\showeprint[2][]{arXiv:#2}

\bibitem[\protect\citeauthoryear{Abbott}{Abbott}{2008}]%
        {abbott2008cambridge}
\bibfield{author}{\bibinfo{person}{H~Porter Abbott}.}
  \bibinfo{year}{2008}\natexlab{}.
\newblock \bibinfo{booktitle}{\emph{The Cambridge introduction to narrative}}.
\newblock \bibinfo{publisher}{Cambridge University Press},
  \bibinfo{address}{One Liberty Plaza, New York, NY, USA}.
\newblock


\bibitem[\protect\citeauthoryear{Abu-Aisheh, Raveaux, Ramel, and
  Martineau}{Abu-Aisheh et~al\mbox{.}}{2015}]%
        {abu2015exact}
\bibfield{author}{\bibinfo{person}{Zeina Abu-Aisheh}, \bibinfo{person}{Romain
  Raveaux}, \bibinfo{person}{Jean-Yves Ramel}, {and} \bibinfo{person}{Patrick
  Martineau}.} \bibinfo{year}{2015}\natexlab{}.
\newblock \showarticletitle{{An Exact Graph Edit Distance Algorithm for Solving
  Pattern Recognition Problems}}. In \bibinfo{booktitle}{\emph{{4th Intl. Conf.
  on Pattern Recognition Applications and Methods 2015}}}.
  \bibinfo{publisher}{HAL Open Science}, \bibinfo{address}{Lisbon, Portugal},
  \bibinfo{pages}{1--9}.
\newblock


\bibitem[\protect\citeauthoryear{Alhussain and Azmi}{Alhussain and
  Azmi}{2021}]%
        {alhussain2021automatic}
\bibfield{author}{\bibinfo{person}{Arwa~I. Alhussain} {and}
  \bibinfo{person}{Aqil~M. Azmi}.} \bibinfo{year}{2021}\natexlab{}.
\newblock \showarticletitle{Automatic Story Generation: A Survey of
  Approaches}.
\newblock \bibinfo{journal}{\emph{ACM Comput. Surv.}} \bibinfo{volume}{54},
  \bibinfo{number}{5}, Article \bibinfo{articleno}{103} (\bibinfo{date}{May}
  \bibinfo{year}{2021}), \bibinfo{numpages}{38}~pages.
\newblock
\showISSN{0360-0300}


\bibitem[\protect\citeauthoryear{Allan}{Allan}{2012}]%
        {allan2012topic}
\bibfield{author}{\bibinfo{person}{James Allan}.}
  \bibinfo{year}{2012}\natexlab{}.
\newblock \bibinfo{booktitle}{\emph{Topic detection and tracking: event-based
  information organization}}. Vol.~\bibinfo{volume}{12}.
\newblock \bibinfo{publisher}{Springer Science \& Business Media},
  \bibinfo{address}{New York}.
\newblock


\bibitem[\protect\citeauthoryear{Ansah, Liu, Kang, Kwashie, Li, and Li}{Ansah
  et~al\mbox{.}}{2019}]%
        {ansah2019graph}
\bibfield{author}{\bibinfo{person}{Jeffery Ansah}, \bibinfo{person}{Lin Liu},
  \bibinfo{person}{Wei Kang}, \bibinfo{person}{Selasie Kwashie},
  \bibinfo{person}{Jixue Li}, {and} \bibinfo{person}{Jiuyong Li}.}
  \bibinfo{year}{2019}\natexlab{}.
\newblock \showarticletitle{A Graph is Worth a Thousand Words: Telling Event
  Stories Using Timeline Summarization Graphs}. In
  \bibinfo{booktitle}{\emph{The World Wide Web Conf.}}
  \emph{(\bibinfo{series}{WWW '19})}. \bibinfo{publisher}{ACM},
  \bibinfo{address}{NY, USA}, \bibinfo{pages}{2565–2571}.
\newblock
\showISBNx{9781450366748}


\bibitem[\protect\citeauthoryear{Avramidis}{Avramidis}{2013}]%
        {avramidis2013rankeval}
\bibfield{author}{\bibinfo{person}{Eleftherios Avramidis}.}
  \bibinfo{year}{2013}\natexlab{}.
\newblock \showarticletitle{RankEval: Open Tool for Evaluation of
  Machine-Learned Ranking.}
\newblock \bibinfo{journal}{\emph{Prague Bull. Math. Linguistics}}
  \bibinfo{volume}{100} (\bibinfo{year}{2013}), \bibinfo{pages}{63--72}.
\newblock


\bibitem[\protect\citeauthoryear{Baber, Andrews, Duffy, and McMaster}{Baber
  et~al\mbox{.}}{2011}]%
        {baber2011sensemaking}
\bibfield{author}{\bibinfo{person}{Chris Baber}, \bibinfo{person}{Dan Andrews},
  \bibinfo{person}{Tom Duffy}, {and} \bibinfo{person}{Richard McMaster}.}
  \bibinfo{year}{2011}\natexlab{}.
\newblock \showarticletitle{Sensemaking as narrative: Visualization for
  collaboration}.
\newblock \bibinfo{journal}{\emph{VAW2011, University London College}}
  \bibinfo{volume}{VAW2011} (\bibinfo{year}{2011}), \bibinfo{pages}{7--8}.
\newblock


\bibitem[\protect\citeauthoryear{Bahmani, Chowdhury, and Goel}{Bahmani
  et~al\mbox{.}}{2010}]%
        {bahmani2010fast}
\bibfield{author}{\bibinfo{person}{Bahman Bahmani}, \bibinfo{person}{Abdur
  Chowdhury}, {and} \bibinfo{person}{Ashish Goel}.}
  \bibinfo{year}{2010}\natexlab{}.
\newblock \bibinfo{title}{Fast Incremental and Personalized PageRank}.
\newblock
\newblock


\bibitem[\protect\citeauthoryear{Bandeli, Hussain, and Agarwal}{Bandeli
  et~al\mbox{.}}{2020}]%
        {bandeli2020framework}
\bibfield{author}{\bibinfo{person}{Kiran~Kumar Bandeli},
  \bibinfo{person}{Muhammad~Nihal Hussain}, {and} \bibinfo{person}{Nitin
  Agarwal}.} \bibinfo{year}{2020}\natexlab{}.
\newblock \showarticletitle{A Framework towards Computational Narrative
  Analysis on Blogs}. In \bibinfo{booktitle}{\emph{Text2Story@ ECIR}}.
  \bibinfo{publisher}{CEUR-WS}, \bibinfo{address}{Lisbon, Portugal},
  \bibinfo{pages}{63--69}.
\newblock


\bibitem[\protect\citeauthoryear{Binh~Tran, Alrifai, and Quoc~Nguyen}{Binh~Tran
  et~al\mbox{.}}{2013}]%
        {binh2013predicting}
\bibfield{author}{\bibinfo{person}{Giang Binh~Tran}, \bibinfo{person}{Mohammad
  Alrifai}, {and} \bibinfo{person}{Dat Quoc~Nguyen}.}
  \bibinfo{year}{2013}\natexlab{}.
\newblock \showarticletitle{Predicting Relevant News Events for Timeline
  Summaries}. In \bibinfo{booktitle}{\emph{Proceedings of the 22nd
  International Conference on World Wide Web}} (Rio de Janeiro, Brazil)
  \emph{(\bibinfo{series}{WWW '13 Companion})}. \bibinfo{publisher}{Association
  for Computing Machinery}, \bibinfo{address}{New York, NY, USA},
  \bibinfo{pages}{91–92}.
\newblock
\showISBNx{9781450320382}
\urldef\tempurl%
\url{https://doi.org/10.1145/2487788.2487829}
\showDOI{\tempurl}


\bibitem[\protect\citeauthoryear{Blei and Lafferty}{Blei and Lafferty}{2006}]%
        {blei2006dynamic}
\bibfield{author}{\bibinfo{person}{David~M. Blei} {and}
  \bibinfo{person}{John~D. Lafferty}.} \bibinfo{year}{2006}\natexlab{}.
\newblock \showarticletitle{Dynamic Topic Models}. In
  \bibinfo{booktitle}{\emph{Proc. of the 23rd Intl. Conf. on Machine
  Learning}}. \bibinfo{publisher}{ACM}, \bibinfo{address}{NY, USA},
  \bibinfo{pages}{113–120}.
\newblock
\showISBNx{1595933832}


\bibitem[\protect\citeauthoryear{Blei, Ng, and Jordan}{Blei
  et~al\mbox{.}}{2003}]%
        {blei2003latent}
\bibfield{author}{\bibinfo{person}{David~M Blei}, \bibinfo{person}{Andrew~Y
  Ng}, {and} \bibinfo{person}{Michael~I Jordan}.}
  \bibinfo{year}{2003}\natexlab{}.
\newblock \showarticletitle{Latent dirichlet allocation}.
\newblock \bibinfo{journal}{\emph{Journal of machine Learning research}}
  \bibinfo{volume}{3}, \bibinfo{number}{Jan} (\bibinfo{year}{2003}),
  \bibinfo{pages}{993--1022}.
\newblock


\bibitem[\protect\citeauthoryear{Bloom}{Bloom}{1956}]%
        {bloom1956taxonomy}
\bibfield{author}{\bibinfo{person}{Benjamin~Samuel Bloom}.}
  \bibinfo{year}{1956}\natexlab{}.
\newblock \bibinfo{booktitle}{\emph{Taxonomy of educational objectives: The
  classification of educational goals}}.
\newblock \bibinfo{publisher}{Longman}, \bibinfo{address}{Ann Arbor, Michigan}.
\newblock


\bibitem[\protect\citeauthoryear{B\"{o}gel and Gertz}{B\"{o}gel and
  Gertz}{2015}]%
        {bogel2015time}
\bibfield{author}{\bibinfo{person}{Thomas B\"{o}gel} {and}
  \bibinfo{person}{Michael Gertz}.} \bibinfo{year}{2015}\natexlab{}.
\newblock \showarticletitle{Time Will Tell: Temporal Linking of News Stories}.
  In \bibinfo{booktitle}{\emph{Proc. of the 15th ACM/IEEE-CS Joint Conf. on
  Digital Libraries}}. \bibinfo{publisher}{ACM}, \bibinfo{address}{NY, USA},
  \bibinfo{pages}{195–204}.
\newblock
\showISBNx{9781450335942}


\bibitem[\protect\citeauthoryear{Burke}{Burke}{1969}]%
        {burke1969grammar}
\bibfield{author}{\bibinfo{person}{Kenneth Burke}.}
  \bibinfo{year}{1969}\natexlab{}.
\newblock \bibinfo{booktitle}{\emph{A grammar of motives}}.
  Vol.~\bibinfo{volume}{177}.
\newblock \bibinfo{publisher}{Univ of California Press}, \bibinfo{address}{155
  Grand Ave. Suite 400. Oakland, CA}.
\newblock


\bibitem[\protect\citeauthoryear{Burns, Xiong, Franconeri, Cairo, and
  Mahyar}{Burns et~al\mbox{.}}{2020}]%
        {burns2020evaluate}
\bibfield{author}{\bibinfo{person}{Alyxander Burns}, \bibinfo{person}{Cindy
  Xiong}, \bibinfo{person}{Steven Franconeri}, \bibinfo{person}{Alberto Cairo},
  {and} \bibinfo{person}{Narges Mahyar}.} \bibinfo{year}{2020}\natexlab{}.
\newblock \showarticletitle{How to evaluate data visualizations across
  different levels of understanding}. In \bibinfo{booktitle}{\emph{2020 IEEE
  Workshop on Evaluation and Beyond - Methodological Approaches to
  Visualization}}. \bibinfo{publisher}{IEEE}, \bibinfo{address}{UT, USA},
  \bibinfo{pages}{19--28}.
\newblock


\bibitem[\protect\citeauthoryear{Cai, Xie, Lau, Li, Wong, and Wang}{Cai
  et~al\mbox{.}}{2019}]%
        {cai2019temporal}
\bibfield{author}{\bibinfo{person}{Yi Cai}, \bibinfo{person}{Haoran Xie},
  \bibinfo{person}{Raymond~YK Lau}, \bibinfo{person}{Qing Li},
  \bibinfo{person}{Tak-Lam Wong}, {and} \bibinfo{person}{Fu~Lee Wang}.}
  \bibinfo{year}{2019}\natexlab{}.
\newblock \showarticletitle{Temporal event searches based on event maps and
  relationships}.
\newblock \bibinfo{journal}{\emph{Applied soft computing}}
  \bibinfo{volume}{85} (\bibinfo{year}{2019}), \bibinfo{pages}{105750}.
\newblock


\bibitem[\protect\citeauthoryear{Camacho~Barranco, Boedihardjo, and
  Hossain}{Camacho~Barranco et~al\mbox{.}}{2019}]%
        {camacho2019analyzing}
\bibfield{author}{\bibinfo{person}{Roberto Camacho~Barranco},
  \bibinfo{person}{Arnold~P Boedihardjo}, {and} \bibinfo{person}{M~Shahriar
  Hossain}.} \bibinfo{year}{2019}\natexlab{}.
\newblock \showarticletitle{Analyzing evolving stories in news articles}.
\newblock \bibinfo{journal}{\emph{Intl. Journal of Data Science and Analytics}}
  \bibinfo{volume}{8}, \bibinfo{number}{3} (\bibinfo{year}{2019}),
  \bibinfo{pages}{241--256}.
\newblock


\bibitem[\protect\citeauthoryear{Cavazza and Pizzi}{Cavazza and Pizzi}{2006}]%
        {cavazza2006narratology}
\bibfield{author}{\bibinfo{person}{Marc Cavazza} {and} \bibinfo{person}{David
  Pizzi}.} \bibinfo{year}{2006}\natexlab{}.
\newblock \showarticletitle{Narratology for Interactive Storytelling: A
  Critical Introduction}. In \bibinfo{booktitle}{\emph{Technologies for
  Interactive Digital Storytelling and Entertainment}},
  \bibfield{editor}{\bibinfo{person}{Stefan G{\"o}bel}, \bibinfo{person}{Rainer
  Malkewitz}, {and} \bibinfo{person}{Ido Iurgel}} (Eds.).
  \bibinfo{publisher}{Springer}, \bibinfo{address}{Berlin, Heidelberg},
  \bibinfo{pages}{72--83}.
\newblock
\showISBNx{978-3-540-49935-0}


\bibitem[\protect\citeauthoryear{Chen and Chen}{Chen and Chen}{2008}]%
        {chen2008tscan}
\bibfield{author}{\bibinfo{person}{Chien~Chin Chen} {and}
  \bibinfo{person}{Meng~Chang Chen}.} \bibinfo{year}{2008}\natexlab{}.
\newblock \showarticletitle{TSCAN: A Novel Method for Topic Summarization and
  Content Anatomy}. In \bibinfo{booktitle}{\emph{Proc. of the 31st Annual Intl.
  ACM SIGIR Conf. on Research and Development in Information Retrieval}}.
  \bibinfo{publisher}{ACM}, \bibinfo{address}{NY, USA},
  \bibinfo{pages}{579–586}.
\newblock
\showISBNx{9781605581644}


\bibitem[\protect\citeauthoryear{Chen and Chen}{Chen and Chen}{2012}]%
        {chen2010tscan}
\bibfield{author}{\bibinfo{person}{Chien~Chin Chen} {and}
  \bibinfo{person}{Meng~Chang Chen}.} \bibinfo{year}{2012}\natexlab{}.
\newblock \showarticletitle{TSCAN: A content anatomy approach to temporal topic
  summarization}.
\newblock \bibinfo{journal}{\emph{IEEE transactions on Knowledge and Data
  Engineering}} \bibinfo{volume}{24}, \bibinfo{number}{1}
  (\bibinfo{year}{2012}), \bibinfo{pages}{170--183}.
\newblock


\bibitem[\protect\citeauthoryear{Chen, Chen, Sun, and Chen}{Chen
  et~al\mbox{.}}{2003}]%
        {chen2003life}
\bibfield{author}{\bibinfo{person}{Chien~Chin Chen}, \bibinfo{person}{Yao-Tsung
  Chen}, \bibinfo{person}{Yeali Sun}, {and} \bibinfo{person}{Meng~Chang Chen}.}
  \bibinfo{year}{2003}\natexlab{}.
\newblock \showarticletitle{Life Cycle Modeling of News Events Using Aging
  Theory}. In \bibinfo{booktitle}{\emph{Machine Learning: ECML 2003}},
  \bibfield{editor}{\bibinfo{person}{Nada Lavra{\v{c}}},
  \bibinfo{person}{Dragan Gamberger}, \bibinfo{person}{Hendrik Blockeel}, {and}
  \bibinfo{person}{Ljup{\v{c}}o Todorovski}} (Eds.).
  \bibinfo{publisher}{Springer}, \bibinfo{address}{Berlin, Heidelberg},
  \bibinfo{pages}{47--59}.
\newblock
\showISBNx{978-3-540-39857-8}


\bibitem[\protect\citeauthoryear{Chen, Niu, and Fu}{Chen et~al\mbox{.}}{2015}]%
        {chen2015multi}
\bibfield{author}{\bibinfo{person}{Jie Chen}, \bibinfo{person}{Zhendong Niu},
  {and} \bibinfo{person}{Hongping Fu}.} \bibinfo{year}{2015}\natexlab{}.
\newblock \showarticletitle{A Multi-news Timeline Summarization Algorithm Based
  on Aging Theory}. In \bibinfo{booktitle}{\emph{Web Technologies and
  Applications}}, \bibfield{editor}{\bibinfo{person}{Reynold Cheng},
  \bibinfo{person}{Bin Cui}, \bibinfo{person}{Zhenjie Zhang},
  \bibinfo{person}{Ruichu Cai}, {and} \bibinfo{person}{Jia Xu}} (Eds.).
  \bibinfo{publisher}{Springer}, \bibinfo{address}{Cham},
  \bibinfo{pages}{449--460}.
\newblock
\showISBNx{978-3-319-25255-1}


\bibitem[\protect\citeauthoryear{Chieu and Lee}{Chieu and Lee}{2004}]%
        {chieu2004query}
\bibfield{author}{\bibinfo{person}{Hai~Leong Chieu} {and}
  \bibinfo{person}{Yoong~Keok Lee}.} \bibinfo{year}{2004}\natexlab{}.
\newblock \showarticletitle{Query Based Event Extraction along a Timeline}. In
  \bibinfo{booktitle}{\emph{Proc. of the 27th Annual Intl. ACM SIGIR Conf. on
  Research and Development in Information Retrieval}}
  \emph{(\bibinfo{series}{SIGIR '04})}. \bibinfo{publisher}{ACM},
  \bibinfo{address}{NY, USA}, \bibinfo{pages}{425–432}.
\newblock
\showISBNx{1581138814}


\bibitem[\protect\citeauthoryear{Collins-Thompson, Rieh, Haynes, and
  Syed}{Collins-Thompson et~al\mbox{.}}{2016}]%
        {collins2016assessing}
\bibfield{author}{\bibinfo{person}{Kevyn Collins-Thompson},
  \bibinfo{person}{Soo~Young Rieh}, \bibinfo{person}{Carl~C. Haynes}, {and}
  \bibinfo{person}{Rohail Syed}.} \bibinfo{year}{2016}\natexlab{}.
\newblock \showarticletitle{Assessing Learning Outcomes in Web Search: A
  Comparison of Tasks and Query Strategies}. In \bibinfo{booktitle}{\emph{Proc.
  of the 2016 ACM on Conf. on Human Information Interaction and Retrieval}}.
  \bibinfo{publisher}{ACM}, \bibinfo{address}{NY, USA},
  \bibinfo{pages}{163–172}.
\newblock
\showISBNx{9781450337519}


\bibitem[\protect\citeauthoryear{Dennis~III}{Dennis~III}{1991}]%
        {dennis1991hyper}
\bibfield{author}{\bibinfo{person}{Samuel~Y Dennis~III}.}
  \bibinfo{year}{1991}\natexlab{}.
\newblock \showarticletitle{On the hyper-Dirichlet type 1 and hyper-Liouville
  distributions}.
\newblock \bibinfo{journal}{\emph{Communications in Statistics-Theory and
  Methods}} \bibinfo{volume}{20}, \bibinfo{number}{12} (\bibinfo{year}{1991}),
  \bibinfo{pages}{4069--4081}.
\newblock


\bibitem[\protect\citeauthoryear{Devlin, Chang, Lee, and Toutanova}{Devlin
  et~al\mbox{.}}{2018}]%
        {devlin2018bert}
\bibfield{author}{\bibinfo{person}{Jacob Devlin}, \bibinfo{person}{Ming-Wei
  Chang}, \bibinfo{person}{Kenton Lee}, {and} \bibinfo{person}{Kristina
  Toutanova}.} \bibinfo{year}{2018}\natexlab{}.
\newblock \bibinfo{title}{BERT: Pre-training of Deep Bidirectional Transformers
  for Language Understanding}.
\newblock
\newblock


\bibitem[\protect\citeauthoryear{Duan, Jatowt, and Yoshikawa}{Duan
  et~al\mbox{.}}{2020}]%
        {duan2020comparative}
\bibfield{author}{\bibinfo{person}{Yijun Duan}, \bibinfo{person}{Adam Jatowt},
  {and} \bibinfo{person}{Masatoshi Yoshikawa}.}
  \bibinfo{year}{2020}\natexlab{}.
\newblock \showarticletitle{Comparative Timeline Summarization via Dynamic
  Affinity-Preserving Random Walk}.
\newblock In \bibinfo{booktitle}{\emph{ECAI 2020}}. \bibinfo{publisher}{IOS
  Press}, \bibinfo{address}{Santiago de Compostela, Spain},
  \bibinfo{pages}{1778--1785}.
\newblock


\bibitem[\protect\citeauthoryear{Erkan and Radev}{Erkan and Radev}{2004a}]%
        {erkan2004lexpagerank}
\bibfield{author}{\bibinfo{person}{Gunes Erkan} {and} \bibinfo{person}{Dragomir
  Radev}.} \bibinfo{year}{2004}\natexlab{a}.
\newblock \showarticletitle{Lexpagerank: Prestige in multi-document text
  summarization}. In \bibinfo{booktitle}{\emph{Proc. of the 2004 Conf. on
  Empirical Methods in Natural Language Processing}}. \bibinfo{publisher}{ACL},
  \bibinfo{address}{Barcelona, Spain}, \bibinfo{pages}{365--371}.
\newblock


\bibitem[\protect\citeauthoryear{Erkan and Radev}{Erkan and Radev}{2004b}]%
        {erkan2004lexrank}
\bibfield{author}{\bibinfo{person}{G{\"u}nes Erkan} {and}
  \bibinfo{person}{Dragomir~R Radev}.} \bibinfo{year}{2004}\natexlab{b}.
\newblock \showarticletitle{Lexrank: Graph-based lexical centrality as salience
  in text summarization}.
\newblock \bibinfo{journal}{\emph{Journal of artificial intelligence research}}
   \bibinfo{volume}{22} (\bibinfo{year}{2004}), \bibinfo{pages}{457--479}.
\newblock


\bibitem[\protect\citeauthoryear{Ermakova, Cossu, and Mothe}{Ermakova
  et~al\mbox{.}}{2019}]%
        {ermakova2019survey}
\bibfield{author}{\bibinfo{person}{Liana Ermakova},
  \bibinfo{person}{Jean~Val{\`e}re Cossu}, {and} \bibinfo{person}{Josiane
  Mothe}.} \bibinfo{year}{2019}\natexlab{}.
\newblock \showarticletitle{A survey on evaluation of summarization methods}.
\newblock \bibinfo{journal}{\emph{Information processing \& management}}
  \bibinfo{volume}{56}, \bibinfo{number}{5} (\bibinfo{year}{2019}),
  \bibinfo{pages}{1794--1814}.
\newblock


\bibitem[\protect\citeauthoryear{Evangelista, Sassi, Romero, and
  Napolitano}{Evangelista et~al\mbox{.}}{2021}]%
        {evangelista2021systematic}
\bibfield{author}{\bibinfo{person}{Jo{\~a}o Rafael~Gon{\c{c}}alves
  Evangelista}, \bibinfo{person}{Renato~Jos{\'e} Sassi},
  \bibinfo{person}{M{\'a}rcio Romero}, {and} \bibinfo{person}{Domingos
  Napolitano}.} \bibinfo{year}{2021}\natexlab{}.
\newblock \showarticletitle{Systematic literature review to investigate the
  application of open source intelligence (OSINT) with artificial
  intelligence}.
\newblock \bibinfo{journal}{\emph{Journal of Applied Security Research}}
  \bibinfo{volume}{16}, \bibinfo{number}{3} (\bibinfo{year}{2021}),
  \bibinfo{pages}{345--369}.
\newblock


\bibitem[\protect\citeauthoryear{Falagas, Pitsouni, Malietzis, and
  Pappas}{Falagas et~al\mbox{.}}{2008}]%
        {falagas2008comparison}
\bibfield{author}{\bibinfo{person}{Matthew~E Falagas}, \bibinfo{person}{Eleni~I
  Pitsouni}, \bibinfo{person}{George~A Malietzis}, {and}
  \bibinfo{person}{Georgios Pappas}.} \bibinfo{year}{2008}\natexlab{}.
\newblock \showarticletitle{Comparison of PubMed, Scopus, web of science, and
  Google scholar: strengths and weaknesses}.
\newblock \bibinfo{journal}{\emph{The FASEB journal}} \bibinfo{volume}{22},
  \bibinfo{number}{2} (\bibinfo{year}{2008}), \bibinfo{pages}{338--342}.
\newblock


\bibitem[\protect\citeauthoryear{Frey and Dueck}{Frey and Dueck}{2007}]%
        {frey2007clustering}
\bibfield{author}{\bibinfo{person}{Brendan~J Frey} {and}
  \bibinfo{person}{Delbert Dueck}.} \bibinfo{year}{2007}\natexlab{}.
\newblock \showarticletitle{Clustering by passing messages between data
  points}.
\newblock \bibinfo{journal}{\emph{science}} \bibinfo{volume}{315},
  \bibinfo{number}{5814} (\bibinfo{year}{2007}), \bibinfo{pages}{972--976}.
\newblock


\bibitem[\protect\citeauthoryear{Gatt and Krahmer}{Gatt and Krahmer}{2018}]%
        {gatt2018survey}
\bibfield{author}{\bibinfo{person}{Albert Gatt} {and} \bibinfo{person}{Emiel
  Krahmer}.} \bibinfo{year}{2018}\natexlab{}.
\newblock \showarticletitle{Survey of the state of the art in natural language
  generation: Core tasks, applications and evaluation}.
\newblock \bibinfo{journal}{\emph{Journal of Artificial Intelligence Research}}
   \bibinfo{volume}{61} (\bibinfo{year}{2018}), \bibinfo{pages}{65--170}.
\newblock


\bibitem[\protect\citeauthoryear{Gerv{\'a}s, Concepci{\'o}n, Le{\'o}n,
  M{\'e}ndez, and Delatorre}{Gerv{\'a}s et~al\mbox{.}}{2019}]%
        {gervas2019long}
\bibfield{author}{\bibinfo{person}{Pablo Gerv{\'a}s}, \bibinfo{person}{Eugenio
  Concepci{\'o}n}, \bibinfo{person}{Carlos Le{\'o}n}, \bibinfo{person}{Gonzalo
  M{\'e}ndez}, {and} \bibinfo{person}{Pablo Delatorre}.}
  \bibinfo{year}{2019}\natexlab{}.
\newblock \showarticletitle{The long path to narrative generation}.
\newblock \bibinfo{journal}{\emph{IBM Journal of Research and Development}}
  \bibinfo{volume}{63}, \bibinfo{number}{1} (\bibinfo{year}{2019}),
  \bibinfo{pages}{8--1}.
\newblock


\bibitem[\protect\citeauthoryear{Ghalandari and Ifrim}{Ghalandari and
  Ifrim}{2020}]%
        {ghalandari2020examining}
\bibfield{author}{\bibinfo{person}{Demian~Gholipour Ghalandari} {and}
  \bibinfo{person}{Georgiana Ifrim}.} \bibinfo{year}{2020}\natexlab{}.
\newblock \showarticletitle{Examining the State-of-the-Art in News Timeline
  Summarization}.
\newblock \bibinfo{journal}{\emph{CoRR}}  \bibinfo{volume}{abs/2005.10107}
  (\bibinfo{year}{2020}), \bibinfo{numpages}{13}~pages.
\newblock
\showeprint[arXiv]{2005.10107}


\bibitem[\protect\citeauthoryear{Gibson}{Gibson}{2016}]%
        {gibson2016acquisition}
\bibfield{author}{\bibinfo{person}{Helen Gibson}.}
  \bibinfo{year}{2016}\natexlab{}.
\newblock \showarticletitle{Acquisition and Preparation of Data for OSINT
  Investigations}.
\newblock In \bibinfo{booktitle}{\emph{Open Source Intelligence Investigation:
  From Strategy to Implementation}}, \bibfield{editor}{\bibinfo{person}{Babak
  Akhgar}, \bibinfo{person}{P.~Saskia Bayerl}, {and} \bibinfo{person}{Fraser
  Sampson}} (Eds.). \bibinfo{publisher}{Springer}, \bibinfo{address}{Cham},
  \bibinfo{pages}{69--93}.
\newblock
\showISBNx{978-3-319-47671-1}


\bibitem[\protect\citeauthoryear{Goldstein, Mittal, Carbonell, and
  Callan}{Goldstein et~al\mbox{.}}{2000}]%
        {goldstein2000creating}
\bibfield{author}{\bibinfo{person}{Jade Goldstein}, \bibinfo{person}{Vibhu
  Mittal}, \bibinfo{person}{Jaime Carbonell}, {and} \bibinfo{person}{Jamie
  Callan}.} \bibinfo{year}{2000}\natexlab{}.
\newblock \showarticletitle{Creating and Evaluating Multi-Document Sentence
  Extract Summaries}. In \bibinfo{booktitle}{\emph{Proc. of the Ninth Intl.
  Conf. on Information and Knowledge Management}} \emph{(\bibinfo{series}{CIKM
  '00})}. \bibinfo{publisher}{ACM}, \bibinfo{address}{NY, USA},
  \bibinfo{pages}{165–172}.
\newblock
\showISBNx{1581133200}


\bibitem[\protect\citeauthoryear{Guha, Kumar, Sivakumar, and Sundaram}{Guha
  et~al\mbox{.}}{2005}]%
        {guha2005unweaving}
\bibfield{author}{\bibinfo{person}{R. Guha}, \bibinfo{person}{Ravi Kumar},
  \bibinfo{person}{D. Sivakumar}, {and} \bibinfo{person}{Ravi Sundaram}.}
  \bibinfo{year}{2005}\natexlab{}.
\newblock \showarticletitle{Unweaving a Web of Documents}. In
  \bibinfo{booktitle}{\emph{Proc. of the Eleventh ACM SIGKDD Intl. Conf. on
  Knowledge Discovery in Data Mining}} \emph{(\bibinfo{series}{KDD '05})}.
  \bibinfo{publisher}{ACM}, \bibinfo{address}{NY, USA},
  \bibinfo{pages}{574–579}.
\newblock
\showISBNx{159593135X}


\bibitem[\protect\citeauthoryear{Guo, Ding, Yao, Liang, and Yu}{Guo
  et~al\mbox{.}}{2019}]%
        {guo2019future}
\bibfield{author}{\bibinfo{person}{Bin Guo}, \bibinfo{person}{Yasan Ding},
  \bibinfo{person}{Lina Yao}, \bibinfo{person}{Yunji Liang}, {and}
  \bibinfo{person}{Zhiwen Yu}.} \bibinfo{year}{2019}\natexlab{}.
\newblock \showarticletitle{The Future of Misinformation Detection: New
  Perspectives and Trends}.
\newblock \bibinfo{journal}{\emph{CoRR}}  \bibinfo{volume}{abs/1909.03654}
  (\bibinfo{year}{2019}), \bibinfo{numpages}{23}~pages.
\newblock


\bibitem[\protect\citeauthoryear{Halverson, Corman, and Goodall}{Halverson
  et~al\mbox{.}}{2011}]%
        {halverson2011master}
\bibfield{author}{\bibinfo{person}{Jeffry Halverson}, \bibinfo{person}{Steven
  Corman}, {and} \bibinfo{person}{H~Lloyd Goodall}.}
  \bibinfo{year}{2011}\natexlab{}.
\newblock \bibinfo{booktitle}{\emph{Master narratives of Islamist extremism}}.
\newblock \bibinfo{publisher}{Springer}, \bibinfo{address}{175 5th Ave., New
  York, NY, USA}.
\newblock


\bibitem[\protect\citeauthoryear{Haveliwala}{Haveliwala}{2003}]%
        {haveliwala2003topic}
\bibfield{author}{\bibinfo{person}{Taher~H Haveliwala}.}
  \bibinfo{year}{2003}\natexlab{}.
\newblock \showarticletitle{Topic-sensitive pagerank: A context-sensitive
  ranking algorithm for web search}.
\newblock \bibinfo{journal}{\emph{IEEE transactions on knowledge and data
  engineering}} \bibinfo{volume}{15}, \bibinfo{number}{4}
  (\bibinfo{year}{2003}), \bibinfo{pages}{784--796}.
\newblock


\bibitem[\protect\citeauthoryear{Hofmann}{Hofmann}{1999}]%
        {hofmann1999probabilistic}
\bibfield{author}{\bibinfo{person}{Thomas Hofmann}.}
  \bibinfo{year}{1999}\natexlab{}.
\newblock \showarticletitle{Probabilistic Latent Semantic Indexing}. In
  \bibinfo{booktitle}{\emph{Proc. of the 22nd Annual Intl. ACM SIGIR Conf. on
  Research and Development in Information Retrieval}}
  \emph{(\bibinfo{series}{SIGIR '99})}. \bibinfo{publisher}{ACM},
  \bibinfo{address}{NY, USA}, \bibinfo{pages}{50–57}.
\newblock
\showISBNx{1581130961}


\bibitem[\protect\citeauthoryear{Hossain, Andrews, Ramakrishnan, and
  North}{Hossain et~al\mbox{.}}{2011}]%
        {hossain2011helping}
\bibfield{author}{\bibinfo{person}{Mahmud~Shahriar Hossain},
  \bibinfo{person}{Christopher Andrews}, \bibinfo{person}{Naren Ramakrishnan},
  {and} \bibinfo{person}{Chris North}.} \bibinfo{year}{2011}\natexlab{}.
\newblock \showarticletitle{Helping intelligence analysts make connections}. In
  \bibinfo{booktitle}{\emph{Workshops at the Twenty-Fifth AAAI Conf. on
  Artificial Intelligence}}. \bibinfo{publisher}{AAAI}, \bibinfo{address}{San
  Francisco, CA, USA}, \bibinfo{pages}{1--10}.
\newblock


\bibitem[\protect\citeauthoryear{Hu, Huang, Xu, Li, Usadi, and Zhu}{Hu
  et~al\mbox{.}}{2011}]%
        {hu2011generating}
\bibfield{author}{\bibinfo{person}{Po Hu}, \bibinfo{person}{Minlie Huang},
  \bibinfo{person}{Peng Xu}, \bibinfo{person}{Weichang Li},
  \bibinfo{person}{Adam~K. Usadi}, {and} \bibinfo{person}{Xiaoyan Zhu}.}
  \bibinfo{year}{2011}\natexlab{}.
\newblock \showarticletitle{Generating Breakpoint-based Timeline Overview for
  News Topic Retrospection}. In \bibinfo{booktitle}{\emph{2011 IEEE 11th Intl.
  Conf. on Data Mining}}. \bibinfo{publisher}{IEEE},
  \bibinfo{address}{Vancouver, Canada}, \bibinfo{pages}{260--269}.
\newblock


\bibitem[\protect\citeauthoryear{Hu, Huang, and Zhu}{Hu et~al\mbox{.}}{2014}]%
        {hu2014exploring}
\bibfield{author}{\bibinfo{person}{Po Hu}, \bibinfo{person}{Min-Lie Huang},
  {and} \bibinfo{person}{Xiao-Yan Zhu}.} \bibinfo{year}{2014}\natexlab{}.
\newblock \showarticletitle{Exploring the interactions of storylines from
  informative news events}.
\newblock \bibinfo{journal}{\emph{J. of Computer Science and Technology}}
  \bibinfo{volume}{29}, \bibinfo{number}{3} (\bibinfo{year}{2014}),
  \bibinfo{pages}{502--518}.
\newblock


\bibitem[\protect\citeauthoryear{Huang, Hu, Cai, and Min}{Huang
  et~al\mbox{.}}{2014}]%
        {huang2014discovering}
\bibfield{author}{\bibinfo{person}{Dongping Huang}, \bibinfo{person}{Shuyu Hu},
  \bibinfo{person}{Yi Cai}, {and} \bibinfo{person}{Huaqing Min}.}
  \bibinfo{year}{2014}\natexlab{}.
\newblock \showarticletitle{Discovering event evolution graphs based on news
  articles relationships}. In \bibinfo{booktitle}{\emph{2014 IEEE 11th Intl.
  Conf. on e-Business Engineering}}. IEEE, \bibinfo{publisher}{IEEE},
  \bibinfo{address}{Guangzhou, China}, \bibinfo{pages}{246--251}.
\newblock


\bibitem[\protect\citeauthoryear{Huang et~al\mbox{.}}{Huang
  et~al\mbox{.}}{2013}]%
        {huang2013optimized}
\bibfield{author}{\bibinfo{person}{Lifu Huang} {et~al\mbox{.}}}
  \bibinfo{year}{2013}\natexlab{}.
\newblock \showarticletitle{Optimized event storyline generation based on
  mixture-event-aspect model}. In \bibinfo{booktitle}{\emph{Proc. of the 2013
  Conf. on Empirical Methods in NLP}}. \bibinfo{publisher}{ACL},
  \bibinfo{address}{Seattle, WA, USA}, \bibinfo{pages}{726--735}.
\newblock


\bibitem[\protect\citeauthoryear{Keith~Norambuena, Horning, and
  Mitra}{Keith~Norambuena et~al\mbox{.}}{2020}]%
        {norambuenaevaluating}
\bibfield{author}{\bibinfo{person}{Brian Keith~Norambuena},
  \bibinfo{person}{Michael Horning}, {and} \bibinfo{person}{Tanushree Mitra}.}
  \bibinfo{year}{2020}\natexlab{}.
\newblock \showarticletitle{Evaluating the Inverted Pyramid Structure through
  Automatic 5W1H Extraction and Summarization}. In
  \bibinfo{booktitle}{\emph{Proc. of the 2020 Computation + Journalism
  Symposium}}. \bibinfo{publisher}{Computation + Journalism 2020},
  \bibinfo{address}{Boston, MA, USA}, \bibinfo{pages}{1--7}.
\newblock


\bibitem[\protect\citeauthoryear{Keith~Norambuena and Mitra}{Keith~Norambuena
  and Mitra}{2021}]%
        {keith2021narrative}
\bibfield{author}{\bibinfo{person}{Brian~Felipe Keith~Norambuena} {and}
  \bibinfo{person}{Tanushree Mitra}.} \bibinfo{year}{2021}\natexlab{}.
\newblock \showarticletitle{Narrative Maps: An Algorithmic Approach to
  Represent and Extract Information Narratives}.
\newblock \bibinfo{journal}{\emph{Proc. of the ACM on Human-Computer
  Interaction}} \bibinfo{volume}{4}, \bibinfo{number}{CSCW3}
  (\bibinfo{year}{2021}), \bibinfo{pages}{1--33}.
\newblock


\bibitem[\protect\citeauthoryear{Khurdiya, Dey, Raj, and Haque}{Khurdiya
  et~al\mbox{.}}{2011}]%
        {khurdiya2011multi}
\bibfield{author}{\bibinfo{person}{Arpit Khurdiya}, \bibinfo{person}{Lipika
  Dey}, \bibinfo{person}{Nidhi Raj}, {and} \bibinfo{person}{Sk~Mirajul Haque}.}
  \bibinfo{year}{2011}\natexlab{}.
\newblock \showarticletitle{Multi-perspective linking of news articles within a
  repository}. In \bibinfo{booktitle}{\emph{Twenty-Second Intl. Joint Conf. on
  Artificial Intelligence}}. \bibinfo{publisher}{AAAI},
  \bibinfo{address}{Barcelona, Spain}, \bibinfo{pages}{2281--2286}.
\newblock


\bibitem[\protect\citeauthoryear{Kullback and Leibler}{Kullback and
  Leibler}{1951}]%
        {kullback1951information}
\bibfield{author}{\bibinfo{person}{Solomon Kullback} {and}
  \bibinfo{person}{Richard~A Leibler}.} \bibinfo{year}{1951}\natexlab{}.
\newblock \showarticletitle{On information and sufficiency}.
\newblock \bibinfo{journal}{\emph{The annals of mathematical statistics}}
  \bibinfo{volume}{22}, \bibinfo{number}{1} (\bibinfo{year}{1951}),
  \bibinfo{pages}{79--86}.
\newblock


\bibitem[\protect\citeauthoryear{Kybartas and Bidarra}{Kybartas and
  Bidarra}{2016}]%
        {kybartas2016survey}
\bibfield{author}{\bibinfo{person}{Ben Kybartas} {and} \bibinfo{person}{Rafael
  Bidarra}.} \bibinfo{year}{2016}\natexlab{}.
\newblock \showarticletitle{A survey on story generation techniques for
  authoring computational narratives}.
\newblock \bibinfo{journal}{\emph{IEEE Transactions on Computational
  Intelligence and AI in Games}} \bibinfo{volume}{9}, \bibinfo{number}{3}
  (\bibinfo{year}{2016}), \bibinfo{pages}{239--253}.
\newblock


\bibitem[\protect\citeauthoryear{La~Quatra, Cagliero, Baralis, Messina, and
  Montagnuolo}{La~Quatra et~al\mbox{.}}{2021}]%
        {la2021summarize}
\bibfield{author}{\bibinfo{person}{Moreno La~Quatra}, \bibinfo{person}{Luca
  Cagliero}, \bibinfo{person}{Elena Baralis}, \bibinfo{person}{Alberto
  Messina}, {and} \bibinfo{person}{Maurizio Montagnuolo}.}
  \bibinfo{year}{2021}\natexlab{}.
\newblock \showarticletitle{Summarize Dates First: A Paradigm Shift in Timeline
  Summarization}. In \bibinfo{booktitle}{\emph{Proc. of the 44th Intl. ACM
  SIGIR Conf. on Research and Development in Information Retrieval}}.
  \bibinfo{publisher}{ACM}, \bibinfo{address}{NY, USA},
  \bibinfo{pages}{418–427}.
\newblock
\showISBNx{9781450380379}


\bibitem[\protect\citeauthoryear{Laban and Hearst}{Laban and Hearst}{2017}]%
        {laban2017newslens}
\bibfield{author}{\bibinfo{person}{Philippe Laban} {and}
  \bibinfo{person}{Marti~A Hearst}.} \bibinfo{year}{2017}\natexlab{}.
\newblock \showarticletitle{newsLens: building and visualizing long-ranging
  news stories}. In \bibinfo{booktitle}{\emph{Proc. of the Events and Stories
  in the News Workshop}}. \bibinfo{publisher}{ACL},
  \bibinfo{address}{Vancouver, Canada}, \bibinfo{pages}{1--9}.
\newblock


\bibitem[\protect\citeauthoryear{Labatut and Bost}{Labatut and Bost}{2019}]%
        {labatut2019extraction}
\bibfield{author}{\bibinfo{person}{Vincent Labatut} {and}
  \bibinfo{person}{Xavier Bost}.} \bibinfo{year}{2019}\natexlab{}.
\newblock \showarticletitle{Extraction and Analysis of Fictional Character
  Networks: A Survey}.
\newblock \bibinfo{journal}{\emph{ACM Comput. Surv.}} \bibinfo{volume}{52},
  \bibinfo{number}{5}, Article \bibinfo{articleno}{89} (\bibinfo{date}{Sept.}
  \bibinfo{year}{2019}), \bibinfo{numpages}{40}~pages.
\newblock
\showISSN{0360-0300}


\bibitem[\protect\citeauthoryear{Li and Li}{Li and Li}{2013a}]%
        {li2013evolutionary}
\bibfield{author}{\bibinfo{person}{Jiwei Li} {and} \bibinfo{person}{Sujian
  Li}.} \bibinfo{year}{2013}\natexlab{a}.
\newblock \showarticletitle{Evolutionary hierarchical dirichlet process for
  timeline summarization}. In \bibinfo{booktitle}{\emph{Proc. of the 51st
  Annual Meeting of the Association for Computational Linguistics (Volume 2:
  Short Papers)}}. \bibinfo{publisher}{ACL}, \bibinfo{address}{Sofia,
  Bulgaria}, \bibinfo{pages}{556--560}.
\newblock


\bibitem[\protect\citeauthoryear{Li and Li}{Li and Li}{2013b}]%
        {li2013empirical}
\bibfield{author}{\bibinfo{person}{Lei Li} {and} \bibinfo{person}{Tao Li}.}
  \bibinfo{year}{2013}\natexlab{b}.
\newblock \showarticletitle{An empirical study of ontology-based multi-document
  summarization in disaster management}.
\newblock \bibinfo{journal}{\emph{IEEE transactions on systems, man, and
  cybernetics: systems}} \bibinfo{volume}{44}, \bibinfo{number}{2}
  (\bibinfo{year}{2013}), \bibinfo{pages}{162--171}.
\newblock


\bibitem[\protect\citeauthoryear{Li, Wang, and Wang}{Li et~al\mbox{.}}{2015}]%
        {li2015tracking}
\bibfield{author}{\bibinfo{person}{Rumeng Li}, \bibinfo{person}{Tao Wang},
  {and} \bibinfo{person}{Xun Wang}.} \bibinfo{year}{2015}\natexlab{}.
\newblock \bibinfo{booktitle}{\emph{Tracking Events Using Time-dependent
  Hierarchical Dirichlet Tree Model}}.
\newblock \bibinfo{publisher}{SIAM}, \bibinfo{address}{Vancouver, Canada},
  \bibinfo{pages}{550--558}.
\newblock


\bibitem[\protect\citeauthoryear{Liao, Wang, and Lee}{Liao
  et~al\mbox{.}}{2021}]%
        {liao2021wilson}
\bibfield{author}{\bibinfo{person}{Yiming Liao}, \bibinfo{person}{Shuguang
  Wang}, {and} \bibinfo{person}{Dongwon Lee}.} \bibinfo{year}{2021}\natexlab{}.
\newblock \showarticletitle{WILSON: A divide and conquer approach for fast and
  effective news timeline summarization}. In \bibinfo{booktitle}{\emph{Advances
  in Database Technology - EDBT 2021}} \emph{(\bibinfo{series}{Advances in
  Database Technology - EDBT})}, \bibfield{editor}{\bibinfo{person}{Yannis
  Velegrakis}, \bibinfo{person}{Yannis Velegrakis}, \bibinfo{person}{Demetris
  Zeinalipour}, \bibinfo{person}{{Panos K.} Chrysanthis},
  \bibinfo{person}{{Panos K.} Chrysanthis}, {and} \bibinfo{person}{Francesco
  Guerra}} (Eds.). \bibinfo{publisher}{OpenProceedings.org},
  \bibinfo{address}{Nicosia, Cyprus}, \bibinfo{pages}{635--645}.
\newblock


\bibitem[\protect\citeauthoryear{Lin, Lin, Li, Wang, Chen, and Li}{Lin
  et~al\mbox{.}}{2012}]%
        {lin2012generating}
\bibfield{author}{\bibinfo{person}{Chen Lin}, \bibinfo{person}{Chun Lin},
  \bibinfo{person}{Jingxuan Li}, \bibinfo{person}{Dingding Wang},
  \bibinfo{person}{Yang Chen}, {and} \bibinfo{person}{Tao Li}.}
  \bibinfo{year}{2012}\natexlab{}.
\newblock \showarticletitle{Generating Event Storylines from Microblogs}. In
  \bibinfo{booktitle}{\emph{Proc. of the 21st ACM Intl. Conf. on Information
  and Knowledge Management}} \emph{(\bibinfo{series}{CIKM '12})}.
  \bibinfo{publisher}{ACM}, \bibinfo{address}{NY, USA},
  \bibinfo{pages}{175–184}.
\newblock
\showISBNx{9781450311564}


\bibitem[\protect\citeauthoryear{Lin, Huang, and Liang}{Lin
  et~al\mbox{.}}{2007}]%
        {lin2007individualized}
\bibfield{author}{\bibinfo{person}{Fu-ren Lin}, \bibinfo{person}{Feng-mei
  Huang}, {and} \bibinfo{person}{Chia-hao Liang}.}
  \bibinfo{year}{2007}\natexlab{}.
\newblock \showarticletitle{Individualized storyline-based news topic
  retrospection}. In \bibinfo{booktitle}{\emph{PACIS 2007 Proc.}}
  \bibinfo{publisher}{AIS}, \bibinfo{address}{Auckland, New Zealand},
  \bibinfo{pages}{140}.
\newblock


\bibitem[\protect\citeauthoryear{Lin and Liang}{Lin and Liang}{2006}]%
        {lin2006topic}
\bibfield{author}{\bibinfo{person}{Fu-ren Lin} {and} \bibinfo{person}{Chia-hao
  Liang}.} \bibinfo{year}{2006}\natexlab{}.
\newblock \showarticletitle{Topic Retrospection with Storyline-based
  Summarization on News Reports}. In \bibinfo{booktitle}{\emph{PACIS 2006
  Proc.}} \bibinfo{publisher}{AIS}, \bibinfo{address}{Kuala Lumpur, Malaysia},
  \bibinfo{numpages}{1320--1334}~pages.
\newblock


\bibitem[\protect\citeauthoryear{Lin and Liang}{Lin and Liang}{2008}]%
        {lin2008storyline}
\bibfield{author}{\bibinfo{person}{Fu-ren Lin} {and} \bibinfo{person}{Chia-Hao
  Liang}.} \bibinfo{year}{2008}\natexlab{}.
\newblock \showarticletitle{Storyline-based summarization for news topic
  retrospection}.
\newblock \bibinfo{journal}{\emph{Decision Support Systems}}
  \bibinfo{volume}{45}, \bibinfo{number}{3} (\bibinfo{year}{2008}),
  \bibinfo{pages}{473--490}.
\newblock


\bibitem[\protect\citeauthoryear{Liu, Han, Niu, Kong, Lai, and Xu}{Liu
  et~al\mbox{.}}{2020}]%
        {liu2020story}
\bibfield{author}{\bibinfo{person}{Bang Liu}, \bibinfo{person}{Fred~X Han},
  \bibinfo{person}{Di Niu}, \bibinfo{person}{Linglong Kong},
  \bibinfo{person}{Kunfeng Lai}, {and} \bibinfo{person}{Yu Xu}.}
  \bibinfo{year}{2020}\natexlab{}.
\newblock \showarticletitle{Story forest: Extracting events and telling stories
  from breaking news}.
\newblock \bibinfo{journal}{\emph{ACM Transactions on Knowledge Discovery from
  Data (TKDD)}} \bibinfo{volume}{14}, \bibinfo{number}{3}
  (\bibinfo{year}{2020}), \bibinfo{pages}{1--28}.
\newblock


\bibitem[\protect\citeauthoryear{Liu, Niu, Lai, Kong, and Xu}{Liu
  et~al\mbox{.}}{2017}]%
        {liu2017growing}
\bibfield{author}{\bibinfo{person}{Bang Liu}, \bibinfo{person}{Di Niu},
  \bibinfo{person}{Kunfeng Lai}, \bibinfo{person}{Linglong Kong}, {and}
  \bibinfo{person}{Yu Xu}.} \bibinfo{year}{2017}\natexlab{}.
\newblock \showarticletitle{Growing Story Forest Online from Massive Breaking
  News}. In \bibinfo{booktitle}{\emph{Proc. of the 2017 ACM on Conf. on
  Information and Knowledge Management}}. \bibinfo{publisher}{ACM},
  \bibinfo{address}{NY, USA}, \bibinfo{pages}{777–785}.
\newblock
\showISBNx{9781450349185}


\bibitem[\protect\citeauthoryear{Luhn}{Luhn}{1958}]%
        {luhn1958automatic}
\bibfield{author}{\bibinfo{person}{Hans~Peter Luhn}.}
  \bibinfo{year}{1958}\natexlab{}.
\newblock \showarticletitle{The automatic creation of literature abstracts}.
\newblock \bibinfo{journal}{\emph{IBM Journal of research and development}}
  \bibinfo{volume}{2}, \bibinfo{number}{2} (\bibinfo{year}{1958}),
  \bibinfo{pages}{159--165}.
\newblock


\bibitem[\protect\citeauthoryear{Mani}{Mani}{2012}]%
        {mani2012computational}
\bibfield{author}{\bibinfo{person}{Inderjeet Mani}.}
  \bibinfo{year}{2012}\natexlab{}.
\newblock \showarticletitle{Computational modeling of narrative}.
\newblock \bibinfo{journal}{\emph{Synthesis Lectures on Human Language
  Technologies}} \bibinfo{volume}{5}, \bibinfo{number}{3}
  (\bibinfo{year}{2012}), \bibinfo{pages}{1--142}.
\newblock


\bibitem[\protect\citeauthoryear{Mei, Guo, and Radev}{Mei
  et~al\mbox{.}}{2010}]%
        {mei2010divrank}
\bibfield{author}{\bibinfo{person}{Qiaozhu Mei}, \bibinfo{person}{Jian Guo},
  {and} \bibinfo{person}{Dragomir Radev}.} \bibinfo{year}{2010}\natexlab{}.
\newblock \showarticletitle{DivRank: The Interplay of Prestige and Diversity in
  Information Networks}. In \bibinfo{booktitle}{\emph{Proc. of the 16th ACM
  SIGKDD Intl. Conf. on Knowledge Discovery and Data Mining}}
  \emph{(\bibinfo{series}{KDD '10})}. \bibinfo{publisher}{ACM},
  \bibinfo{address}{NY, USA}, \bibinfo{pages}{1009–1018}.
\newblock
\showISBNx{9781450300551}


\bibitem[\protect\citeauthoryear{Mei and Zhai}{Mei and Zhai}{2005}]%
        {mei2005discovering}
\bibfield{author}{\bibinfo{person}{Qiaozhu Mei} {and}
  \bibinfo{person}{ChengXiang Zhai}.} \bibinfo{year}{2005}\natexlab{}.
\newblock \showarticletitle{Discovering Evolutionary Theme Patterns from Text:
  An Exploration of Temporal Text Mining}. In \bibinfo{booktitle}{\emph{Proc.
  of the Eleventh ACM SIGKDD Intl. Conf. on Knowledge Discovery in Data
  Mining}} \emph{(\bibinfo{series}{KDD '05})}. \bibinfo{publisher}{ACM},
  \bibinfo{address}{NY, USA}, \bibinfo{pages}{198–207}.
\newblock
\showISBNx{159593135X}


\bibitem[\protect\citeauthoryear{Mihalcea, Corley, Strapparava,
  et~al\mbox{.}}{Mihalcea et~al\mbox{.}}{2006}]%
        {mihalcea2006corpus}
\bibfield{author}{\bibinfo{person}{Rada Mihalcea}, \bibinfo{person}{Courtney
  Corley}, \bibinfo{person}{Carlo Strapparava}, {et~al\mbox{.}}}
  \bibinfo{year}{2006}\natexlab{}.
\newblock \showarticletitle{Corpus-based and knowledge-based measures of text
  semantic similarity}. In \bibinfo{booktitle}{\emph{AAAI}},
  Vol.~\bibinfo{volume}{6}. \bibinfo{publisher}{AAAI},
  \bibinfo{address}{Boston, MA, USA}, \bibinfo{pages}{775--780}.
\newblock


\bibitem[\protect\citeauthoryear{Mihalcea and Tarau}{Mihalcea and
  Tarau}{2004}]%
        {mihalcea2004textrank}
\bibfield{author}{\bibinfo{person}{Rada Mihalcea} {and} \bibinfo{person}{Paul
  Tarau}.} \bibinfo{year}{2004}\natexlab{}.
\newblock \showarticletitle{Textrank: Bringing order into text}. In
  \bibinfo{booktitle}{\emph{Proc. of the 2004 Conf. on empirical methods in
  natural language processing}}. \bibinfo{publisher}{ACL},
  \bibinfo{address}{Barcelona, Spain}, \bibinfo{pages}{404--411}.
\newblock


\bibitem[\protect\citeauthoryear{Minard, Speranza, Agirre, Aldabe, van Erp,
  Magnini, Rigau, and Urizar}{Minard et~al\mbox{.}}{2015}]%
        {minard2015semeval}
\bibfield{author}{\bibinfo{person}{Anne-Lyse~Myriam Minard},
  \bibinfo{person}{Manuela Speranza}, \bibinfo{person}{Eneko Agirre},
  \bibinfo{person}{Itziar Aldabe}, \bibinfo{person}{Marieke van Erp},
  \bibinfo{person}{Bernardo Magnini}, \bibinfo{person}{German Rigau}, {and}
  \bibinfo{person}{Ruben Urizar}.} \bibinfo{year}{2015}\natexlab{}.
\newblock \showarticletitle{Semeval-2015 task 4: Timeline: Cross-document event
  ordering}. In \bibinfo{booktitle}{\emph{9th Intl. workshop on semantic
  evaluation}}. \bibinfo{publisher}{ACL}, \bibinfo{address}{CO, USA},
  \bibinfo{pages}{778--786}.
\newblock


\bibitem[\protect\citeauthoryear{Miskimmon, O'loughlin, and Roselle}{Miskimmon
  et~al\mbox{.}}{2014}]%
        {miskimmon2014strategic}
\bibfield{author}{\bibinfo{person}{Alister Miskimmon}, \bibinfo{person}{Ben
  O'loughlin}, {and} \bibinfo{person}{Laura Roselle}.}
  \bibinfo{year}{2014}\natexlab{}.
\newblock \bibinfo{booktitle}{\emph{Strategic narratives: Communication power
  and the new world order}}.
\newblock \bibinfo{publisher}{Routledge}, \bibinfo{address}{711 3rd Ave. \#8,
  New York, NY, USA}.
\newblock


\bibitem[\protect\citeauthoryear{Nallapati, Feng, Peng, and Allan}{Nallapati
  et~al\mbox{.}}{2004}]%
        {nallapati2004event}
\bibfield{author}{\bibinfo{person}{Ramesh Nallapati}, \bibinfo{person}{Ao
  Feng}, \bibinfo{person}{Fuchun Peng}, {and} \bibinfo{person}{James Allan}.}
  \bibinfo{year}{2004}\natexlab{}.
\newblock \showarticletitle{Event Threading within News Topics}. In
  \bibinfo{booktitle}{\emph{Proc. of the Thirteenth ACM Intl. Conf. on
  Information and Knowledge Management}} \emph{(\bibinfo{series}{CIKM '04})}.
  \bibinfo{publisher}{ACM}, \bibinfo{address}{NY, USA},
  \bibinfo{pages}{446–453}.
\newblock
\showISBNx{1581138741}


\bibitem[\protect\citeauthoryear{Nguyen, Tannier, and Moriceau}{Nguyen
  et~al\mbox{.}}{2014}]%
        {nguyen2014ranking}
\bibfield{author}{\bibinfo{person}{Kiem-Hieu Nguyen}, \bibinfo{person}{Xavier
  Tannier}, {and} \bibinfo{person}{V{\'e}ronique Moriceau}.}
  \bibinfo{year}{2014}\natexlab{}.
\newblock \showarticletitle{{Ranking Multidocument Event Descriptions for
  Building Thematic Timelines}}. In \bibinfo{booktitle}{\emph{{COLING 2014, the
  25th Intl. Conf. on Computational Linguistic}}}. \bibinfo{publisher}{HAL Open
  Science}, \bibinfo{address}{Dublin, Ireland}, \bibinfo{pages}{1208 -- 1217}.
\newblock


\bibitem[\protect\citeauthoryear{North, Saraiya, and Duca}{North
  et~al\mbox{.}}{2011}]%
        {north2011comparison}
\bibfield{author}{\bibinfo{person}{Chris North}, \bibinfo{person}{Purvi
  Saraiya}, {and} \bibinfo{person}{Karen Duca}.}
  \bibinfo{year}{2011}\natexlab{}.
\newblock \showarticletitle{A comparison of benchmark task and insight
  evaluation methods for information visualization}.
\newblock \bibinfo{journal}{\emph{Information Visualization}}
  \bibinfo{volume}{10}, \bibinfo{number}{3} (\bibinfo{year}{2011}),
  \bibinfo{pages}{162--181}.
\newblock


\bibitem[\protect\citeauthoryear{Ofek, Dar{\'a}nyi, and Rokach}{Ofek
  et~al\mbox{.}}{2013}]%
        {ofek2013linking}
\bibfield{author}{\bibinfo{person}{Nir Ofek}, \bibinfo{person}{S{\'a}ndor
  Dar{\'a}nyi}, {and} \bibinfo{person}{Lior Rokach}.}
  \bibinfo{year}{2013}\natexlab{}.
\newblock \showarticletitle{Linking motif sequences with tale types by machine
  learning}. In \bibinfo{booktitle}{\emph{2013 Workshop on Computational Models
  of Narrative}}. Schloss Dagstuhl-Leibniz-Zentrum fuer Informatik,
  \bibinfo{publisher}{Dagstuhl Publishing}, \bibinfo{address}{Germany},
  \bibinfo{pages}{166--182}.
\newblock


\bibitem[\protect\citeauthoryear{Ogata}{Ogata}{2016}]%
        {ogata2016computational}
\bibfield{author}{\bibinfo{person}{Takashi Ogata}.}
  \bibinfo{year}{2016}\natexlab{}.
\newblock \showarticletitle{Computational and cognitive approaches to
  narratology from the perspective of narrative generation}.
\newblock In \bibinfo{booktitle}{\emph{Computational and cognitive approaches
  to narratology}}. \bibinfo{publisher}{IGI Global}, \bibinfo{address}{Hershey,
  PA, USA}, \bibinfo{pages}{1--74}.
\newblock


\bibitem[\protect\citeauthoryear{Ogawa, Mori, and Toyama}{Ogawa
  et~al\mbox{.}}{2012}]%
        {ogawa2011recall}
\bibfield{author}{\bibinfo{person}{Yasuhiro Ogawa}, \bibinfo{person}{Masaki
  Mori}, {and} \bibinfo{person}{Katsuhiko Toyama}.}
  \bibinfo{year}{2012}\natexlab{}.
\newblock \showarticletitle{Recall-Oriented Evaluation Metrics for Consistent
  Translation of Japanese Legal Sentences}. In \bibinfo{booktitle}{\emph{New
  Frontiers in Artificial Intelligence}},
  \bibfield{editor}{\bibinfo{person}{Manabu Okumura}, \bibinfo{person}{Daisuke
  Bekki}, {and} \bibinfo{person}{Ken Satoh}} (Eds.).
  \bibinfo{publisher}{Springer}, \bibinfo{address}{Berlin, Heidelberg},
  \bibinfo{pages}{141--154}.
\newblock
\showISBNx{978-3-642-32090-3}


\bibitem[\protect\citeauthoryear{Page, Brin, Motwani, and Winograd}{Page
  et~al\mbox{.}}{1999}]%
        {page1999pagerank}
\bibfield{author}{\bibinfo{person}{Lawrence Page}, \bibinfo{person}{Sergey
  Brin}, \bibinfo{person}{Rajeev Motwani}, {and} \bibinfo{person}{Terry
  Winograd}.} \bibinfo{year}{1999}\natexlab{}.
\newblock \bibinfo{booktitle}{\emph{The PageRank citation ranking: Bringing
  order to the web.}}
\newblock \bibinfo{type}{{T}echnical {R}eport}. \bibinfo{institution}{Stanford
  InfoLab}.
\newblock


\bibitem[\protect\citeauthoryear{Pasquali, Campos, Ribeiro, Santana, Jorge, and
  Jatowt}{Pasquali et~al\mbox{.}}{2021}]%
        {pasquali2021tls}
\bibfield{author}{\bibinfo{person}{Arian Pasquali}, \bibinfo{person}{Ricardo
  Campos}, \bibinfo{person}{Alexandre Ribeiro}, \bibinfo{person}{Brenda
  Santana}, \bibinfo{person}{Al{\'i}pio Jorge}, {and} \bibinfo{person}{Adam
  Jatowt}.} \bibinfo{year}{2021}\natexlab{}.
\newblock \showarticletitle{TLS-Covid19: A New Annotated Corpus for Timeline
  Summarization}. In \bibinfo{booktitle}{\emph{Advances in Information
  Retrieval}}, \bibfield{editor}{\bibinfo{person}{Djoerd Hiemstra},
  \bibinfo{person}{Marie-Francine Moens}, \bibinfo{person}{Josiane Mothe},
  \bibinfo{person}{Raffaele Perego}, \bibinfo{person}{Martin Potthast}, {and}
  \bibinfo{person}{Fabrizio Sebastiani}} (Eds.). \bibinfo{publisher}{Springer
  International Publishing}, \bibinfo{address}{Cham},
  \bibinfo{pages}{497--512}.
\newblock
\showISBNx{978-3-030-72113-8}


\bibitem[\protect\citeauthoryear{Pemantle}{Pemantle}{1992}]%
        {pemantle1992vertex}
\bibfield{author}{\bibinfo{person}{Robin Pemantle}.}
  \bibinfo{year}{1992}\natexlab{}.
\newblock \showarticletitle{Vertex-reinforced random walk}.
\newblock \bibinfo{journal}{\emph{Probability Theory and Related Fields}}
  \bibinfo{volume}{92}, \bibinfo{number}{1} (\bibinfo{year}{1992}),
  \bibinfo{pages}{117--136}.
\newblock


\bibitem[\protect\citeauthoryear{Puckett}{Puckett}{2016}]%
        {puckett2016narrative}
\bibfield{author}{\bibinfo{person}{Kent Puckett}.}
  \bibinfo{year}{2016}\natexlab{}.
\newblock \bibinfo{booktitle}{\emph{Narrative theory}}.
\newblock \bibinfo{publisher}{Cambridge University Press},
  \bibinfo{address}{One Liberty Plaza, New York, NY, USA}.
\newblock


\bibitem[\protect\citeauthoryear{Pustejovsky, Castano, Ingria, Sauri,
  Gaizauskas, Setzer, Katz, and Radev}{Pustejovsky et~al\mbox{.}}{2003}]%
        {pustejovsky2003timeml}
\bibfield{author}{\bibinfo{person}{James Pustejovsky},
  \bibinfo{person}{Jos{\'e}~M Castano}, \bibinfo{person}{Robert Ingria},
  \bibinfo{person}{Roser Sauri}, \bibinfo{person}{Robert~J Gaizauskas},
  \bibinfo{person}{Andrea Setzer}, \bibinfo{person}{Graham Katz}, {and}
  \bibinfo{person}{Dragomir~R Radev}.} \bibinfo{year}{2003}\natexlab{}.
\newblock \showarticletitle{TimeML: Robust specification of event and temporal
  expressions in text.}
\newblock \bibinfo{journal}{\emph{New directions in question answering}}
  \bibinfo{volume}{3} (\bibinfo{year}{2003}), \bibinfo{pages}{28--34}.
\newblock


\bibitem[\protect\citeauthoryear{Qiu, Li, Qiao, Li, and Zhu}{Qiu
  et~al\mbox{.}}{2008}]%
        {qiu2008timeline}
\bibfield{author}{\bibinfo{person}{Jiangtao Qiu}, \bibinfo{person}{Chuan Li},
  \bibinfo{person}{Shaojie Qiao}, \bibinfo{person}{Taiyong Li}, {and}
  \bibinfo{person}{Jun Zhu}.} \bibinfo{year}{2008}\natexlab{}.
\newblock \showarticletitle{Timeline Analysis of Web News Events}. In
  \bibinfo{booktitle}{\emph{Advanced Data Mining and Applications}},
  \bibfield{editor}{\bibinfo{person}{Changjie Tang},
  \bibinfo{person}{Charles~X. Ling}, \bibinfo{person}{Xiaofang Zhou},
  \bibinfo{person}{Nick~J. Cercone}, {and} \bibinfo{person}{Xue Li}} (Eds.).
  \bibinfo{publisher}{Springer}, \bibinfo{address}{Berlin, Heidelberg},
  \bibinfo{pages}{123--134}.
\newblock
\showISBNx{978-3-540-88192-6}


\bibitem[\protect\citeauthoryear{Qiu and Tang}{Qiu and Tang}{2007}]%
        {qiu2007topic}
\bibfield{author}{\bibinfo{person}{J Qiu} {and} \bibinfo{person}{C Tang}.}
  \bibinfo{year}{2007}\natexlab{}.
\newblock \showarticletitle{Topic oriented semi-supervised document
  clustering}. In \bibinfo{booktitle}{\emph{Proc. of SIGMOD 2007 Workshop
  IDAR}}. \bibinfo{publisher}{DISC}, \bibinfo{address}{Beijing, China},
  \bibinfo{pages}{57--62}.
\newblock


\bibitem[\protect\citeauthoryear{Qiu, Tang, Zeng, Qiao, Zuo, Chen, and Zhu}{Qiu
  et~al\mbox{.}}{2007}]%
        {qiu2007novel}
\bibfield{author}{\bibinfo{person}{Jiangtao Qiu}, \bibinfo{person}{Changjie
  Tang}, \bibinfo{person}{Tao Zeng}, \bibinfo{person}{Shaojie Qiao},
  \bibinfo{person}{Jie Zuo}, \bibinfo{person}{Peng Chen}, {and}
  \bibinfo{person}{Jun Zhu}.} \bibinfo{year}{2007}\natexlab{}.
\newblock \showarticletitle{A Novel Text Classification Approach Based on
  Enhanced Association Rule}. In \bibinfo{booktitle}{\emph{Advanced Data Mining
  and Applications}}, \bibfield{editor}{\bibinfo{person}{Reda Alhajj},
  \bibinfo{person}{Hong Gao}, \bibinfo{person}{Jianzhong Li},
  \bibinfo{person}{Xue Li}, {and} \bibinfo{person}{Osmar~R. Za{\"i}ane}}
  (Eds.). \bibinfo{publisher}{Springer}, \bibinfo{address}{Berlin, Heidelberg},
  \bibinfo{pages}{252--263}.
\newblock
\showISBNx{978-3-540-73871-8}


\bibitem[\protect\citeauthoryear{Reimers and Gurevych}{Reimers and
  Gurevych}{2019}]%
        {reimers2019sentence}
\bibfield{author}{\bibinfo{person}{Nils Reimers} {and} \bibinfo{person}{Iryna
  Gurevych}.} \bibinfo{year}{2019}\natexlab{}.
\newblock \showarticletitle{Sentence-BERT: Sentence Embeddings using Siamese
  BERT-Networks}.
\newblock \bibinfo{journal}{\emph{CoRR}}  \bibinfo{volume}{abs/1908.10084}
  (\bibinfo{year}{2019}), \bibinfo{numpages}{11}~pages.
\newblock
\showeprint[arXiv]{1908.10084}


\bibitem[\protect\citeauthoryear{Richards, Winston, Finlayson,
  et~al\mbox{.}}{Richards et~al\mbox{.}}{2009}]%
        {richards2009advancing}
\bibfield{author}{\bibinfo{person}{Whitman Richards},
  \bibinfo{person}{Patrick~Henry Winston}, \bibinfo{person}{Mark~Alan
  Finlayson}, {et~al\mbox{.}}} \bibinfo{year}{2009}\natexlab{}.
\newblock \bibinfo{booktitle}{\emph{Advancing computational models of
  narrative}}.
\newblock \bibinfo{type}{{T}echnical {R}eport}.
  \bibinfo{institution}{MIT-CSAIL}.
\newblock


\bibitem[\protect\citeauthoryear{Rigsby and Barbar{\'a}}{Rigsby and
  Barbar{\'a}}{2018}]%
        {rigsby2018storytelling}
\bibfield{author}{\bibinfo{person}{J.~T. Rigsby} {and} \bibinfo{person}{Daniel
  Barbar{\'a}}.} \bibinfo{year}{2018}\natexlab{}.
\newblock \showarticletitle{Storytelling with Signal Injection: Focusing
  Stories with Domain Knowledge}. In \bibinfo{booktitle}{\emph{Machine Learning
  and Data Mining in Pattern Recognition}},
  \bibfield{editor}{\bibinfo{person}{Petra Perner}} (Ed.).
  \bibinfo{publisher}{Springer}, \bibinfo{address}{Cham},
  \bibinfo{pages}{425--439}.
\newblock
\showISBNx{978-3-319-96133-0}


\bibitem[\protect\citeauthoryear{Rosenberg and Hirschberg}{Rosenberg and
  Hirschberg}{2007}]%
        {rosenberg2007v}
\bibfield{author}{\bibinfo{person}{Andrew Rosenberg} {and}
  \bibinfo{person}{Julia Hirschberg}.} \bibinfo{year}{2007}\natexlab{}.
\newblock \showarticletitle{V-measure: A conditional entropy-based external
  cluster evaluation measure}. In \bibinfo{booktitle}{\emph{Proc. of the 2007
  joint Conf. on empirical methods in natural language processing and
  computational natural language learning}}. \bibinfo{publisher}{ACL},
  \bibinfo{address}{Prague, Czech Republic}, \bibinfo{pages}{410--420}.
\newblock


\bibitem[\protect\citeauthoryear{Ryan}{Ryan}{2017}]%
        {ryan2017grimes}
\bibfield{author}{\bibinfo{person}{James Ryan}.}
  \bibinfo{year}{2017}\natexlab{}.
\newblock \showarticletitle{Grimes' Fairy Tales: A 1960s Story Generator}. In
  \bibinfo{booktitle}{\emph{Interactive Storytelling}},
  \bibfield{editor}{\bibinfo{person}{Nuno Nunes}, \bibinfo{person}{Ian Oakley},
  {and} \bibinfo{person}{Valentina Nisi}} (Eds.).
  \bibinfo{publisher}{Springer}, \bibinfo{address}{Cham},
  \bibinfo{pages}{89--103}.
\newblock
\showISBNx{978-3-319-71027-3}


\bibitem[\protect\citeauthoryear{Schlachter, Ruvinsky, Reynoso, Muthiah, and
  Ramakrishnan}{Schlachter et~al\mbox{.}}{2015}]%
        {schlachter2015leveraging}
\bibfield{author}{\bibinfo{person}{Jason Schlachter}, \bibinfo{person}{Alicia
  Ruvinsky}, \bibinfo{person}{Luis~Asencios Reynoso},
  \bibinfo{person}{Sathappan Muthiah}, {and} \bibinfo{person}{Naren
  Ramakrishnan}.} \bibinfo{year}{2015}\natexlab{}.
\newblock \showarticletitle{Leveraging topic models to develop metrics for
  evaluating the quality of narrative threads extracted from news stories}.
\newblock \bibinfo{journal}{\emph{Procedia Manufacturing}}  \bibinfo{volume}{3}
  (\bibinfo{year}{2015}), \bibinfo{pages}{4028--4035}.
\newblock


\bibitem[\protect\citeauthoryear{Shahaf and Guestrin}{Shahaf and
  Guestrin}{2010}]%
        {shahaf2010connecting}
\bibfield{author}{\bibinfo{person}{Dafna Shahaf} {and} \bibinfo{person}{Carlos
  Guestrin}.} \bibinfo{year}{2010}\natexlab{}.
\newblock \showarticletitle{Connecting the Dots between News Articles}. In
  \bibinfo{booktitle}{\emph{Proc. of the 16th ACM SIGKDD Intl. Conf. on
  Knowledge Discovery and Data Mining}}. \bibinfo{publisher}{ACM},
  \bibinfo{address}{NY, USA}, \bibinfo{pages}{623–632}.
\newblock
\showISBNx{9781450300551}


\bibitem[\protect\citeauthoryear{Shahaf and Guestrin}{Shahaf and
  Guestrin}{2012}]%
        {shahaf2012connecting}
\bibfield{author}{\bibinfo{person}{Dafna Shahaf} {and} \bibinfo{person}{Carlos
  Guestrin}.} \bibinfo{year}{2012}\natexlab{}.
\newblock \showarticletitle{Connecting two (or less) dots: Discovering
  structure in news articles}.
\newblock \bibinfo{journal}{\emph{ACM Trans. on Knowledge Discovery from Data
  (TKDD)}} \bibinfo{volume}{5}, \bibinfo{number}{4} (\bibinfo{year}{2012}),
  \bibinfo{pages}{1--31}.
\newblock


\bibitem[\protect\citeauthoryear{Shahaf, Guestrin, and Horvitz}{Shahaf
  et~al\mbox{.}}{2012}]%
        {shahaf2012trains}
\bibfield{author}{\bibinfo{person}{Dafna Shahaf}, \bibinfo{person}{Carlos
  Guestrin}, {and} \bibinfo{person}{Eric Horvitz}.}
  \bibinfo{year}{2012}\natexlab{}.
\newblock \showarticletitle{Trains of Thought: Generating Information Maps}. In
  \bibinfo{booktitle}{\emph{Proc. of the 21st Intl. Conf. on World Wide Web}}.
  \bibinfo{publisher}{ACM}, \bibinfo{address}{NY, USA},
  \bibinfo{pages}{899–908}.
\newblock
\showISBNx{9781450312295}


\bibitem[\protect\citeauthoryear{Shahaf, Guestrin, and Horvitz}{Shahaf
  et~al\mbox{.}}{2013a}]%
        {shahaf2013metro}
\bibfield{author}{\bibinfo{person}{Dafna Shahaf}, \bibinfo{person}{Carlos
  Guestrin}, {and} \bibinfo{person}{Eric Horvitz}.}
  \bibinfo{year}{2013}\natexlab{a}.
\newblock \showarticletitle{Metro Maps of Information}.
\newblock \bibinfo{journal}{\emph{SIGWEB Newsletter}} \bibinfo{volume}{April
  2013}, \bibinfo{number}{Spring}, Article \bibinfo{articleno}{4}
  (\bibinfo{date}{April} \bibinfo{year}{2013}), \bibinfo{numpages}{9}~pages.
\newblock
\showISSN{1931-1745}


\bibitem[\protect\citeauthoryear{Shahaf, Guestrin, Horvitz, and
  Leskovec}{Shahaf et~al\mbox{.}}{2015}]%
        {shahaf2015information}
\bibfield{author}{\bibinfo{person}{Dafna Shahaf}, \bibinfo{person}{Carlos
  Guestrin}, \bibinfo{person}{Eric Horvitz}, {and} \bibinfo{person}{Jure
  Leskovec}.} \bibinfo{year}{2015}\natexlab{}.
\newblock \showarticletitle{Information cartography}.
\newblock \bibinfo{journal}{\emph{Commun. ACM}} \bibinfo{volume}{58},
  \bibinfo{number}{11} (\bibinfo{year}{2015}), \bibinfo{pages}{62--73}.
\newblock


\bibitem[\protect\citeauthoryear{Shahaf, Yang, Suen, Jacobs, Wang, and
  Leskovec}{Shahaf et~al\mbox{.}}{2013b}]%
        {shahaf2013information}
\bibfield{author}{\bibinfo{person}{Dafna Shahaf}, \bibinfo{person}{Jaewon
  Yang}, \bibinfo{person}{Caroline Suen}, \bibinfo{person}{Jeff Jacobs},
  \bibinfo{person}{Heidi Wang}, {and} \bibinfo{person}{Jure Leskovec}.}
  \bibinfo{year}{2013}\natexlab{b}.
\newblock \showarticletitle{Information Cartography: Creating Zoomable,
  Large-Scale Maps of Information}. In \bibinfo{booktitle}{\emph{Proc. of the
  19th ACM SIGKDD Intl. Conf. on Knowledge Discovery and Data Mining}}.
  \bibinfo{publisher}{ACM}, \bibinfo{address}{NY, USA},
  \bibinfo{pages}{1097–1105}.
\newblock
\showISBNx{9781450321747}


\bibitem[\protect\citeauthoryear{Shen and Li}{Shen and Li}{2010}]%
        {shen2010multi}
\bibfield{author}{\bibinfo{person}{Chao Shen} {and} \bibinfo{person}{Tao Li}.}
  \bibinfo{year}{2010}\natexlab{}.
\newblock \showarticletitle{Multi-document summarization via the minimum
  dominating set}. In \bibinfo{booktitle}{\emph{Proc. of the 23rd Intl. Conf.
  on Computational Linguistics (Coling 2010)}}. \bibinfo{publisher}{ACL},
  \bibinfo{address}{Beijing, China}, \bibinfo{pages}{984--992}.
\newblock


\bibitem[\protect\citeauthoryear{Swan and Allan}{Swan and Allan}{2000}]%
        {swan2000automatic}
\bibfield{author}{\bibinfo{person}{Russell Swan} {and} \bibinfo{person}{James
  Allan}.} \bibinfo{year}{2000}\natexlab{}.
\newblock \showarticletitle{Automatic Generation of Overview Timelines}. In
  \bibinfo{booktitle}{\emph{Proc. of the 23rd Annual Intl. ACM SIGIR Conf. on
  Research and Development in Information Retrieval}}
  \emph{(\bibinfo{series}{SIGIR '00})}. \bibinfo{publisher}{ACM},
  \bibinfo{address}{NY, USA}, \bibinfo{pages}{49–56}.
\newblock
\showISBNx{1581132263}


\bibitem[\protect\citeauthoryear{Swanson}{Swanson}{1991}]%
        {swanson1991complementary}
\bibfield{author}{\bibinfo{person}{Don~R. Swanson}.}
  \bibinfo{year}{1991}\natexlab{}.
\newblock \showarticletitle{Complementary Structures in Disjoint Science
  Literatures}. In \bibinfo{booktitle}{\emph{Proc. of the 14th Annual Intl. ACM
  SIGIR Conf. on Research and Development in Information Retrieval}}
  \emph{(\bibinfo{series}{SIGIR '91})}. \bibinfo{publisher}{ACM},
  \bibinfo{address}{NY, USA}, \bibinfo{pages}{280–289}.
\newblock
\showISBNx{0897914481}


\bibitem[\protect\citeauthoryear{Szostek}{Szostek}{2017}]%
        {szostek2017defence}
\bibfield{author}{\bibinfo{person}{Joanna Szostek}.}
  \bibinfo{year}{2017}\natexlab{}.
\newblock \showarticletitle{Defence and Promotion of Desired State Identity in
  Russia’s Strategic Narrative}.
\newblock \bibinfo{journal}{\emph{Geopolitics}} \bibinfo{volume}{22},
  \bibinfo{number}{3} (\bibinfo{year}{2017}), \bibinfo{pages}{571--593}.
\newblock


\bibitem[\protect\citeauthoryear{Tannier and Moriceau}{Tannier and
  Moriceau}{2013}]%
        {tannier2013building}
\bibfield{author}{\bibinfo{person}{Xavier Tannier} {and}
  \bibinfo{person}{V{\'e}ronique Moriceau}.} \bibinfo{year}{2013}\natexlab{}.
\newblock \showarticletitle{Building event threads out of multiple news
  articles}. In \bibinfo{booktitle}{\emph{Proc. of the 2013 Conf. on Empirical
  Methods in NLP}}. \bibinfo{publisher}{ACL}, \bibinfo{address}{Seattle, WA,
  USA}, \bibinfo{pages}{958--967}.
\newblock


\bibitem[\protect\citeauthoryear{Teh, Jordan, Beal, and Blei}{Teh
  et~al\mbox{.}}{2006}]%
        {teh2006hierarchical}
\bibfield{author}{\bibinfo{person}{Yee~Whye Teh}, \bibinfo{person}{Michael~I
  Jordan}, \bibinfo{person}{Matthew~J Beal}, {and} \bibinfo{person}{David~M
  Blei}.} \bibinfo{year}{2006}\natexlab{}.
\newblock \showarticletitle{Hierarchical dirichlet processes}.
\newblock \bibinfo{journal}{\emph{Journal of the american statistical
  association}} \bibinfo{volume}{101}, \bibinfo{number}{476}
  (\bibinfo{year}{2006}), \bibinfo{pages}{1566--1581}.
\newblock


\bibitem[\protect\citeauthoryear{Tikhomirov and Dobrov}{Tikhomirov and
  Dobrov}{2018}]%
        {tikhomirov2017news}
\bibfield{author}{\bibinfo{person}{Mikhail Tikhomirov} {and}
  \bibinfo{person}{Boris Dobrov}.} \bibinfo{year}{2018}\natexlab{}.
\newblock \showarticletitle{News Timeline Generation: Accounting for Structural
  Aspects and Temporal Nature of News Stream}. In
  \bibinfo{booktitle}{\emph{Data Analytics and Management in Data Intensive
  Domains}}, \bibfield{editor}{\bibinfo{person}{Leonid Kalinichenko},
  \bibinfo{person}{Yannis Manolopoulos}, \bibinfo{person}{Oleg Malkov},
  \bibinfo{person}{Nikolay Skvortsov}, \bibinfo{person}{Sergey Stupnikov},
  {and} \bibinfo{person}{Vladimir Sukhomlin}} (Eds.).
  \bibinfo{publisher}{Springer}, \bibinfo{address}{Cham},
  \bibinfo{pages}{267--280}.
\newblock
\showISBNx{978-3-319-96553-6}


\bibitem[\protect\citeauthoryear{Tran, Alrifai, and Herder}{Tran
  et~al\mbox{.}}{2015a}]%
        {tran2015timeline}
\bibfield{author}{\bibinfo{person}{Giang Tran}, \bibinfo{person}{Mohammad
  Alrifai}, {and} \bibinfo{person}{Eelco Herder}.}
  \bibinfo{year}{2015}\natexlab{a}.
\newblock \showarticletitle{Timeline Summarization from Relevant Headlines}. In
  \bibinfo{booktitle}{\emph{Advances in Information Retrieval}},
  \bibfield{editor}{\bibinfo{person}{Allan Hanbury}, \bibinfo{person}{Gabriella
  Kazai}, \bibinfo{person}{Andreas Rauber}, {and} \bibinfo{person}{Norbert
  Fuhr}} (Eds.). \bibinfo{publisher}{Springer}, \bibinfo{address}{Cham},
  \bibinfo{pages}{245--256}.
\newblock
\showISBNx{978-3-319-16354-3}


\bibitem[\protect\citeauthoryear{Tran, Tran, Tran, Alrifai, and Kanhabua}{Tran
  et~al\mbox{.}}{2013}]%
        {tran2013leveraging}
\bibfield{author}{\bibinfo{person}{{Giang Binh} Tran}, \bibinfo{person}{Tuan
  Tran}, \bibinfo{person}{Nam-Khanh Tran}, \bibinfo{person}{Mohammad Alrifai},
  {and} \bibinfo{person}{Nattiya Kanhabua}.} \bibinfo{year}{2013}\natexlab{}.
\newblock \showarticletitle{Leveraging Learning To Rank in an Optimization
  Framework for Timeline Summarization}. In \bibinfo{booktitle}{\emph{SIGIR
  2013 Workshop on Time-aware Information Access (TAIA'2013)}}.
  \bibinfo{publisher}{ACM}, \bibinfo{address}{Dublin, Ireland},
  \bibinfo{numpages}{4}~pages.
\newblock


\bibitem[\protect\citeauthoryear{Tran, Niederee, Kanhabua, Gadiraju, and
  Anand}{Tran et~al\mbox{.}}{2015b}]%
        {tran2015balancing}
\bibfield{author}{\bibinfo{person}{Tuan~A. Tran}, \bibinfo{person}{Claudia
  Niederee}, \bibinfo{person}{Nattiya Kanhabua}, \bibinfo{person}{Ujwal
  Gadiraju}, {and} \bibinfo{person}{Avishek Anand}.}
  \bibinfo{year}{2015}\natexlab{b}.
\newblock \showarticletitle{Balancing Novelty and Salience: Adaptive Learning
  to Rank Entities for Timeline Summarization of High-Impact Events}. In
  \bibinfo{booktitle}{\emph{Proc. of the 24th ACM Intl. on Conf. on Information
  and Knowledge Management}} \emph{(\bibinfo{series}{CIKM '15})}.
  \bibinfo{publisher}{ACM}, \bibinfo{address}{NY, USA},
  \bibinfo{pages}{1201–1210}.
\newblock
\showISBNx{9781450337946}


\bibitem[\protect\citeauthoryear{Uramoto and Takeda}{Uramoto and
  Takeda}{1998}]%
        {uramoto1998method}
\bibfield{author}{\bibinfo{person}{Naohiko Uramoto} {and}
  \bibinfo{person}{Koichi Takeda}.} \bibinfo{year}{1998}\natexlab{}.
\newblock \showarticletitle{A Method for Relating Multiple Newspaper Articles
  by Using Graphs, and Its Application to Webcasting}. In
  \bibinfo{booktitle}{\emph{Proc. of the 36th Annual Meeting of the Association
  for Computational Linguistics and 17th Intl. Conf. on Computational
  Linguistics - Volume 2}} \emph{(\bibinfo{series}{ACL '98/COLING '98})}.
  \bibinfo{publisher}{Association for Computational Linguistics},
  \bibinfo{address}{USA}, \bibinfo{pages}{1307–1313}.
\newblock


\bibitem[\protect\citeauthoryear{Valls-Vargas, Zhu, and
  Onta\~{n}\'{o}n}{Valls-Vargas et~al\mbox{.}}{2017}]%
        {valls2017computational}
\bibfield{author}{\bibinfo{person}{Josep Valls-Vargas}, \bibinfo{person}{Jichen
  Zhu}, {and} \bibinfo{person}{Santiago Onta\~{n}\'{o}n}.}
  \bibinfo{year}{2017}\natexlab{}.
\newblock \showarticletitle{From Computational Narrative Analysis to
  Generation: A Preliminary Review}. In \bibinfo{booktitle}{\emph{Proc. of the
  12th Intl. Conf. on the Foundations of Digital Games}}
  \emph{(\bibinfo{series}{FDG '17})}. \bibinfo{publisher}{ACM},
  \bibinfo{address}{NY, USA}, Article \bibinfo{articleno}{55},
  \bibinfo{numpages}{4}~pages.
\newblock
\showISBNx{9781450353199}


\bibitem[\protect\citeauthoryear{Voorhees}{Voorhees}{1986}]%
        {voorhees1986implementing}
\bibfield{author}{\bibinfo{person}{Ellen~M Voorhees}.}
  \bibinfo{year}{1986}\natexlab{}.
\newblock \showarticletitle{Implementing agglomerative hierarchic clustering
  algorithms for use in document retrieval}.
\newblock \bibinfo{journal}{\emph{Information Processing \& Management}}
  \bibinfo{volume}{22}, \bibinfo{number}{6} (\bibinfo{year}{1986}),
  \bibinfo{pages}{465--476}.
\newblock


\bibitem[\protect\citeauthoryear{Wake}{Wake}{2013}]%
        {wake2013narrative}
\bibfield{author}{\bibinfo{person}{Paul Wake}.}
  \bibinfo{year}{2013}\natexlab{}.
\newblock \showarticletitle{Narrative and narratology}.
\newblock In \bibinfo{booktitle}{\emph{The Routledge Companion to Critical and
  Cultural Theory}}. \bibinfo{publisher}{Routledge}, \bibinfo{address}{NY,
  USA}, \bibinfo{pages}{39--52}.
\newblock


\bibitem[\protect\citeauthoryear{Wang, He, and Zhou}{Wang
  et~al\mbox{.}}{2018}]%
        {wang2018event}
\bibfield{author}{\bibinfo{person}{Chengyu Wang}, \bibinfo{person}{Xiaofeng
  He}, {and} \bibinfo{person}{Aoying Zhou}.} \bibinfo{year}{2018}\natexlab{}.
\newblock \showarticletitle{Event phase oriented news summarization}.
\newblock \bibinfo{journal}{\emph{World Wide Web}} \bibinfo{volume}{21},
  \bibinfo{number}{4} (\bibinfo{year}{2018}), \bibinfo{pages}{1069--1092}.
\newblock


\bibitem[\protect\citeauthoryear{Wei, Lee, Chiang, Chen, and Yang}{Wei
  et~al\mbox{.}}{2014}]%
        {wei2014exploiting}
\bibfield{author}{\bibinfo{person}{Chih-Ping Wei}, \bibinfo{person}{Yen-Hsien
  Lee}, \bibinfo{person}{Yu-Sheng Chiang}, \bibinfo{person}{Chun-Ta Chen},
  {and} \bibinfo{person}{Christopher~CC Yang}.}
  \bibinfo{year}{2014}\natexlab{}.
\newblock \showarticletitle{Exploiting temporal characteristics of features for
  effectively discovering event episodes from news corpora}.
\newblock \bibinfo{journal}{\emph{Journal of the Association for Information
  Science and Technology}} \bibinfo{volume}{65}, \bibinfo{number}{3}
  (\bibinfo{year}{2014}), \bibinfo{pages}{621--634}.
\newblock


\bibitem[\protect\citeauthoryear{Weigel and Fein}{Weigel and Fein}{1994}]%
        {weigel1994normalizing}
\bibfield{author}{\bibinfo{person}{Achim Weigel} {and} \bibinfo{person}{Frank
  Fein}.} \bibinfo{year}{1994}\natexlab{}.
\newblock \showarticletitle{Normalizing the weighted edit distance}. In
  \bibinfo{booktitle}{\emph{Proc. of the 12th IAPR Intl. Conf. on Pattern
  Recognition, Vol. 3-Conf. C: Signal Processing (Cat. No. 94CH3440-5)}},
  Vol.~\bibinfo{volume}{2}. IEEE, \bibinfo{publisher}{IEEE},
  \bibinfo{address}{Jerusalem, Israel}, \bibinfo{pages}{399--402}.
\newblock


\bibitem[\protect\citeauthoryear{Wenskovitch, Bradel, Dowling, House, and
  North}{Wenskovitch et~al\mbox{.}}{2018}]%
        {wenskovitch2018effect}
\bibfield{author}{\bibinfo{person}{John Wenskovitch}, \bibinfo{person}{Lauren
  Bradel}, \bibinfo{person}{Michelle Dowling}, \bibinfo{person}{Leanna House},
  {and} \bibinfo{person}{Chris North}.} \bibinfo{year}{2018}\natexlab{}.
\newblock \showarticletitle{The effect of semantic interaction on foraging in
  text analysis}. In \bibinfo{booktitle}{\emph{2018 IEEE Conf. on Visual
  Analytics Science and Technology (VAST)}}. IEEE, \bibinfo{publisher}{IEEE},
  \bibinfo{address}{Berlin, Germany}, \bibinfo{pages}{13--24}.
\newblock


\bibitem[\protect\citeauthoryear{Wilson, Zhou, and Starbird}{Wilson
  et~al\mbox{.}}{2018}]%
        {wilson2018assembling}
\bibfield{author}{\bibinfo{person}{Tom Wilson}, \bibinfo{person}{Kaitlyn Zhou},
  {and} \bibinfo{person}{Kate Starbird}.} \bibinfo{year}{2018}\natexlab{}.
\newblock \showarticletitle{Assembling strategic narratives: Information
  operations as collaborative work within an online community}.
\newblock \bibinfo{journal}{\emph{Proc. of the ACM on HCI}}
  \bibinfo{volume}{2}, \bibinfo{number}{CSCW} (\bibinfo{year}{2018}),
  \bibinfo{pages}{1--26}.
\newblock


\bibitem[\protect\citeauthoryear{Wu, Morstatter, Carley, and Liu}{Wu
  et~al\mbox{.}}{2019}]%
        {wu2019misinformation}
\bibfield{author}{\bibinfo{person}{Liang Wu}, \bibinfo{person}{Fred
  Morstatter}, \bibinfo{person}{Kathleen~M Carley}, {and} \bibinfo{person}{Huan
  Liu}.} \bibinfo{year}{2019}\natexlab{}.
\newblock \showarticletitle{Misinformation in social media: definition,
  manipulation, and detection}.
\newblock \bibinfo{journal}{\emph{ACM SIGKDD Explorations Newsletter}}
  \bibinfo{volume}{21}, \bibinfo{number}{2} (\bibinfo{year}{2019}),
  \bibinfo{pages}{80--90}.
\newblock


\bibitem[\protect\citeauthoryear{Wu, Sun, and Yan}{Wu et~al\mbox{.}}{2017}]%
        {wu2017event}
\bibfield{author}{\bibinfo{person}{Yaguang Wu}, \bibinfo{person}{Haichun Sun},
  {and} \bibinfo{person}{Chungang Yan}.} \bibinfo{year}{2017}\natexlab{}.
\newblock \showarticletitle{An event timeline extraction method based on news
  corpus}. In \bibinfo{booktitle}{\emph{2017 IEEE 2nd Intl. Conf. on Big Data
  Analysis (ICBDA)}}. \bibinfo{publisher}{IEEE}, \bibinfo{address}{Beijing,
  China}, \bibinfo{pages}{697--702}.
\newblock


\bibitem[\protect\citeauthoryear{Xu and Tang}{Xu and Tang}{2018}]%
        {xu2018generating}
\bibfield{author}{\bibinfo{person}{Nuo Xu} {and} \bibinfo{person}{Xijin Tang}.}
  \bibinfo{year}{2018}\natexlab{}.
\newblock \showarticletitle{Generating Risk Maps for Evolution Analysis of
  Societal Risk Events}. In \bibinfo{booktitle}{\emph{Knowledge and Systems
  Sciences}}, \bibfield{editor}{\bibinfo{person}{Jian Chen},
  \bibinfo{person}{Yuji Yamada}, \bibinfo{person}{Mina Ryoke}, {and}
  \bibinfo{person}{Xijin Tang}} (Eds.). \bibinfo{publisher}{Springer},
  \bibinfo{address}{Singapore}, \bibinfo{pages}{115--128}.
\newblock
\showISBNx{978-981-13-3149-7}


\bibitem[\protect\citeauthoryear{Yan, Kong, Huang, Wan, Li, and Zhang}{Yan
  et~al\mbox{.}}{2011a}]%
        {yan2011timeline}
\bibfield{author}{\bibinfo{person}{Rui Yan}, \bibinfo{person}{Liang Kong},
  \bibinfo{person}{Congrui Huang}, \bibinfo{person}{Xiaojun Wan},
  \bibinfo{person}{Xiaoming Li}, {and} \bibinfo{person}{Yan Zhang}.}
  \bibinfo{year}{2011}\natexlab{a}.
\newblock \showarticletitle{Timeline generation through evolutionary
  trans-temporal summarization}. In \bibinfo{booktitle}{\emph{Proc. of the 2011
  Conf. on Empirical Methods in Natural Language Processing}}.
  \bibinfo{publisher}{ACL}, \bibinfo{address}{Edinburgh, Scotland, UK},
  \bibinfo{pages}{433--443}.
\newblock


\bibitem[\protect\citeauthoryear{Yan, Wan, Otterbacher, Kong, Li, and
  Zhang}{Yan et~al\mbox{.}}{2011b}]%
        {yan2011evolutionary}
\bibfield{author}{\bibinfo{person}{Rui Yan}, \bibinfo{person}{Xiaojun Wan},
  \bibinfo{person}{Jahna Otterbacher}, \bibinfo{person}{Liang Kong},
  \bibinfo{person}{Xiaoming Li}, {and} \bibinfo{person}{Yan Zhang}.}
  \bibinfo{year}{2011}\natexlab{b}.
\newblock \showarticletitle{Evolutionary Timeline Summarization: A Balanced
  Optimization Framework via Iterative Substitution}. In
  \bibinfo{booktitle}{\emph{Proc. of the 34th Intl. ACM SIGIR Conf. on Research
  and Development in Information Retrieval}}. \bibinfo{publisher}{ACM},
  \bibinfo{address}{NY, USA}, \bibinfo{pages}{745–754}.
\newblock
\showISBNx{9781450307574}


\bibitem[\protect\citeauthoryear{Yang, Shi, and Wei}{Yang
  et~al\mbox{.}}{2006}]%
        {yang2006tracing}
\bibfield{author}{\bibinfo{person}{Christopher~C. Yang},
  \bibinfo{person}{Xiaodong Shi}, {and} \bibinfo{person}{Chih-Ping Wei}.}
  \bibinfo{year}{2006}\natexlab{}.
\newblock \showarticletitle{Tracing the Event Evolution of Terror Attacks from
  On-Line News}. In \bibinfo{booktitle}{\emph{Intelligence and Security
  Informatics}}, \bibfield{editor}{\bibinfo{person}{Sharad Mehrotra},
  \bibinfo{person}{Daniel~D. Zeng}, \bibinfo{person}{Hsinchun Chen},
  \bibinfo{person}{Bhavani Thuraisingham}, {and} \bibinfo{person}{Fei-Yue
  Wang}} (Eds.). \bibinfo{publisher}{Springer}, \bibinfo{address}{Berlin,
  Heidelberg}, \bibinfo{pages}{343--354}.
\newblock
\showISBNx{978-3-540-34479-7}


\bibitem[\protect\citeauthoryear{Yang, Shi, and Wei}{Yang
  et~al\mbox{.}}{2009}]%
        {yang2009discovering}
\bibfield{author}{\bibinfo{person}{Christopher~C Yang},
  \bibinfo{person}{Xiaodong Shi}, {and} \bibinfo{person}{Chih-Ping Wei}.}
  \bibinfo{year}{2009}\natexlab{}.
\newblock \showarticletitle{Discovering event evolution graphs from news
  corpora}.
\newblock \bibinfo{journal}{\emph{IEEE Transactions on Systems, Man, and
  Cybernetics-Part A: Systems and Humans}} \bibinfo{volume}{39},
  \bibinfo{number}{4} (\bibinfo{year}{2009}), \bibinfo{pages}{850--863}.
\newblock


\bibitem[\protect\citeauthoryear{Yilmaz, Kanoulas, and Aslam}{Yilmaz
  et~al\mbox{.}}{2008}]%
        {yilmaz2008simple}
\bibfield{author}{\bibinfo{person}{Emine Yilmaz}, \bibinfo{person}{Evangelos
  Kanoulas}, {and} \bibinfo{person}{Javed~A. Aslam}.}
  \bibinfo{year}{2008}\natexlab{}.
\newblock \showarticletitle{A Simple and Efficient Sampling Method for
  Estimating AP and NDCG}. In \bibinfo{booktitle}{\emph{Proc. of the 31st
  Annual Intl. ACM SIGIR Conf. on Research and Development in Information
  Retrieval}} \emph{(\bibinfo{series}{SIGIR '08})}. \bibinfo{publisher}{ACM},
  \bibinfo{address}{NY, USA}, \bibinfo{pages}{603–610}.
\newblock
\showISBNx{9781605581644}


\bibitem[\protect\citeauthoryear{Yu and Hatzivassiloglou}{Yu and
  Hatzivassiloglou}{2003}]%
        {yu2003towards}
\bibfield{author}{\bibinfo{person}{Hong Yu} {and} \bibinfo{person}{Vasileios
  Hatzivassiloglou}.} \bibinfo{year}{2003}\natexlab{}.
\newblock \showarticletitle{Towards answering opinion questions: Separating
  facts from opinions and identifying the polarity of opinion sentences}. In
  \bibinfo{booktitle}{\emph{Proc. of the 2003 Conf. on Empirical methods in
  natural language processing}}. \bibinfo{publisher}{ACL},
  \bibinfo{address}{Sapporo, Japan}, \bibinfo{pages}{129--136}.
\newblock


\bibitem[\protect\citeauthoryear{Yu, Jatowt, Doucet, Sugiyama, and
  Yoshikawa}{Yu et~al\mbox{.}}{2021}]%
        {yu2021multi}
\bibfield{author}{\bibinfo{person}{Yi Yu}, \bibinfo{person}{Adam Jatowt},
  \bibinfo{person}{Antoine Doucet}, \bibinfo{person}{Kazunari Sugiyama}, {and}
  \bibinfo{person}{Masatoshi Yoshikawa}.} \bibinfo{year}{2021}\natexlab{}.
\newblock \showarticletitle{Multi-timeline summarization (mtls): Improving
  timeline summarization by generating multiple summaries}. In
  \bibinfo{booktitle}{\emph{Proc. of the 59th Annual Meeting of the Association
  for Computational Linguistics and the 11th Intl. Joint Conf. on Natural
  Language Processing (Volume 1: Long Papers)}}. \bibinfo{publisher}{ACL},
  \bibinfo{address}{Online}, \bibinfo{pages}{377--387}.
\newblock


\bibitem[\protect\citeauthoryear{Yuan, Zhou, and Zhou}{Yuan
  et~al\mbox{.}}{2019}]%
        {yuan2019dtexsl}
\bibfield{author}{\bibinfo{person}{Ruifeng Yuan}, \bibinfo{person}{Qifeng
  Zhou}, {and} \bibinfo{person}{Wubai Zhou}.} \bibinfo{year}{2019}\natexlab{}.
\newblock \showarticletitle{dTexSL: A dynamic disaster textual storyline
  generating framework}.
\newblock \bibinfo{journal}{\emph{World Wide Web}} \bibinfo{volume}{22},
  \bibinfo{number}{5} (\bibinfo{year}{2019}), \bibinfo{pages}{1913--1933}.
\newblock


\bibitem[\protect\citeauthoryear{Zhao}{Zhao}{2021}]%
        {zhao2021event}
\bibfield{author}{\bibinfo{person}{Liang Zhao}.}
  \bibinfo{year}{2021}\natexlab{}.
\newblock \showarticletitle{Event Prediction in the Big Data Era: A Systematic
  Survey}.
\newblock \bibinfo{journal}{\emph{ACM Computing Surveys (CSUR)}}
  \bibinfo{volume}{54}, \bibinfo{number}{5} (\bibinfo{year}{2021}),
  \bibinfo{pages}{1--37}.
\newblock


\bibitem[\protect\citeauthoryear{Zhou, Shen, Li, Chen, and Xie}{Zhou
  et~al\mbox{.}}{2014}]%
        {zhou2014generating}
\bibfield{author}{\bibinfo{person}{Wubai Zhou}, \bibinfo{person}{Chao Shen},
  \bibinfo{person}{Tao Li}, \bibinfo{person}{Shu-Ching Chen}, {and}
  \bibinfo{person}{Ning Xie}.} \bibinfo{year}{2014}\natexlab{}.
\newblock \showarticletitle{Generating textual storyline to improve situation
  awareness in disaster management}. In \bibinfo{booktitle}{\emph{Proc. of the
  2014 IEEE 15th Intl. Conf. on Information Reuse and Integration}}. IEEE,
  \bibinfo{publisher}{IEEE}, \bibinfo{address}{Redwood City, CA, USA},
  \bibinfo{pages}{585--592}.
\newblock


\bibitem[\protect\citeauthoryear{Zhou, Shen, Li, Chen, Xie, and Iyengar}{Zhou
  et~al\mbox{.}}{2018}]%
        {zhou2018new}
\bibfield{author}{\bibinfo{person}{Wubai Zhou}, \bibinfo{person}{Chao Shen},
  \bibinfo{person}{Tao Li}, \bibinfo{person}{Shu-Ching Chen},
  \bibinfo{person}{Ning Xie}, {and} \bibinfo{person}{SS Iyengar}.}
  \bibinfo{year}{2018}\natexlab{}.
\newblock \showarticletitle{A New Two-layer Storyline Generation Framework for
  Disaster Management}.
\newblock \bibinfo{journal}{\emph{Intl. Journal of Next-Generation Computing}}
  \bibinfo{volume}{9}, \bibinfo{number}{3} (\bibinfo{year}{2018}),
  \bibinfo{numpages}{13}~pages.
\newblock


\bibitem[\protect\citeauthoryear{Zhu and Oates}{Zhu and Oates}{2012}]%
        {zhu2012finding}
\bibfield{author}{\bibinfo{person}{Xianshu Zhu} {and} \bibinfo{person}{Tim
  Oates}.} \bibinfo{year}{2012}\natexlab{}.
\newblock \showarticletitle{Finding story chains in newswire articles}. In
  \bibinfo{booktitle}{\emph{2012 IEEE 13th Intl. Conf. on Information Reuse \&
  Integration (IRI)}}. \bibinfo{publisher}{IEEE}, \bibinfo{address}{Las Vegas,
  NV, USA}, \bibinfo{pages}{93--100}.
\newblock


\bibitem[\protect\citeauthoryear{Zhu and Oates}{Zhu and Oates}{2014}]%
        {zhu2014finding}
\bibfield{author}{\bibinfo{person}{Xianshu Zhu} {and} \bibinfo{person}{Tim
  Oates}.} \bibinfo{year}{2014}\natexlab{}.
\newblock \showarticletitle{Finding story chains in newswire articles using
  random walks}.
\newblock \bibinfo{journal}{\emph{Information Systems Frontiers}}
  \bibinfo{volume}{16}, \bibinfo{number}{5} (\bibinfo{year}{2014}),
  \bibinfo{pages}{753--769}.
\newblock


\end{thebibliography}

\end{document}